\pdfoutput=1
\documentclass[10pt,twocolumn,letterpaper]{article}
\usepackage[pagenumbers]{cvpr}
\ifdefined\pdfomitinfodict \pdfomitinfodict=1 \fi
\ifdefined\pdfsuppressptexinfo \pdfsuppressptexinfo=-1 \fi
\usepackage[utf8]{inputenc}
\usepackage[T1]{fontenc}
\usepackage{microtype,nicefrac,multirow,enumitem,placeins}
\usepackage{algorithm,algpseudocode,array,tabularx,colortbl,makecell}
\newcolumntype{Y}{>{\centering\arraybackslash\leavevmode}X}
\definecolor{cvprblue}{rgb}{0.21,0.49,0.74}
\usepackage[pagebackref,breaklinks,colorlinks,allcolors=cvprblue]{hyperref}
\makeatletter
\def\theHALG@line{\thealgorithm.\arabic{ALG@line}}
\makeatother
\newcommand{\PaperTitle}{OA-WAM: Object-Addressable World Action Model\\ for Robust Robot Manipulation}
\newcommand{\method}{OA-WAM}
\newcommand{\tok}[1]{\textnormal{\textsc{#1}}}
\definecolor{ourshl}{RGB}{218,236,255}
\newcommand{\hl}[1]{\cellcolor{ourshl}#1}
\DeclareRobustCommand{\statstd}[1]{%
  \nobreak\hspace{0.06em}\textsubscript{\normalfont\ensuremath{\pm}#1}%
}
\newcommand{\methodours}{\textbf{OA-WAM (Ours)}}
\newcommand{\sectionrow}[2]{\multicolumn{#1}{@{}l}{\textit{\hspace{0.15em}#2}}\\}

\def\confName{CVPR}
\def\confYear{2027}

\usepackage[pass]{geometry}
\usepackage{needspace}
\title{\PaperTitle}
\author{%
\setlength{\tabcolsep}{0.75em}%
\begin{tabular}[b]{@{}ccccc@{}}
Yushan Liu$^{1}$ & Peibo Sun$^{2}$ & Shoujie Li$^{3}$ & Yifan Xie$^{1}$ & Lingfeng Zhang$^{1}$\\[0.25em]
Xintao Chao$^{1}$ & Shiyuan Dong$^{2}$ & Fang Chen$^{2}$ & Xiao-Ping Zhang$^{1}$ & Wenbo Ding$^{1\dagger}$
\end{tabular}\\[1.4em]
{\normalsize $^{1}$Tsinghua University\hspace{1.8em}$^{2}$Shanghai Jiao Tong University\hspace{1.8em}$^{3}$Nanyang Technological University}%
}
\newcommand{\AfterPaperTitle}{%
  \begingroup
  \renewcommand\thefootnote{}%
  \footnotetext{$^{\dagger}$Corresponding author.}%
  \endgroup
}

\newcommand{\ArxivEdition}{}
\begin{document}
\providecommand{\AfterPaperTitle}{}
\providecommand{\LLMUsageStatement}{}
\maketitle
\AfterPaperTitle
\pdfbookmark[0]{Main paper}{oawam-main}
\begin{abstract}
World action models (WAMs) learn manipulation by predicting actions and future observations, yet their success drops sharply when the camera viewpoint, object layout or robot initial state changes. Such changes keep the instructed object the same but alter the appearance and pose features the policy observes, which in standard attention also form the keys that locate the object. We introduce \textbf{OA-WAM}, an object-addressable WAM that finds each object by an identity address fixed for the episode and reads its changing state. In every layer of a pretrained multimodal trunk, object-slot keys depend only on the address and frame position, and values carry the current state; the same addresses key the action pointer and index per-object forecasts, which can trigger re-grounding. Fine-tuned on standard LIBERO demonstrations, OA-WAM transfers zero-shot to LIBERO-Plus with $90.5\%$ success, outperforming all ten published baselines. It also leads on LIBERO ($99.0\%$) and, fine-tuned on BridgeData V2, on SimplerEnv WidowX ($88.9\%$). On five real-robot tasks, it succeeds in 89 of 100 nominal trials, versus 73 for the strongest baseline. Ablations trace its robustness to addressing rather than to object slots: with standard attention, slots do no better than a slot-free policy, and addressing them by identity raises geometric success by $10.7$ points. Forecasting the addressed objects is likewise a stronger training signal than forecasting image codes.
\end{abstract}

\section{Introduction}
\label{sec:intro}

Robot manipulation unfolds at the level of objects. An instruction such as ``put the red mug on the green tray'' asks the robot to find the named objects among distractors and act on them in closed loop. Vision-language-action (VLA) policies have made rapid progress by scaling vision-language backbones, robot data and action decoders~\citep{rt2,openvla,pi05}, and world action models (WAMs) add supervision by jointly predicting actions and future world states~\citep{rynnvla002,adaworldpolicy,dreamzero,liu2026pact}.

\begin{figure*}[!tbp]
  \centering
  \includegraphics[width=\linewidth]{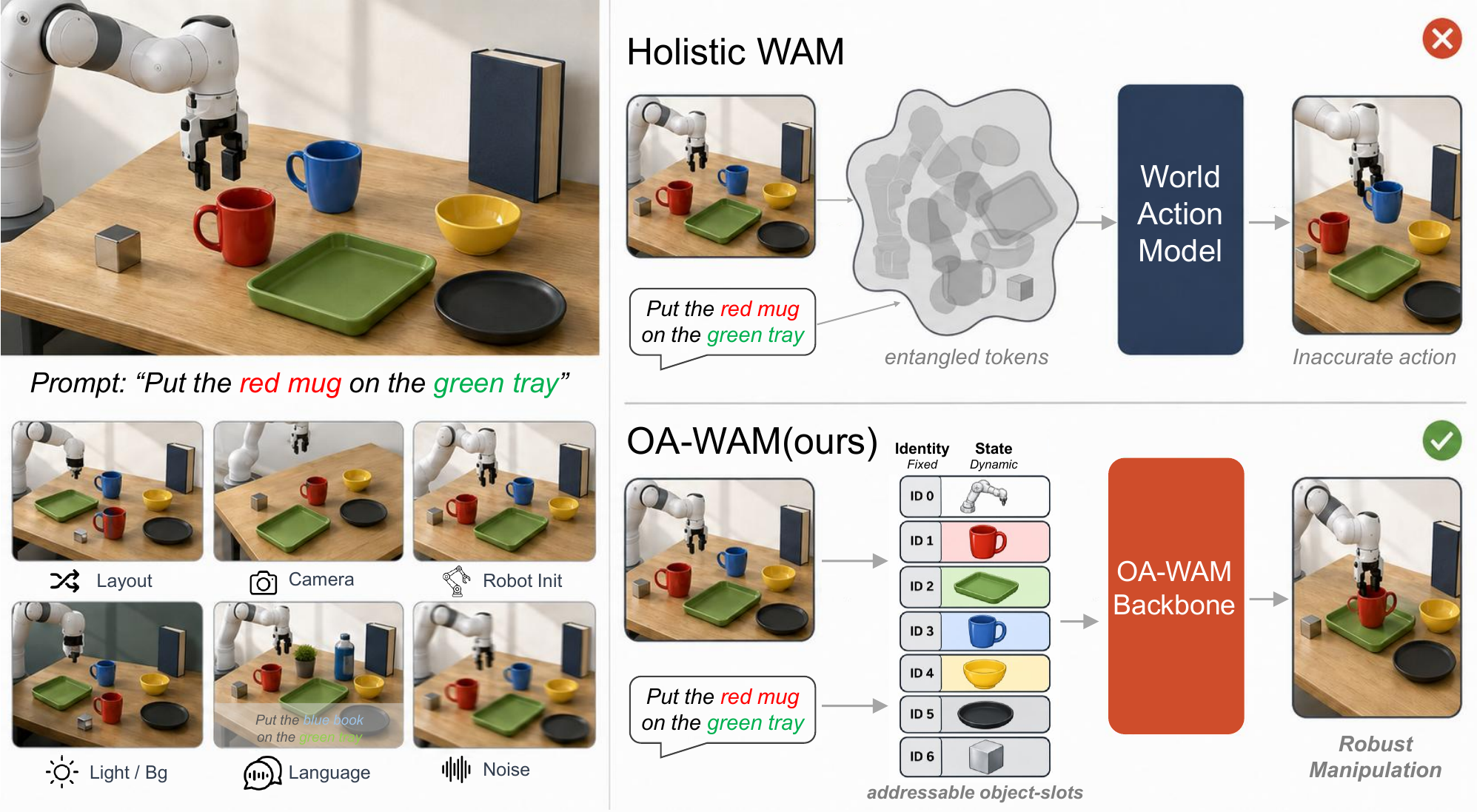}

  \caption{\textbf{Addressing objects by identity.} In this illustration, a holistic WAM entangles objects in scene tokens and grasps the blue mug instead of the requested red one (top right), whereas \method{} finds the red mug by its fixed identity and reads its current state (bottom right). Left: the task and its perturbed variants.}
  \label{fig:pipeline}

\end{figure*}

Yet these policies remain fragile when the camera viewpoint, object layout or robot initial state changes~\citep{liberoplus,colosseum,robustvla}. Such changes leave the instructed object the same but alter its appearance, its spatial relations and the actions needed to reach it. A robust policy must therefore keep two things apart: which object the instruction refers to, which stays fixed, and that object's state, which changes with every observation. Scene-level forecasts do not draw this line: future frames, visual tokens or global latents~\citep{uwm,worldvla,cotvla} describe how the scene will change, not what will happen to the red mug (Fig.~\ref{fig:pipeline}).

Object-centric representations make objects explicit. Slot-based world models improve sample efficiency and relational reasoning~\citep{focus,ocwm}, object-grounded VLAs use per-object states as the interface between language and action~\citep{goalvla,slotvla,lilovla}, and recent methods separate a persistent identity from a dynamic state~\citep{dssa,slotnarrative}, which ObjDream uses to bias attention~\citep{objdream}. In standard attention, however, an object's key is computed from its full representation. The features that locate the object are thus the very features a perturbation alters.

\begin{figure*}[!tbp]
  \centering
  \includegraphics[width=\linewidth]{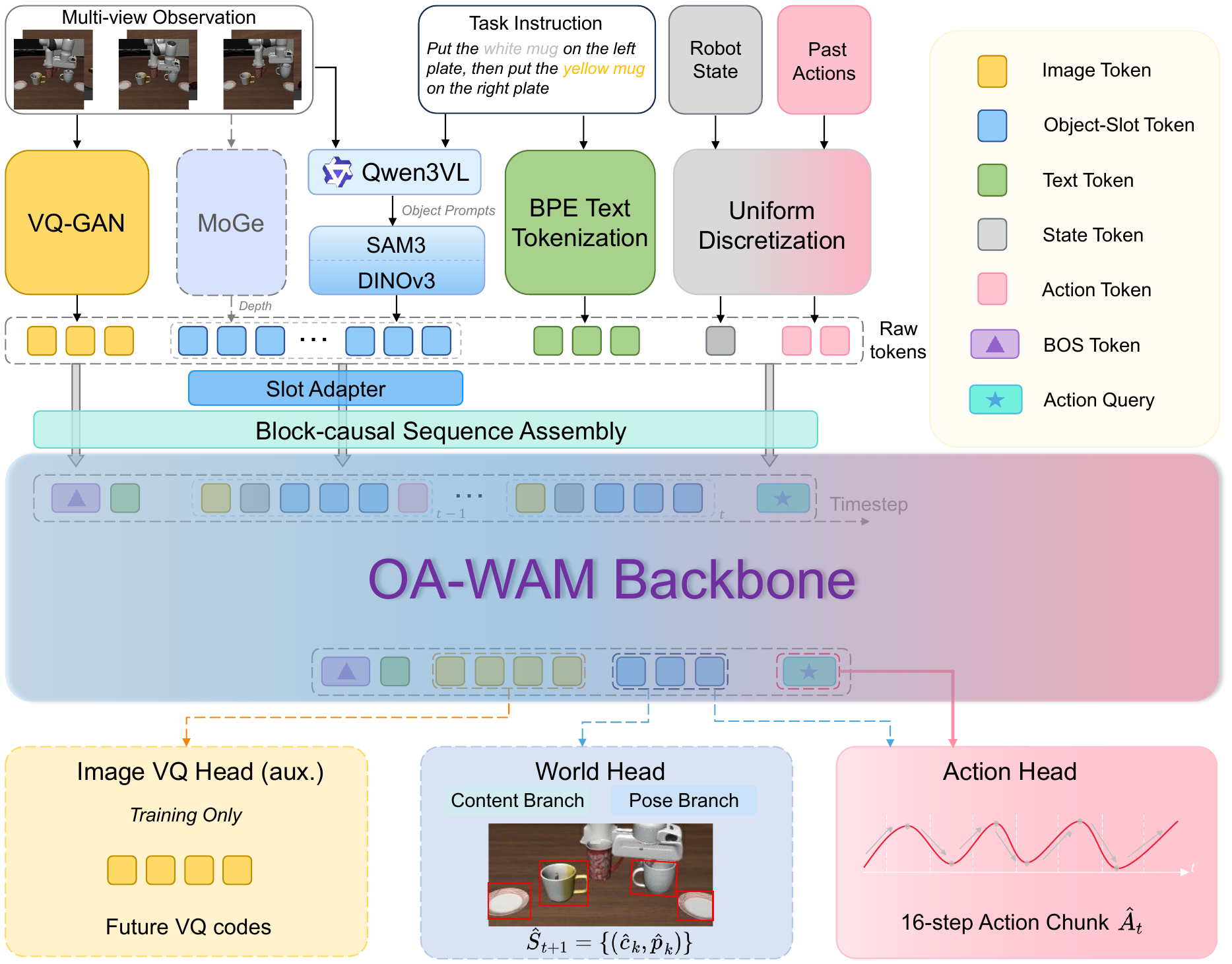}

  \caption{\textbf{\method{} architecture.} One identity address per object keys attention in every trunk layer, keys the action pointer and indexes the forecast. Object slots pair this fixed address with a dynamic state and join text, image, state and action tokens in one block-causal sequence. The action head outputs a $16$-step action chunk, and the world head forecasts each object's next content and pose before execution; the forecast can be checked against the next observation (Supp.~\ref{app:inference-detail}). An auxiliary image head predicts future VQ codes during training only.}
  \label{fig:arch}

\end{figure*}

We present \textbf{\method{}}, an object-addressable world action model that separates how an object is found from what is read from it (Fig.~\ref{fig:arch}). Inspired by object files and visual indexes~\citep{objectfiles,finst}, \method{} gives each object an identity address fixed for the episode and a state updated with every observation. Each layer of the pretrained trunk computes the slot key from the identity address and frame position with its existing attention parameters, while the value carries the object's current state. The key that finds the red mug thus never sees its pose, and moving the mug leaves it unchanged. The same addresses key the action pointer and index the world head's forecast of each object's next state.

Fine-tuned on standard LIBERO demonstrations, \method{} transfers zero-shot to LIBERO-Plus with $90.5\%$ overall success, ahead of all ten published baselines and of the public $\pi_{0.5}$ and Cosmos Policy checkpoints evaluated on the same trials (Tab.~\ref{tab:libero-plus}). It also leads on standard LIBERO ($99.0\%$) and SimplerEnv WidowX ($+8.1$ points; Tab.~\ref{tab:libero-simpler}) and completes 89 of 100 nominal real-robot trials versus 73 for the strongest baseline trained on the same demonstrations (Tab.~\ref{tab:real}). Ablations trace its robustness to addressing rather than to object slots: with standard attention, object slots reach $78.6\%$ success under geometric perturbations (Geo Avg), below a slot-free policy at $79.7\%$, and identity keys raise it to $89.3\%$.

Our contributions are as follows:
\begin{itemize}[leftmargin=1.2em]
    \item Object-addressable attention writes each object's identity address into every trunk layer and computes its slot key from the address and frame position alone, reusing the pretrained key projection without adding attention parameters (Sec.~\ref{sec:method:oa}).
    \item Address-indexed forecasts supervise the shared slot features during training and can trigger re-grounding at test time when the next observation departs from them (Sec.~\ref{sec:method:heads}).
    \item Matched ablations isolate each use of the address (Sec.~\ref{sec:experiments:ablations}): from the same correct binding, trunk identity keys select the instructed object more often; the identity-key pointer places a new target more often than a state-key pointer; and object-state forecasts outperform image-code forecasts as a training signal and add $1.8$ points as a recovery trigger beyond object disappearance.
\end{itemize}

\section{Related Work}

\paragraph{World action models for control.}
World action models couple policy learning with dynamics modeling by predicting future visual observations jointly with actions~\citep{uwm,worldvla,rynnvla002,dreamzero,liu2026pact}, a formulation also adopted in end-to-end autonomous driving~\citep{liu2026perceptdrive,zhong2026forgedrive}. CheckVLA verifies and revises an action suffix with an action-conditioned world model~\citep{checkvla}, and FFDC compares predicted visual dynamics with observations to adapt the executed horizon~\citep{ffdc}. Other monitors use forecast consistency~\citep{favor}, robot-conditioned object-motion likelihoods (RC-NF)~\citep{rcnf} or geometric predicates over persistent object records (POT-VLA)~\citep{potvla}. \method{} instead predicts the next state of each tracked object under its identity address, which supervises the features shared with the action head and can trigger re-grounding at test time.

\paragraph{Robust manipulation policies.}
VLA policies acquire general manipulation skills from vision-language pretraining and large-scale robot data~\citep{rt2,openvla,pi0,pi05}, typically collected by teleoperation~\citep{chao2025exo}. Controlled perturbation benchmarks show that high nominal success does not guarantee robustness to scene and instruction changes~\citep{liberoplus,liberopro}. Existing methods broaden training coverage through perturbation-based training and novel-view augmentation~\citep{strongvla,roviaug,vista}, or strengthen spatial modeling with 3D scene representations, multi-view fusion and active viewpoint control~\citep{peract,rvt,liu2025avr}. VIOLA attends to object proposals~\citep{violaZhu}, and GROOT uses object-centric 3D representations for control~\citep{grootZhu}. \method{} instead targets the interface through which the policy addresses objects, so that state updates leave the basis for finding an object unchanged.

\paragraph{Object-centric representations and identity-based addressing.}
Object-centric methods decompose scenes into object representations~\citep{slotattention,slotformer} and use them for language-conditioned control, relational reasoning and dynamics prediction~\citep{ocwm,slotvla,srwm}. Later work separates persistent identity from dynamic state~\citep{dssa,slotnarrative,objdream}. Perceptual anchoring links symbols to tracked observations~\citep{anchoringCS}, referent grounding resolves the instructed instance~\citep{refguard}, and neural binding and static-key, dynamic-value memories distinguish identity from changing attributes~\citep{bindingGreff,dkvmnZhang}. Compositional and positional attention decouple addressing from retrieval~\citep{mittal2022compositional,posattn}. \method{} brings this decoupling into a world action model built on a pretrained multimodal trunk, where one episode-fixed address keys the object in every trunk layer and in the action pointer and indexes its forecast.

\section{Method}
\label{sec:method}

\subsection{Problem setup}
\label{sec:method:setup}

At decision time $t$, the robot observes RGB images $\mathbf I_t$, proprioception $\mathbf q_t\in\mathbb R^{d_q}$, a language instruction $\ell$ and its executed action history $\mathbf a^{\mathrm{exec}}_{<t}$. Before execution, the policy jointly predicts a chunk of $H$ actions, $\mathbf A_t\in\mathbb R^{H\times d_a}$, and the content features and poses of $N$ objects at the next decision, $\widehat{\mathcal S}_{t+1}=\{(\hat{\mathbf c}_k^{t+1},\hat{\mathbf p}_k^{t+1})\}_{k=1}^N$:
\begin{equation}
\bigl(\mathbf{A}_t,\widehat{\mathcal{S}}_{t+1}\bigr)\;\sim\;\pi_\theta\ \!\bigl(\,\cdot\mid\mathbf{I}_{\le t},\,\mathbf{q}_{\le t},\,\ell,\,\mathbf{a}^{\mathrm{exec}}_{<t}\bigr).
\label{eq:policy}
\end{equation}
The robot makes its next decision at $t+1$ after executing the first $K\le H$ actions.

\subsection{Object-slot tokenization and unified sequence}
\label{sec:method:slot}

At the start of an episode, Qwen3-VL~\citep{qwen3vl} extracts object descriptions from the instruction and initial observation to prompt SAM~3~\citep{sam3}, which segments and tracks the objects. DINOv3~\citep{dinov3} descriptors pooled within each mask encode appearance, and poses are estimated from the geometry within each mask, which frozen MoGe-3 ViT-L~\citep{moge3} predicts from RGB.

The address of object $k$ combines a label $\boldsymbol\ell_k$ that embeds its description, its initial visual descriptor $\mathbf f_k^{(0)}$ and a learned identity code $\mathbf c_{\iota_k}$:
\begin{equation}
\mathbf{addr}_k
=f_{\mathrm{addr}}\!\left([\boldsymbol{\ell}_k;\mathbf{f}_k^{(0)}]\right)
+\mathbf{c}_{\iota_k}.
\label{eq:addr}
\end{equation}
The codes are sampled without replacement for each episode, independently of slot order~\citep{aot,motip}, so objects with the same description still receive distinct addresses. At each decision, $f_{\mathrm{cnt}}$ encodes the object's current descriptor, pose and label into the content $\mathbf{cnt}_k^t$ that represents its state. A frame encoding $\boldsymbol\pi^t$ and a role encoding $\boldsymbol\rho_k$ (robot, object or padding) complete the slot:
\begin{equation}
\mathbf{s}_k^t
=\bigl[\underbrace{\mathbf{addr}_k}_{32}\,\Vert\,
\underbrace{\mathbf{cnt}_k^t}_{256}\,\Vert\,
\underbrace{\boldsymbol{\pi}^t}_{16}\,\Vert\,
\underbrace{\boldsymbol{\rho}_k}_{16}\bigr]
\in\mathbb{R}^{320}.
\label{eq:slot}
\end{equation}
A shared adapter embeds each slot as one trunk token~\citep{llava,idefics2}.
The sequence begins with the BPE-tokenized instruction, followed by the last $T=4$ frame groups in temporal order. Each group contains external-camera image VQ codes, quantized proprioception and the slots of that frame; the wrist view contributes to slot appearance and pose. Historical groups also contain tokens for the first action executed after their observation, and a learnable action query \tok{[act\_q]} closes the sequence (Supp.~\ref{app:tokenizer-detail}).

\subsection{Object-addressable attention}
\label{sec:method:oa}

\begin{figure}[t]
  \centering
  \includegraphics[width=\linewidth]{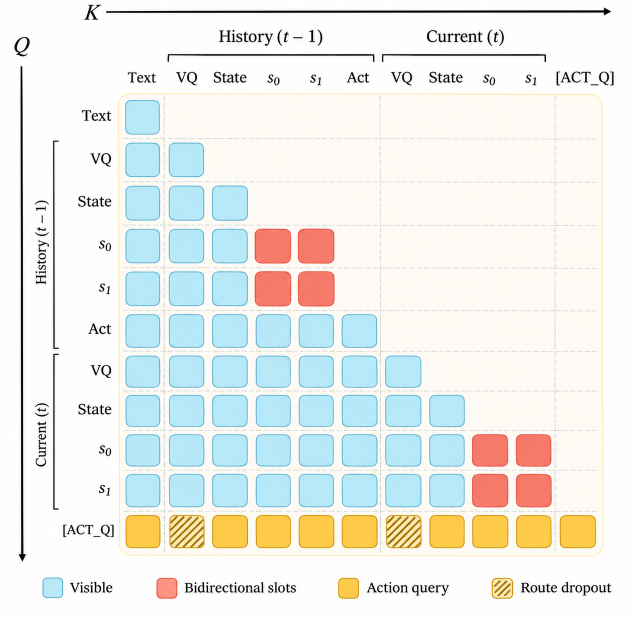}
  \caption{\textbf{OA-WAM attention mask.} Attention is causal across frames and bidirectional among slots within each frame (red). Hatched cells mark image codes directly accessible to the action query. During Stage~II training, these connections are masked jointly with probability $0.5$ (route dropout); they are open at inference.}
  \label{fig:attn-mask}
\end{figure}
\begin{table*}[t]
\centering
\caption{\textbf{Standard manipulation benchmarks.} Task success (\%) on four LIBERO suites and four SimplerEnv WidowX visual-matching tasks. Baseline task scores are from published evaluations; the \method{} row averages three training seeds. Dashes (---) indicate no matching WidowX score in the cited sources. \textbf{Bold} and \underline{underlining} mark the highest and second-highest scores in each column; ties share a rank. Sources and evaluation settings are in Supp.~\ref{app:eval}.}
\label{tab:libero-simpler}
\small
\setlength{\tabcolsep}{2.5pt}
\renewcommand{\arraystretch}{1.08}
\begin{tabularx}{\textwidth}{@{}l*{10}{Y}@{}}
\toprule
 & \multicolumn{5}{c}{\textbf{LIBERO}} & \multicolumn{5}{c}{\textbf{SimplerEnv WidowX (Visual Matching)}} \\
\cmidrule(lr){2-6} \cmidrule(lr){7-11}
Method & Spatial & Object & Goal & Long & Avg & \makecell{Spoon\\Towel} & \makecell{Carrot\\Plate} & \makecell{Stack\\Block} & \makecell{Eggplant\\Basket} & Avg \\
\midrule
\sectionrow{11}{Vision-Language-Action Models (VLA)}
OpenVLA-OFT~\citep{openvlaoft} & 97.6 & 98.4 & 97.9 & 94.5 & 97.1 & 34.2 & 30.0 & 30.0 & 72.5 & 41.7 \\
$\pi_0$-FAST~\citep{pi0fast} & 96.4 & 96.8 & 88.6 & 60.2 & 85.5 & 66.7 & 70.8 & 41.7 & 8.3 & 46.9 \\
$\pi_{0.5}$~\citep{pi05} & \underline{98.8} & 98.2 & 98.0 & 92.4 & 96.9 & 49.3 & 64.7 & 44.7 & 69.7 & 57.1 \\
GR00T N1.5~\citep{groot15} & 93.4 & 89.4 & 85.0 & 78.4 & 86.6 & 75.3 & 54.3 & 57.0 & 61.3 & 62.0 \\
UniVLA~\citep{bu2025univla} & 96.5 & 96.8 & 95.6 & 92.0 & 95.2 & 52.8 & 55.6 & 2.8 & \underline{80.6} & 48.0 \\
\midrule
\sectionrow{11}{World Action Models (WAM)}
Fast-WAM~\citep{fastwam} & 98.2 & \textbf{100.0} & 97.0 & 95.2 & 97.6 & --- & --- & --- & --- & --- \\
Cosmos Policy~\citep{cosmospolicy} & 98.1 & \textbf{100.0} & 98.2 & 97.6 & 98.5 & --- & --- & --- & --- & --- \\
LingBot-VA~\citep{lingbotva} & 98.5 & 99.6 & 97.2 & \textbf{98.5} & 98.5 & --- & --- & --- & --- & --- \\
VLA-JEPA~\citep{vlajepa} & 96.2 & 99.6 & 97.2 & 95.8 & 97.2 & 75.0 & 70.8 & 12.5 & 70.8 & 57.3 \\
InternVLA-A1.5~\citep{internvlaa15} & 98.6 & 99.8 & \underline{98.6} & \underline{98.4} & \underline{98.9} & \underline{92.4} & \underline{85.4} & \underline{69.5} & 75.7 & \underline{80.8} \\
\midrule
\hl{\methodours{}} & \hl{\textbf{99.0}} & \hl{\underline{99.9}} & \hl{\textbf{98.8}} & \hl{\underline{98.4}} & \hl{\textbf{99.0}} & \hl{\textbf{93.1}} & \hl{\textbf{93.1}} & \hl{\textbf{72.2}} & \hl{\textbf{97.2}} & \hl{\textbf{88.9}} \\
\bottomrule\end{tabularx}
\end{table*}

\begin{table*}[t]
\centering
\caption{\textbf{Zero-shot transfer from LIBERO to LIBERO-Plus.} Task success (\%). Local rows share 10,030 trial conditions: three runs per public checkpoint and three training seeds for \method{}. Other rows are published; $^{\dagger}$also evaluated locally. Geo Avg weights Camera, Robot init and Layout equally; local Overall pools all seven axes. Bold/underline: best/second-best; dashes: unreported. Sources: Supp.~\ref{app:eval}.}
\label{tab:libero-plus}
\small
\setlength{\tabcolsep}{2.5pt}
\renewcommand{\arraystretch}{1.08}
\begin{tabularx}{\textwidth}{@{}l*{9}{Y}@{}}
\toprule
Method & Camera & \makecell{Robot\\init} & Layout & \makecell{Geo\\Avg} & Light & BG & Lang & Noise & Overall \\
\midrule
\sectionrow{10}{Vision-Language-Action Models (VLA)}
OpenVLA-OFT & 56.4 & 31.9 & 74.2 & 54.2 & 88.7 & 93.3 & 79.5 & 75.8 & 69.6 \\
$\pi_0$-FAST & 65.1 & 21.6 & 68.8 & 51.8 & 73.2 & 73.2 & 61.0 & 74.4 & 61.6 \\
$\pi_{0.5}{}^{\dagger}$ & 75.4 & 77.5 & \underline{85.7} & 79.5 & \textbf{96.9} & 94.6 & 85.6 & 89.7 & \underline{85.7} \\
GR00T N1.5 & --- & --- & --- & --- & --- & --- & --- & --- & 66.4 \\
UniVLA & 1.8 & 46.2 & 31.9 & 26.6 & 69.0 & 81.0 & 69.6 & 21.2 & 42.9 \\
\midrule
\sectionrow{10}{World Action Models (WAM)}
Fast-WAM & 16.4 & 44.5 & 60.7 & 40.5 & 78.2 & 53.7 & 68.9 & 37.7 & 51.5 \\
Cosmos Policy$^{\dagger}$ & 75.8 & 63.3 & 82.2 & 73.8 & 96.5 & 88.9 & 81.7 & \underline{92.7} & 82.2 \\
LingBot-VA & 40.9 & \underline{83.0} & 76.2 & 66.7 & 82.3 & 53.1 & 86.4 & 64.4 & 69.5 \\
VLA-JEPA & 63.3 & 67.1 & 85.1 & 71.8 & 95.6 & 93.6 & 85.4 & 66.3 & 79.5 \\
InternVLA-A1.5 & \underline{83.1} & 55.1 & 85.2 & 74.5 & 96.4 & \underline{98.2} & \underline{86.9} & \textbf{95.6} & 84.8 \\
\midrule
$\pi_{0.5}$ (local) & 77.2 & 77.6 & 84.3 & \underline{79.7} & 95.9 & 94.9 & 84.7 & 83.1 & 84.5 \\
Cosmos Policy (local) & 75.4 & 64.2 & 81.7 & 73.7 & 96.4 & 88.4 & 81.6 & 92.4 & 82.1 \\
\hl{\methodours{}} & \hl{\textbf{89.9}} & \hl{\textbf{90.7}} & \hl{\textbf{87.4}} & \hl{\textbf{89.3}} & \hl{\underline{96.6}} & \hl{\textbf{98.4}} & \hl{\textbf{89.0}} & \hl{85.8} & \hl{\textbf{90.5}} \\
\bottomrule
\end{tabularx}
\end{table*}

\begin{figure*}[t]
  \centering
  \includegraphics[width=\linewidth]{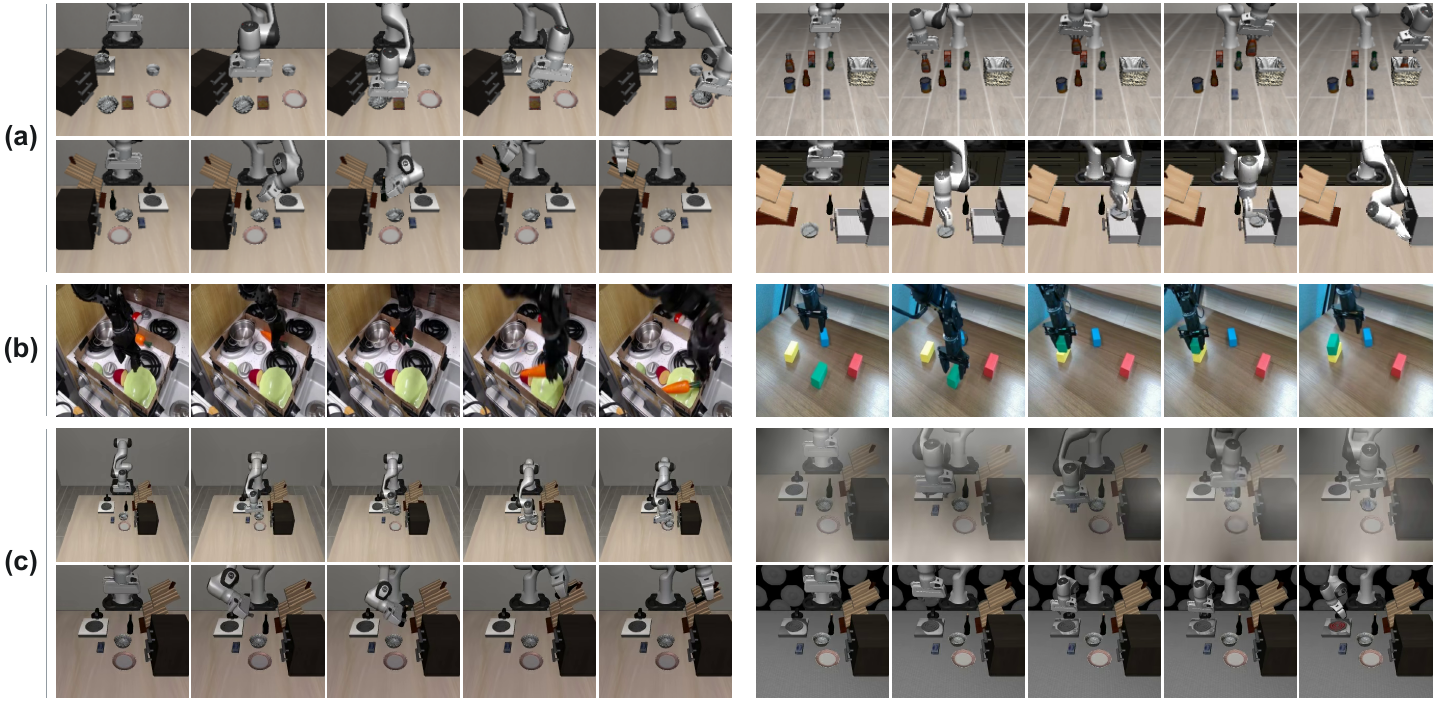}
  \caption{\textbf{Closed-loop \method{} rollouts.} Each strip shows five frames of one rollout in temporal order. \textbf{(a) LIBERO:} move the black bowl beside the cookie box to the plate (Spatial), the ketchup to the basket (Object) and the wine bottle to the rack (Goal); put the bowl in the bottom drawer and close it (Long). \textbf{(b) WidowX:} put the carrot on the plate; stack the green block on the yellow block. \textbf{(c) LIBERO-Plus:} bowl to plate (viewpoint), cream cheese into bowl (sensor noise), wine bottle to rack (lighting) and stove on (background texture).}
  \label{fig:sim-benchmarks}
\end{figure*}

The Chameleon-7B trunk~\citep{chameleon,luminamgpt} (the backbone in Fig.~\ref{fig:arch}) processes frame groups in causal order (Fig.~\ref{fig:attn-mask}). Within each group, attention follows token order, with bidirectional attention among valid slots, which share the rotary position $p_t$ of their slot-block start marker. Padded slots are masked as keys.

Standard attention computes a slot's key from its full hidden state, which mixes the address with the changing state; object-addressable attention (OA) computes it from the address alone. OA reserves a fixed set of residual coordinates $\mathcal A$ for each slot's identity address, writes the address before the first layer and restores it after every layer. At layer $m$, RMS normalization over $\mathcal A$ alone gives $\hat{\mathbf a}_k^{(m)}$, which keeps changes in the object's state from altering the scale of the address. The slot key reuses the corresponding columns of the pretrained key projection:
\begin{equation}
\mathbf K_k^{(m)}=\mathrm{RoPE}_{p_t}\!\left(
\mathrm{LN}_K\!\left(W_{K,\mathcal A}^{(m)}\hat{\mathbf a}_k^{(m)}\right)\right).
\label{eq:oa}
\end{equation}
This construction adds no attention parameters, and at a fixed frame position a change in appearance or pose leaves the slot key unchanged: observations reach the address only through the first-frame descriptor, never through the pose or later frames. Queries and values use the full hidden state: the query can shift attention with the instruction or scene, and the value carries the object's current state. Non-slot tokens retain the original attention rule. One trunk pass yields the current object-slot states $\mathbf H_{k,t}$ and the action-query state $\mathbf h_{\mathrm q}$ (Supp.~\ref{app:oa-impl}).

\subsection{Prediction heads}
\label{sec:method:heads}

\noindent\textbf{Object-conditioned actions.} The action head selects objects through the same addresses. It projects $\mathbf h_{\mathrm q}$ into $R$ role queries $\mathbf q_r$; for each valid object $k\in\mathcal V_t$, the pointer projects its identity address to a key $\boldsymbol\kappa_k$ and its current state $\mathbf H_{k,t}$ to a value $\mathbf v_{k,t}$. Each role weights the objects, and the weighted states form the readout:
\begin{equation}
\begin{aligned}
\boldsymbol\alpha_r
&=\operatorname{softmax}_{k\in\mathcal V_t}
  \!\left(\frac{\mathbf q_r^\top\boldsymbol\kappa_k}{\sqrt d}\right),\\
\mathbf r_t
&=\operatorname{LN}\!\Bigl(
  \bigl[\sum\nolimits_{k\in\mathcal V_t}\alpha_{rk}\mathbf v_{k,t}\bigr]_{r=1}^{R}
  \Bigr),
\end{aligned}
\label{eq:pointer}
\end{equation}
where $d$ is the key dimension. The action condition $\mathbf c_t=[W_c\mathbf h_{\mathrm q};\mathbf r_t]$ combines the task context with the states selected by the pointer. A conditional flow head~\citep{flowmatching,pi0} learns the velocity from Gaussian noise to the demonstrated chunk and, at inference, integrates it with $\mathbf c_t$ fixed to update all $H$ actions jointly (Supp.~\ref{app:heads}).

\noindent\textbf{Future object states and joint supervision.} The world head predicts each object's next content features and pose, $(\hat{\mathbf c}_k^{t+1},\hat{\mathbf p}_k^{t+1})=h_\psi(\mathbf H_{k,t})$, and its loss trains the slot representations shared with the action head. An exponential moving average (EMA) copy of $f_{\mathrm{cnt}}$ encodes the content target from the next observation. Content, position and rotation have separate validity masks, so an observed object can supply a content target even when its geometry is missing. During training, an auxiliary image head predicts the next decision's VQ codes from the current image-token states. The policy objective is
\begin{equation}
\begin{aligned}
\mathcal L(\theta)
={}&\mathcal L_{\mathrm{act}}
  +\lambda_w\mathcal L_{\mathrm{world}}
  +\lambda_v\mathcal L_{\mathrm{vq}}\\
 &+\lambda_c\mathcal L_{\mathrm{compose}}
  +\lambda_r\mathcal L_{\mathrm{role}}.
\end{aligned}
\label{eq:loss}
\end{equation}
The composition loss encourages consistent object selection and actions across identity-code redraws and distractor insertions by aligning pointer distributions over the original objects and action chunks generated from the same noise. The role loss supervises target and reference selection with pseudo-labels parsed from the instruction (Supp.~\ref{app:loss-detail}).

\noindent\textbf{Closed-loop execution.}\phantomsection\label{sec:method:inference}
Before taking the first $K$ actions, the policy caches each object's forecast, which serves as a check on the tracked state at the next query. With this check enabled, a missing named object or a residual above threshold triggers re-grounding (re-detection and reassignment) before the pointer reads the updated states (Supp.~\ref{app:reassignment}).

\begin{figure*}[t]
  \centering
  \includegraphics[width=\textwidth]{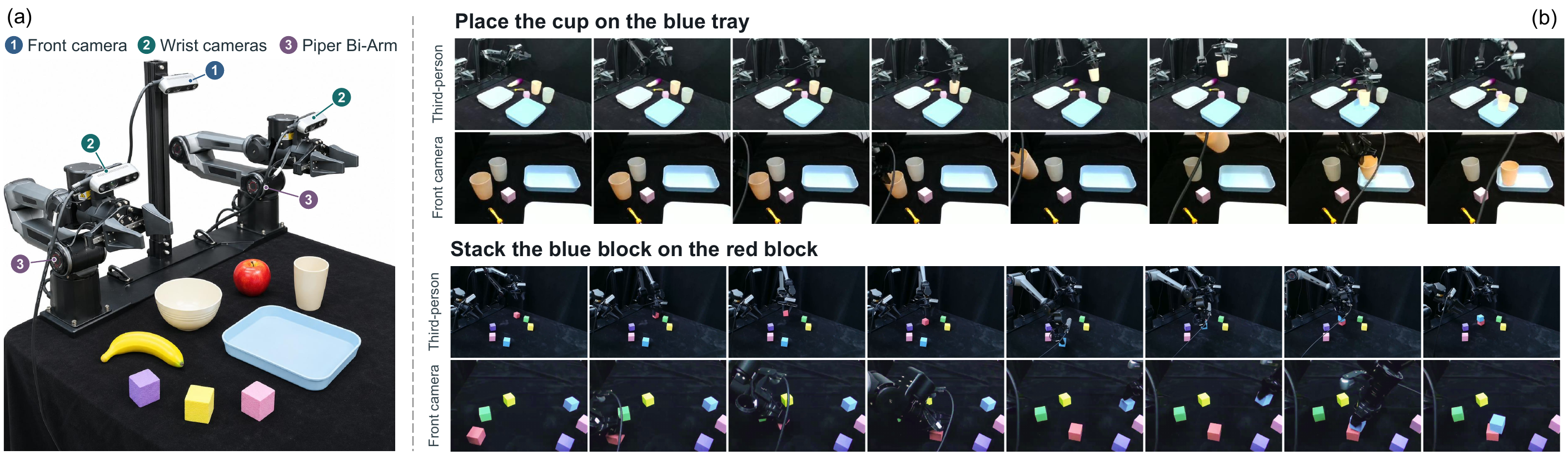}
  \caption{\textbf{Real-world setup and closed-loop \method{} rollouts.}
    (a) Two PiPER arms with a fixed front camera and one camera on each wrist. (b) \method{} places the cup on the blue tray (T1) and stacks the blue block on the red block (T3). For each task, the upper row shows a third-person view recorded only for visualization, not a policy input, and the lower row shows the front camera; time runs from left to right.}
  \label{fig:real-platform}
\end{figure*}

\subsection{Training setup}
\label{sec:method:training}

Stage~0 fine-tunes the full Chameleon-7B trunk on DROID~\citep{droid} and 100,000 Open X-Embodiment trajectories~\citep{openx}, with causal attention and cross-entropy only on valid future-action tokens. Stage~I freezes the trunk and trains the slot modules and world head with OA enabled on the same data plus LIBERO. Stage~II adds the pointer-conditioned action head and rank-$32$ LoRA~\citep{hu2022lora} adapters, and optimizes Eq.~\eqref{eq:loss} on standard LIBERO demonstrations, or on BridgeData~V2~\citep{bridgev2} for WidowX. Slot-based models are trained with route dropout (Fig.~\ref{fig:attn-mask}; Supp.~\ref{app:training-detail}).

\section{Experiments}
\label{sec:experiments}

\subsection{Simulation Benchmarks}
\label{sec:experiments:benchmarks}

\noindent\textbf{Setup and baselines.}
We fine-tune \method{} on LIBERO~\citep{libero} for four suites and on BridgeData V2~\citep{bridgev2} for four SimplerEnv WidowX visual-matching tasks~\citep{simplerenv} (Fig.~\ref{fig:sim-benchmarks}; Supp.~\ref{app:sim-deployments}). On all three benchmarks, \method{} estimates geometry from RGB without simulator depth. We compare three-seed means of \method{} with the published results of ten baselines (five VLAs, five WAMs), whose protocols can differ from ours (Supp.~\ref{app:baseline-sources}). LIBERO-Plus tests the LIBERO policies zero-shot on seven perturbation axes~\citep{liberoplus}.

\noindent\textbf{Standard manipulation.}
\method{} leads both standard benchmarks (Tab.~\ref{tab:libero-simpler}): $99.0\%$ on LIBERO and $88.9\%$ on WidowX (256/288 trials), where it is highest on all four tasks and exceeds the best of seven baselines by $8.1$ points.

\noindent\textbf{Zero-shot robustness.}
\method{} leads the published baselines on five axes and Overall (90.5\%); its Geo Avg exceeds the best published baseline by 9.8 points (Tab.~\ref{tab:libero-plus}). On the same 10,030 conditions (Supp.~\ref{app:baseline-artifacts}), it outperforms the public $\pi_{0.5}$ checkpoint on all seven axes (Geo Avg $+9.6$, Overall $+6.1$ points) and Cosmos Policy on six (Geo Avg $+15.6$, Overall $+8.5$); Cosmos Policy is higher only under sensor noise (92.4\% vs.\ 85.8\%).

\begin{table}[!htb]
\centering
\caption{\textbf{Nominal task success on PiPER.} All methods are fine-tuned on the same PiPER demonstrations. Entries give successes out of 20 trials; Avg is the five-task mean (\%). Wilson 95\% intervals for \method{} are in Supp.~\ref{app:real}. \textbf{Bold} and \underline{underlining} mark the best and second-best scores in each column, including ties. Bimanual OpenVLA-OFT uses FiLM~\citep{perez2018film}.}
\label{tab:real}
\small
\setlength{\tabcolsep}{2pt}
\renewcommand{\arraystretch}{1.02}
\begin{tabular*}{\linewidth}{@{}l@{\extracolsep{\fill}}cccccc@{}}
\toprule
 & \multicolumn{2}{c}{Single arm} & \multicolumn{3}{c}{Bimanual} & \\
\cmidrule(lr){2-3}\cmidrule(lr){4-6}
Method & T1 & T2 & T3 & T4 & T5 & Avg\\
\midrule
$\pi_{0.5}$ & 14/20 & 9/20 & 13/20 & 12/20 & 7/20 & 55 \\
OpenVLA-OFT & 15/20 & 8/20 & 10/20 & 7/20 & 4/20 & 44 \\
GR00T N1.5 & 16/20 & \underline{18/20} & 15/20 & 8/20 & 6/20 & 63 \\
Fast-WAM & \underline{17/20} & 17/20 & \underline{16/20} & \underline{14/20} & 9/20 & \underline{73} \\
Cosmos Policy & \underline{17/20} & 16/20 & 15/20 & 12/20 & \underline{11/20} & 71 \\
\textbf{\method{}} & \textbf{19/20} & \textbf{20/20} & \textbf{17/20} & \textbf{18/20} & \textbf{15/20} & \textbf{89} \\
\bottomrule
\end{tabular*}
\end{table}

\subsection{Real-Robot Experiments}
\label{sec:experiments:real}

\noindent\textbf{Platform and tasks.}
On AgileX PiPER (Fig.~\ref{fig:real-platform}), one arm places a cup on a tray, both specified by color (T1), or selects one of three identical cups by initial position (T2). Two arms stack blocks (T3), sort fruit and vegetables (T4) or relay a named object from clutter to a tray (T5). We fine-tune \method{} and five of the baselines (three VLAs, two WAMs) once per arm configuration on the same 20 teleoperated RGB demonstrations per task (Supp.~\ref{app:real-training}).

\noindent\textbf{Nominal and perturbed scenes.}
On the same held-out scene--instruction pairs (20 trials per task), \method{} succeeds in 89 of 100 nominal trials against 73 for Fast-WAM, the strongest baseline, and leads on every task (Tab.~\ref{tab:real}). Under matched single-arm perturbations, evaluated against $\pi_{0.5}$ and Fast-WAM, \method{} leads in every condition and has the smallest drop from nominal success in each, tied with Fast-WAM under occlusion (Tab.~\ref{tab:real-perturbations}). With a lookalike cup, it keeps 19/20 on T1, while Fast-WAM falls to 9/20.

\FloatBarrier

\subsection{Ablation Studies}
\label{sec:experiments:ablations}

\noindent\textbf{Addressing, not slots, drives robustness.}\phantomsection\label{sec:experiments:controls}
Model (a) is an object-slot WAM with our perception, pointer and world head; (b) differs from it only in how the trunk addresses the slots (Tab.~\ref{tab:controls}A). With standard attention, object slots reach 78.6\% mean Geo Avg, below the slot-free policy (h) at 79.7\%. Addressing the same slots by identity raises Geo Avg to 89.3\% and improves every LIBERO suite and LIBERO-Plus axis in all three seeds, with mean gains of 13.3 points on Camera and 12.4 under sensor noise (Tab.~\ref{tab:per-seed}). With only the external camera in Stage~II and at evaluation, OA raises Geo Avg from 78.8\% (a$'$) to 82.3\% (b$'$).

Trunk identity keys improve target choice itself. On paired trials, OA raises the pointer's hit rate on the instructed instance by 13.2, 9.3 and 4.6 points on Camera, Robot init and Layout and lowers wrong first grasps on every axis. On the 935 of 1,084 pairs where both policies bind the target correctly at the start, the gains change by less than 0.1 points and stay positive in every seed (Tab.~\ref{tab:controls}C). Both trunk operations contribute: with Stage~I shared, Geo Avg is 84.5\% with both off, 86.7\% with per-layer writes alone, 86.9\% with the address key path alone and 89.3\% with both (Tab.~\ref{tab:abl-train}). Route dropout (Fig.~\ref{fig:attn-mask}) raises Geo Avg from 87.4\% to 89.3\%, and with the image route closed at test time, the policy retains 88.5\% (Tabs.~\ref{tab:abl-train}, \ref{tab:abl-infer}).

\noindent\textbf{The identity-key pointer follows a new target.}\phantomsection\label{sec:experiments:retarget}
On LIBERO-Object, we change the requested object but keep the scene and basket fixed (Tab.~\ref{tab:controls}B). Without role supervision, replacing mean pooling with the identity-key pointer raises new-target placement from 56.9\% to 64.6\%, and role supervision to 77.4\%, against 69.8\% for an equally supervised state-key pointer. In every seed, more trials end with only the new target in the basket (65.9\% vs.\ 60.3\%; Tab.~\ref{tab:readout-seeds}).

\begin{table}[!t]
\centering
\caption{\textbf{Object addressing and target selection.} A: (a)/(b) share perception, readout and training budget. A/B: three-seed means (\%); Cosmos Policy uses one public checkpoint. B: 500 rewrites per checkpoint. Re-target includes trials with both objects in the basket; New only excludes them; Orig.\ only counts trials with only the original object in it. C: mean within-seed (b)$-$(a) in pp. Both correct: both initial target bindings are correct (Tab.~\ref{tab:per-seed}B--D). Bold/underline: best/next distinct within view (A) or panel (B).}
\label{tab:controls}\label{tab:retarget}
\footnotesize
\setlength{\tabcolsep}{2pt}
\renewcommand{\arraystretch}{1.06}
\begin{tabularx}{\linewidth}{@{}l*{3}{Y}@{}}
\toprule
\multicolumn{4}{@{}l}{\textbf{A. Addressing under geometric perturbations}}\\
Model & LIBERO $\uparrow$ & Camera $\uparrow$ & Geo Avg $\uparrow$\\
\midrule
(h) No object slots & \underline{97.4} & 76.0 & \underline{79.7}\\
(a) Object slots, no OA & \underline{97.4} & \underline{76.6} & 78.6 \\
(b) \method{} & \textbf{99.0} & \textbf{89.9} & \textbf{89.3} \\
\midrule
\multicolumn{4}{@{}l}{\emph{External camera only}}\\
(a$'$) Object slots, no OA & \underline{97.7} & \underline{73.6} & \underline{78.8} \\
(b$'$) \method{} & \textbf{97.8} & \textbf{76.4} & \textbf{82.3} \\
\bottomrule
\end{tabularx}\par\smallskip
\begin{tabular*}{\linewidth}{@{}l@{\extracolsep{\fill}}cccc@{}}
\multicolumn{5}{@{}l}{\textbf{B. Instruction changes; scene stays fixed}}\\
Readout & Role loss & Re-target $\uparrow$ & New only $\uparrow$ & Orig.\ only $\downarrow$\\
\midrule
Mean pooling & Off & 56.9 & --- & 29.1 \\
Identity-key pointer & Off & 64.6 & --- & 20.3 \\
State-key pointer & On & \underline{69.8} & \underline{60.3} & \underline{15.2} \\
Identity-key pointer & On & \textbf{77.4} & \textbf{65.9} & \textbf{11.3} \\
\midrule
Cosmos Policy (public) & --- & 53.4 & --- & 30.6 \\
\bottomrule
\end{tabular*}\par\smallskip
\begin{tabularx}{\linewidth}{@{}l*{3}{Y}@{}}
\toprule
\multicolumn{4}{@{}l}{\textbf{C. Target choice and first grasp on paired trials}}\\
Axis & \shortstack{Pointer hit\\all pairs} & \shortstack{Pointer hit\\both correct} & \shortstack{Wrong first grasp\\all pairs}\\
\midrule
Camera & $+13.22$ & $+13.15$ & $-2.24$ \\
Robot init & $+9.30$ & $+9.29$ & $-4.51$ \\
Layout & $+4.57$ & $+4.63$ & $-8.72$ \\
\bottomrule\end{tabularx}
\end{table}

\noindent\textbf{Object forecasts help training and recovery.}
With recovery off and a shared Stage-I initialization, Geo Avg is 80.7\% for image-code targets, 83.8\% for object-state targets and 84.3\% for both (Tab.~\ref{tab:world-ablation}A; Supp.~\ref{app:recovery-controls}). At fixed checkpoints, re-grounding on disappearance raises it to 87.5\%. A trigger for visible objects then raises Geo Avg to 88.7\% with observation change and 89.3\% with the forecast residual. The residual outperforms observation change on all three axes in every seed and intervenes in 11.3\% of trials, versus 9.2\%.

\phantomsection\label{sec:experiments:mechanism}The forecast residual detects more injected target--distractor swaps than observation change (76.2\% vs.\ 61.7\%) with fewer clean interventions per episode (0.051 vs.\ 0.082). After a swap, task success reaches 68.0\%, versus 64.7\% with observation change and 56.3\% without re-grounding (Tabs.~\ref{tab:world-ablation}B, \ref{tab:abl-monitor}).

\begin{table}[!t]
\centering
\caption{\textbf{Future-state supervision and recovery.} Three-seed means (\%). A: With both targets, only the trigger varies within a seed; change compares successive observations (Supp.~\ref{app:recovery-controls}). Dash: not evaluated. B: 400 swap and 400 clean trials per rule and seed. Bold/underline: best/next distinct success per column and panel.}
\label{tab:world-ablation}
\footnotesize
\setlength{\tabcolsep}{2pt}
\renewcommand{\arraystretch}{1.06}
\begin{tabular*}{\linewidth}{@{}l@{\extracolsep{\fill}}lrr@{}}
\toprule
\multicolumn{4}{@{}l}{\textbf{A. Forecast targets and test-time recovery}}\\
Forecast targets & Recovery trigger & LIBERO $\uparrow$ & Geo Avg $\uparrow$\\
\midrule
Image codes & Off & 98.0 & 80.7 \\
Object states & Off & --- & 83.8 \\
Both & Off & \underline{98.8} & 84.3 \\
Both & Disappearance only & --- & 87.5 \\
Both & Disappearance or change & --- & \underline{88.7} \\
Both & Disappearance or residual & \textbf{99.0} & \textbf{89.3} \\
\bottomrule
\end{tabular*}\par\smallskip
\begin{tabularx}{\linewidth}{@{}l*{3}{Y}@{}}
\multicolumn{4}{@{}l}{\textbf{B. Injected track swaps; checkpoint fixed within seed}}\\
Recovery rule & Swap $\uparrow$ & Clean $\uparrow$ & Clean events/ep.\\
\midrule
No re-grounding & 56.3 & \textbf{98.1} & 0 (fixed) \\
Observation change & \underline{64.7} & 97.4 & 0.082 \\
Forecast residual & \textbf{68.0} & \underline{97.5} & 0.051 \\
\bottomrule
\end{tabularx}
\end{table}

\section{Conclusion}

\method{} finds each object by an episode-fixed identity address and reads its changing state. Keying slots by this address in every trunk layer makes the policy robust to geometric change, whereas the same slots under standard attention do no better than a slot-free policy: robustness depends on how a policy addresses objects, not only on how it represents them. From the same correct binding, trunk identity keys select the instructed object more often. The same address keys the action pointer, which lets the policy follow a newly requested object, and indexes per-object forecasts, which train a more robust policy than image-code forecasts and detect most injected track swaps. \method{} achieves the highest overall success on LIBERO-Plus, SimplerEnv WidowX and five real-robot tasks. Together, these results favor identity-addressed objects as the unit of attention and prediction in world action models.

\paragraph{Limitations and future work.} \method{} relies on binding each address to the right instance in the first frame. Swapping the initial bindings of two instances lowers same-class distractor success from 80.0\% to 54.3\% (Tab.~\ref{tab:abl-infer}), and a binding that is wrong from the start can yield a self-consistent track with a small residual. Poses from the external camera also assume nominal extrinsics: moving it lowers PiPER success by 20 and 30 points on T1 and T2 (Tab.~\ref{tab:real-perturbations}). These limits point to concrete next steps. Online extrinsic estimation is one: a wrist-anchored offset estimate already raises simulated Camera success from 89.9\% to 92.6\% (Tab.~\ref{tab:abl-infer}), and visual servoing~\citep{hager1997} or markerless camera-to-robot estimation~\citep{ctrnet,robopepp} could bring it to hardware. Further directions include re-grounding after missed first-frame detections, a composition objective for distractors that should change the action, other trunk sizes, and faster, more accurate perception: the frozen perception stack accounts for most of the per-query latency (Supp.~\ref{app:latency}), and simulator poses raise Geo Avg from 89.3\% to 93.3\% (Tab.~\ref{tab:abl-train}).

\ifdefined\ArxivEdition\else\clearpage\fi
\LLMUsageStatement
{\small
\bibliographystyle{ieeenat_fullname}
\bibliography{references}
}

\onecolumn
\newgeometry{onecolumn,layoutwidth=8.5in,layoutheight=11in,textwidth=6.875in,textheight=8.875in,centering}
\setlength{\paperwidth}{8.5in}
\setlength{\paperheight}{11in}
\pdfpagewidth=\paperwidth
\pdfpageheight=\paperheight
\pdfbookmark[0]{Supplementary material}{oawam-supplement}
\begin{center}
  {\Large\bfseries \PaperTitle\par}
  \vspace{.5em}
  {\large Supplementary Material\par}
\end{center}
\setcounter{figure}{0}
\setcounter{table}{0}
\setcounter{equation}{0}
\setcounter{algorithm}{0}
\renewcommand{\thefigure}{S\arabic{figure}}
\renewcommand{\thetable}{S\arabic{table}}
\renewcommand{\theequation}{S\arabic{equation}}
\renewcommand{\thealgorithm}{S\arabic{algorithm}}
\renewcommand*{\theHfigure}{supp.\arabic{figure}}
\renewcommand*{\theHtable}{supp.\arabic{table}}
\renewcommand*{\theHequation}{supp.\arabic{equation}}
\providecommand*{\theHalgorithm}{\arabic{algorithm}}
\renewcommand*{\theHalgorithm}{supp.\arabic{algorithm}}
\appendix
\makeatletter
\setlength{\@fptop}{0pt}
\setlength{\@fpbot}{0pt plus 1fil}
\makeatother
\raggedbottom
\section{From observations to object sequences}\label{app:tokenizer-detail}

\subsection{Observations at a decision time}

At control step $s$, the simulation policy receives the instruction, RGB observations and proprioception up to $s$, together with previously executed actions. LIBERO supplies scene and wrist images at $20$\,Hz, seven-dimensional end-effector states and actions. Its four suites contain ten tasks each, with 50 released demonstrations per task (training and validation splits in Supp.~\ref{app:training-detail}). BridgeData V2~\citep{bridgev2} provides third-person RGB training sequences at $5$\,Hz for WidowX. During evaluation, the policy receives online observations from the corresponding third-person camera in SimplerEnv. The dimensions in Supp.~\ref{app:tokenizer-detail}--\ref{app:inference-detail} refer to simulation unless otherwise stated. The PiPER inputs and commands are in Supp.~\ref{app:real}.

In the default simulation settings, the policy predicts $H=16$ actions and executes a prefix of length $K\le H$. For a query at control step $s$, the $T=4$ history observations have indices $\max(0,s-jK)$ for $j=3,2,1,0$, and the next-decision target has index $s+K$. LIBERO uses $K=16$ at $20$\,Hz; BridgeData~V2 and SimplerEnv use $K=4$ at $5$\,Hz. In both settings the executed prefix spans a nominal $0.8$\,s. Thus, BridgeData~V2 training supervises all $16$ predicted actions while its world and image targets describe the state after the first $4$ actions. Each completed history group contains the first action executed after its observation. Left-padded history positions and missing next-decision observations are excluded from world and image losses. Action chunks are right-padded with the terminal action, and padded steps are excluded from action losses. Stage-I sampling uses the timestamps and query stride of each source.

For the simulation policies, the six continuous action dimensions are clipped to the training split's $1$st/$99$th percentiles and scaled to $[-1,1]$; the gripper is mapped to $\{-1,+1\}$. Stage~0 quantizes each action coordinate into $256$ bins. Stages~I and II reuse this per-dataset mapping for historical actions. The normalized $H=16$ future actions form the Stage-II flow target. Proprioception is discretized into $256$ bins per coordinate; normalized bin centers give $\bar{\mathbf q}$ for the auxiliary monitor. The calibrated workspace box maps object positions into $[-1,1]^3$, and rotations use the six-dimensional representation of~\citep{zhou2019continuity}. All normalization statistics and workspace bounds are fixed from training data and reused at deployment.

\subsection{Grounding, geometry and object state}

At the first observation, Qwen3-VL-4B~\citep{qwen3vl} parses the instruction into object phrases and target--reference relations. Each phrase prompts SAM~3~\citep{sam3}; an additional pass finds unmentioned distractors. Masks are deduplicated when both IoU exceeds $0.5$ and DINOv3 cosine similarity exceeds $0.9$. The label and the initial mask-pooled descriptor define the linguistic and visual parts of the identity address. SAM~3 tracks the mask at subsequent decisions; appearance and pose at those decisions update the slot content while the original address stays fixed.

\paragraph{Geometry from RGB.}\phantomsection\label{app:monocular-depth}
The image tokenizer resizes RGB to $256\times256$ and scales it to $[0,1]$. Geometry estimation keeps the original aspect ratio and transforms the image, masks, depth and intrinsics identically. The RGB benchmark configurations use frozen MoGe-3 ViT-L~\citep{moge3} (\texttt{Ruicheng/moge-3-vitl}), with resolution level 9, three refinement steps and projection-consistent point maps. A calibrated horizontal field of view is supplied when available; otherwise, the estimator predicts normalized intrinsics. Only positive, finite depths inside both the predicted validity mask and the object mask contribute. Simulated depth is excluded from the RGB benchmark inputs, including offline features. Stage-I sources with sensor depth can use their recorded measurements.

Let $\mathbf K^{\rm px}$ be the intrinsics for the pixel grid on which depth and masks are aligned. A depth sample $d(u,v)$ is mapped into the robot base frame by
\[
\mathbf X^{\rm cam}=d(u,v)(\mathbf K^{\rm px})^{-1}
  \begin{bmatrix}u&v&1\end{bmatrix}^{\!\top},
\qquad
\mathbf X^{\rm base}=\mathbf R_{\rm base\leftarrow cam}\mathbf X^{\rm cam}
  +\mathbf t_{\rm base\leftarrow cam}.
\]
On the returned image grid, $\mathbf K^{\rm px}=\operatorname{diag}(W_{\rm im},H_{\rm im},1)\mathbf K^{\rm norm}$. Camera-to-base transforms come from the recorded calibration or from robot kinematics and hand--eye calibration. A frame without an aligned image transform or camera calibration supplies appearance and action targets but no base-frame geometry. Nominal training data with sensor depth or known workspace distances provide the reference for monocular scale and position uncertainty.

\paragraph{Multi-view, multi-frame pose tracking.}
\phantomsection\label{app:pose-tracking} \emph{Measurement.} For each view, the object's mask pixels with valid depth are back-projected into the robot base frame. External cameras keep their nominal, training-time extrinsics even when moved; the wrist camera uses forward kinematics and hand--eye calibration. Points farther than $3$ median absolute deviations from the median are trimmed. The median of the remaining points is the position measurement, with covariance $\mathrm{diag}(1.57\,\boldsymbol{\sigma}^2/n+\sigma_z^2)+\sigma_v^2\mathbf{I}$ from the point spread $\boldsymbol{\sigma}^2$, the number of points $n$, a depth-source-specific error scale $\sigma_z$ and a systematic error $\sigma_v$ of view $v$, since views see different sides of an object and their extrinsics never agree exactly. For multiple views, $\sigma_v$ is the RMS disagreement between the views on nominal training episodes. Single-view MoGe-3 uses scale and covariance fitted against the reference geometry described above. Views with fewer than $30$ points or unavailable camera calibration give no base-frame measurement. When the trimmed point cloud is elongated (eigenvalue ratios above $1.3$), its principal axes give an orientation measurement; without one, the rotation validity flag is zero. These axes describe visible geometry and are used as orientation pseudo-targets; symmetric or poorly observed axes receive no rotation loss.
\emph{Fusion.} Each object has a constant-velocity Kalman filter on position (process noise $0.02$\,m/s$^2$), updated with every view; the wrist view, whose extrinsics come from the robot's kinematics, is applied first. A measurement inside the $99\%$ Mahalanobis gate ($\chi^2_3=11.34$) updates the filter. A measurement outside the gate is compared with the last accepted measurement of the same view. If it also jumps relative to that measurement and its mask size stays between $1/1.4$ and $1.4$ times that view's last accepted mask, as for a grasped object, the object is treated as moved and the track restarts at the measurement. If it does not jump, the views are treated as disagreeing: the measurement is skipped if the wrist view measured the object at this query and is otherwise applied with its covariance inflated by the innovation. A size change is treated as a corrupted mask (leakage onto a neighbor, partial occlusion by the arm), and the measurement is rejected; after $3$ consecutive rejections the track restarts. Without measurements (occlusion) the filter predicts. Orientation is averaged on SO(3) with weight $0.3$ and a $45^\circ$ gate, and the sign ambiguity of principal axes is resolved by choosing the proper sign flip closest to the current estimate. Tracks are reset at episode start and for re-grounded objects. The tracker uses elapsed timestamps; the nominal query interval is $\Delta t=0.8$\,s for LIBERO and SimplerEnv. In the wrist-anchored offset diagnostic (P7 in Tab.~\ref{tab:abl-infer}), we estimate the external-camera position offset relative to the wrist view online as the running median of their position differences over jointly observed objects, and subtract it from the external-camera measurements. This sensor-based calibration correction is related to camera-to-robot estimation~\citep{ctrnet,robopepp} and is enabled only for this diagnostic.

\paragraph{Temporal filter.}
For each object the filter keeps a state $\tilde{\mathbf{x}}$ and updates $\tilde{\mathbf{x}}_t=\tilde{\mathbf{x}}_{t-1}+w_t(\mathbf{x}_t-\tilde{\mathbf{x}}_{t-1})$ with $w_t=\alpha\,\mathrm{clip}(\gamma/d_t,w_{\min},1)$, $\alpha=0.6$, $w_{\min}=0.15$. Here $d_t$ is the cosine distance for the DINOv3 descriptor ($\gamma=0.15$) and the L2 distance for mask shape ($\gamma=0.25$). Poses are not filtered again; they come from the Kalman tracker. Before mask pooling, soft masks are averaged with weight $\alpha$ when their IoU with the filtered mask is at least $0.5$, and with weight $\alpha w_{\min}$ otherwise. Objects seen for the first time start from the raw observation, invisible objects keep their state, and the state is reset at the start of each episode. The filter is causal and has no learned parameters; it uses the same settings on every benchmark and axis, chosen on unperturbed LIBERO validation episodes whose initial states are not test states.

\paragraph{Slot features.}\phantomsection\label{app:slot-features}
A fixed projection maps the mask-pooled DINOv3 ViT-L/16 feature~\citep{dinov3} to 256 dimensions. The label is a 256-dimensional average T5 phrase encoding~\citep{t5}; automatic detections receive a learned unknown label. The raw vector concatenates the visual descriptor (256), pose (3 position and 6 rotation), label (256), image-plane mask shape (15), validity flags for visibility, position and rotation ($v^{\rm vis},v^{\rm pos},v^{\rm rot}$) and one zero padding entry, for 540 entries. Shape encodes the centroid, bounding box, second moments, area and convexity; it is available to the temporal filter and zeroed before content encoding. An unavailable pose component is zero-filled and masked from its regression loss. Thus, a visible symmetric object can still supply appearance supervision when its orientation is unresolved.

The encoders $f_\mathrm{addr}:\mathbb{R}^{512}\!\to\!\mathbb{R}^{128}\!\to\!\mathbb{R}^{32}$ and $f_\mathrm{cnt}:\mathbb{R}^{540}\!\to\!\mathbb{R}^{512}\!\to\!\mathbb{R}^{256}$ are two-layer MLPs with input LayerNorm and GELU. The identity-code bank holds $32$ learnable $32$-d codes initialized orthonormally. Code $0$ is reserved for the robot. At the start of every episode (every sample during training), the valid objects receive distinct codes drawn uniformly without replacement from codes $1$--$31$. Addresses of padded slots are zero. The slot adapter $f_\phi(\mathbf{s})=W_2\,\mathrm{GELU}(W_1\,\mathrm{LN}(\mathbf{s}))$ maps $\mathbb{R}^{320}\!\to\!\mathbb{R}^{4096}\!\to\!\mathbb{R}^{4096}$. $\boldsymbol{\pi}^t$ is a $16$-d sinusoidal embedding of the frame index and $\boldsymbol{\rho}_k$ a $16$-d learned role embedding.

\subsection{Input paths and causal sequence}\label{app:sequence}

\begin{center}\small
\begin{tabularx}{\linewidth}{@{}l l X@{}}\toprule
Path & Source & Representation and destination\\\midrule
Parser & Instruction and first image & Object phrases prompt grounding; relations supply role labels. Neither enters the trunk as parser output.\\
Text & Instruction & Chameleon BPE tokens embedded by the trunk~\citep{chameleon}.\\
Image & Scene RGB & Frozen VQ-GAN: $16\times16$ codes from a vocabulary of 8,192. The slot-free control encodes all available views.\\
Slot & Tracked objects & Appearance from available views, pose and identity form one embedding per slot.\\
State & Proprioception & Seven discretized state tokens in each observation's own frame group.\\
Action & Past actions & Seven discretized tokens for the first executed action in each completed group.\\\bottomrule
\end{tabularx}\end{center}

The complete sequence template is shown below, with frame groups indexed locally by $t=0,\ldots,T-1$:
\[
\bigl[\tok{bos};\,\mathbf{x}^{\text{text}};\,\bigl[\tok{f-bos};\,\mathbf{x}_t^{\text{image}};\,\mathbf{x}_t^{\text{state}};\,\tok{s-bos};\,\mathbf{x}_t^{\text{slot}};\,\tok{s-eos};\,\mathbf{x}_t^{\text{action}};\,\tok{f-eos}\bigr]_{t=0}^{T-1};\,\tok{[act\_q]}\bigr].
\]

Each group has 256 image codes, seven proprioception tokens and up to 17 slots: the robot at index zero and up to 16 objects. Padding cannot serve as an attention key. The current group has no executed-action tokens. With padded slots, the four groups contain 1,157 tokens. Six reserved symbols delimit frames and slot blocks, mark slot-embedding positions and supply the action query, without changing vocabulary size. State, action, image and structural symbols occupy disjoint token IDs.

The mask is causal across frame groups. Within each group, queries follow the token order $\mathrm{VQ};\mathrm{STATE};\mathrm{SLOT}$, with executed-action tokens appended only in historical groups. Valid slots form a bidirectional block, and padded slots are masked as keys. A historical action token can therefore read its frame's observations and slots. The final \tok{[act\_q]} can attend to all preceding valid tokens. During Stage~II, slot-based models jointly mask all image-code columns of this query in a random half of the samples (route dropout). The route is open at inference; the closed-route evaluation (K5 in Tab.~\ref{tab:abl-infer}) closes it at every query.

For cached training inputs, each stride phase $p=0,\ldots,K-1$ starts from the first episode observation and advances through $p,p+K,\ldots$. Mask tracking, feature pooling, pose estimation and filtering use only observations available by that query. The resulting batch has $T=4$ history groups and $S=17$ slot positions per group, with raw features of shape $B\times T\times S\times540$. Each group carries its own proprioception; subsequent observations enter only as targets for earlier groups. At deployment the same state updates are applied online.

\FloatBarrier
\section{Object addressing and its guarantees}\label{app:oa-impl}

\subsection{Address writes and slot keys}

The trunk follows the Lumina-mGPT-7B-768 architecture~\citep{luminamgpt}: $32$ pre-norm layers, width $4096$, $32$ heads without GQA, SwiGLU FFN of size $11{,}008$, vocabulary $65{,}536$, RMSNorm, per-head Q/K LayerNorm and RoPE with $\theta\!=\!10{,}000$ ($7.0$B parameters). Stage~0 starts from the public Chameleon-7B weights~\citep{chameleon} and trains the full trunk without OA (Supp.~\ref{app:training-detail}). Each layer's attention is replaced by a slot-aware operation that reuses the pretrained projections and Q/K LayerNorms; without slot rows, it reproduces the original attention exactly. Slot embeddings replace the \tok{slot} positions of the embedded sequence~\citep{llava,idefics2}. Attention uses PyTorch scaled-dot-product attention with an additive mask, because the bidirectional slot blocks cannot be expressed in FlashAttention-2's causal mode.

Before the first layer and after each layer, every slot receives its identity address in the same 32 residual coordinates $\mathcal A$. To select these coordinates on the Stage-0 trunk, we compute the $99.9$th percentile of absolute activation over robot-training tokens for each layer and coordinate. We exclude coordinates exceeding ten times the layer median in any layer~\citep{massiveact} and keep the 32 remaining coordinates with the lowest mean percentile across layers. All variants with 32-dimensional address writes use this same set of coordinates. The write passes gradients to the address encoder and identity codes but not to the residual coordinates it overwrites.

\paragraph{Slot key path.}
The address normalization in Eq.~\eqref{eq:oa} is
\[
\hat{\mathbf{a}}_k^{(m)}
=\mathrm{RMSNorm}_{\mathcal{A}}^{(m)}(\mathbf{addr}_k)
=\mathbf{g}^{(m)}_{\mathcal{A}}\odot
\frac{\mathbf{addr}_k}{\sqrt{|\mathcal{A}|^{-1}\sum_{d=1}^{|\mathcal{A}|}\mathbf{addr}_{k,d}^{2}+\epsilon}},
\]
where $\mathbf{g}^{(m)}_{\mathcal{A}}$ and $\epsilon$ are the gains at coordinates $\mathcal{A}$ and the numerical constant of the layer's input RMSNorm. Let $P_{\mathcal{A}}$ place the normalized address at the coordinates $\mathcal{A}$ of an otherwise zero vector of width $4096$. The implementation applies the original key projection to this vector, which is equivalent to selecting its address-coordinate columns:
\[
W_K^{(m)}\,P_{\mathcal{A}}(\hat{\mathbf{a}}_k^{(m)})
=W_{K,\mathcal{A}}^{(m)}\hat{\mathbf{a}}_k^{(m)}.
\]
Per-head $\mathrm{LN}_K$ and RoPE are then applied as in Eq.~\eqref{eq:oa}. For the query and value of a slot, let $\mathbf h_k^{(m)}=\mathrm{RMSNorm}^{(m)}(\mathbf x_k^{(m)})$ denote its full normalized residual. The pretrained projections give
\[
\begin{aligned}
\mathbf Q_k^{(m)}&=\mathrm{RoPE}_{p_t}\!\left(\mathrm{LN}_Q\!\left(W_Q^{(m)}\mathbf h_k^{(m)}\right)\right),\\
\mathbf V_k^{(m)}&=W_V^{(m)}\mathbf h_k^{(m)}.
\end{aligned}
\]
Here $\mathrm{LN}_Q$ and $\mathrm{LN}_K$ are per-head LayerNorms. Non-slot tokens retain all pretrained projections and their original positions. The address writes and key-input construction introduce no additional parameters.

\paragraph{Masked versus exact key input.}
The masked normalized hidden state, $P_{\mathcal{A}}(\mathbf{g}^{(m)}_{\mathcal{A}}\odot\mathbf{x}^{(m)}_{k,\mathcal{A}}/\mathrm{rms}(\mathbf{x}^{(m)}_k))$, equals the exact key input scaled by $\mathrm{rms}(\mathbf{x}^{(m)}_{k,\mathcal{A}})/\mathrm{rms}(\mathbf{x}^{(m)}_k)$, a factor that depends on the slot's content and shrinks with depth as the residual norm grows. Since $\mathrm{LN}_K$ removes a positive scale only up to its $\epsilon$, this construction is only approximately invariant, and not at all in a trunk without Q/K LayerNorm. Empirically, the exact key input has higher three-seed mean success than the masked variant on all six summary metrics in Tab.~\ref{tab:abl-train}, including Geo Avg (89.3\% vs.\ 83.7\%) and Multi-inst.\ (80.0\% vs.\ 74.5\%).

\subsection{Key invariance and its scope}\label{app:key-invariance}

\paragraph{Proposition 1.}

\emph{For fixed model parameters, the key $\mathbf{K}_k^{(m)}$ at every layer $m$ is a deterministic function of the identity address $\mathbf{addr}_k$ and frame position $p_t$. With these inputs fixed, the key is invariant to the slot's dynamic content and pose and to changes in other tokens.}

\paragraph{Proof of Proposition~1.}
The write before the first layer gives $\mathbf{x}^{(0)}_{k,\mathcal{A}}=\mathbf{addr}_k$. Likewise, the write after layer $m-1$ gives $\mathbf{x}^{(m)}_{k,\mathcal{A}}=\mathbf{addr}_k$ for every subsequent layer. The address normalization therefore depends only on $\mathbf{addr}_k$ and the fixed normalization parameters. The projection $W_{K,\mathcal{A}}^{(m)}$, per-head $\mathrm{LN}_K$ and RoPE at $p_t$ consequently produce a key determined by $\mathbf{addr}_k$ and $p_t$ alone. LoRA adapters on $W_K$ act on the same address input and preserve this property. \hfill$\square$

Consequently, a change of scene leaves all slot keys unchanged if the addresses and frame positions are preserved. A viewpoint change at initialization can alter the first visual descriptor and hence the address itself. Even with an unchanged key, a different instruction or observation can change the query, attention weights and selected object; the values still carry the current state. Numerically perturbing slot content and non-slot embeddings changes the exact-path keys by less than $10^{-5}$ (maximum absolute change) at every layer. Without per-layer writes (V3), invariance holds only at the first layer; without the dedicated key path (V1), keys change at every layer. Both variants also have lower Geo Avg than the exact design (86.9\% and 86.7\% vs.\ 89.3\%; Tab.~\ref{tab:abl-train}).

\subsection{Permuting object slots}\label{app:equivariance}

All $N{+}1$ slot tokens of frame group $G_t$ share the RoPE position $p_t$ of its marker \tok{s-bos}. This marker follows the frame's image codes and proprioception, so a slot reads its own frame at positive offsets, as every token does in causal pretraining. Other tokens keep sequential positions. A shared rotation commutes with slot permutations and introduces no slot-index dependence; it is implemented through explicit position indices and adds no parameters.

\paragraph{Proposition 2.}
Assume (i) object slots carry no slot-index embeddings; (ii) all slot tokens of a frame group share one RoPE position; (iii) the same projections, attention, feed-forward blocks and heads are applied to every object slot; (iv) the addresses (including identity codes), all history-frame slot rows and the padding mask are permuted consistently by any object-slot permutation $\pi$ that fixes the robot; (v) the model runs in evaluation mode. Then the trunk is permutation-equivariant over object slots, the world predictions and pointer distributions $\boldsymbol{\alpha}$ are permuted by $\pi$, and the readout $\mathbf{r}$, the $\textnormal{\textsc{[act\_q]}}$ state and the action chunk (for identical noise) are invariant.

\paragraph{Proof sketch of Proposition~2.}
Row-wise maps commute with $\pi$; the shared RoPE rotation commutes with row permutation; mask-aware attention computes pairwise scores, so permuting rows, columns and masks consistently permutes the attention output; the address writes commute with $\pi$ under (iv). Induction over layers gives equivariant slot states. The pointer logits are computed row-wise over objects, so $\boldsymbol{\alpha}$ is permuted and $\mathbf{r}=\sum_j\alpha_{rj}\mathbf{v}_j$ is invariant. Identity codes are drawn independently of slot order, so the model cannot use them as slot-index embeddings. \hfill$\square$

Because the equivariance is exact, a consistency loss under distractor permutation would vanish up to dropout noise. Hence $\mathcal{L}_\mathrm{compose}$ uses identity-code redraws and distractor insertions, which the architecture does not make invariant (Supp.~\ref{app:loss-detail}).

\FloatBarrier
\section{Prediction heads, supervision and training}\label{app:heads}

\subsection{Predicting actions and future object states}

\paragraph{Action head $h_\xi$.}
\emph{Conditioning.} The projection $W_c\mathbf h_{\mathrm q}\in\mathbb R^{1024}$ has $4{,}194{,}304$ parameters. The action pointer in Eq.~\eqref{eq:pointer} uses
\[
\begin{aligned}
\mathbf q_r&=\bigl[W_q\operatorname{LN}_q(\mathbf h_{\mathrm q})\bigr]_r
  \in\mathbb R^{64},\qquad r=1,\ldots,R,\\
\boldsymbol\kappa_k&=W_k\operatorname{LN}_k(\mathbf{addr}_k)
  \in\mathbb R^{64},\\
\mathbf v_{k,t}&=W_v\operatorname{LN}_v(\mathbf H_{k,t})
  \in\mathbb R^{128}.
\end{aligned}
\]
Here $R=4$, and the projections have separate input LayerNorms. The four role readouts are concatenated and passed through a final LayerNorm, giving $\mathbf r_t\in\mathbb R^{512}$. The pointer has $1{,}592{,}384$ parameters. Invalid, padded and robot slots are excluded from the pointer support. When no objects are valid, the weighted sums are set to zero before the final LayerNorm.

\emph{Flow MLP.} The conditioning vector is $\mathbf c_t=[W_c\mathbf h_{\mathrm q};\mathbf r_t]\in\mathbb R^{1536}$. The network concatenates this vector with the flattened noisy chunk and a $128$-dimensional sinusoidal embedding of $\tau$:
\[
\bigl[\mathbf c_t;\operatorname{vec}(\mathbf A_t^\tau);\operatorname{emb}(\tau)\bigr]
\in\mathbb R^{1536+112+128}.
\]
An input projection maps this vector to width $1024$. Eight residual blocks apply $\mathrm{LN}\to\mathrm{Linear}\to\mathrm{GELU}\to\mathrm{Linear}$, followed by an output projection to $\mathbb R^{16\times7}$. The flow MLP has $18{,}744{,}432$ parameters, giving $24{,}531{,}120$ parameters for the complete action head.

\paragraph{Inference-time integration.}
A single trunk pass supplies $\mathbf h_{\mathrm q}$ and $\mathbf H_{k,t}$ for both heads. The action condition $\mathbf c_t$ is computed once, and the world forecast is cached before execution. Starting from $\mathbf A_t^0\sim\mathcal N(\mathbf0,\mathbf I)$, four Euler steps produce the action chunk:
\begin{equation}
\begin{aligned}
\mathbf A_t^{\tau+\Delta\tau}
&=\mathbf A_t^\tau
  +\Delta\tau\,\mathbf v_\xi(\mathbf A_t^\tau,\tau,\mathbf c_t),\\
\Delta\tau&=\tfrac14,\qquad
\tau\in\{0,\tfrac14,\tfrac12,\tfrac34\}.
\end{aligned}
\label{eq:euler-app}
\end{equation}
The returned chunk is $\mathbf A_t=\mathbf A_t^1$; every step updates all $H\times7$ action coordinates with the same condition.

\paragraph{World head $h_\psi$.}
Separate LayerNorms and two-layer MLPs predict content and pose:
\begin{equation}
\begin{aligned}
\hat{\mathbf c}_k^{t+1}
&=\operatorname{MLP}_{4096\to1024\to256}
  \!\bigl(\operatorname{LN}_{\mathrm c}(\mathbf H_{k,t})\bigr),\\
\hat{\mathbf p}_k^{t+1}
&=\operatorname{MLP}_{4096\to256\to9}
  \!\bigl(\operatorname{LN}_{\mathrm p}(\mathbf H_{k,t})\bigr).
\end{aligned}
\label{eq:worldhead}
\end{equation}
Each MLP uses GELU between its two linear layers. The head has $5{,}525{,}257$ parameters and produces next-decision forecasts before execution. Content targets use the EMA encoder $\bar f_{\mathrm{cnt}}$ with decay $0.996$ and no target gradients. Pose targets contain the workspace-normalized position and the six-dimensional rotation representation. Supervision uses separate content, position and rotation masks for objects visible at both decision times and updates the head and trainable shared features.

\paragraph{Auxiliary motion monitor $g_\omega$.}
The optional monitor combines stored trunk features with the realized motion between decision times:
\[
\begin{aligned}
\mathbf u_t
&=f_u\!\left([\bar{\mathbf q}_t;
  \bar{\mathbf q}_{t+1}-\bar{\mathbf q}_t]\right),\\
\widetilde{\mathcal S}^{\mathrm{mon}}_{k,t+1}
&=g_\omega\!\left(\operatorname{sg}(\mathbf H_{k,t}),\mathbf u_t\right).
\end{aligned}
\]
Here $\bar{\mathbf q}$ is normalized proprioception and $\operatorname{sg}$ stops gradients. The motion encoder $f_u$ applies $\mathrm{LN}\to\mathrm{Linear}_{14\to256}\to\mathrm{GELU}\to\mathrm{Linear}_{256\to256}$. Its output is concatenated with $\mathrm{LN}(\operatorname{sg}(\mathbf H_{k,t}))$, and the resulting $4352$-dimensional input feeds content and pose branches with the world head's hidden and output widths. This auxiliary head has $5{,}922{,}597$ parameters and is evaluated after $\mathbf q_{t+1}$ is observed. Its regression loss updates only $g_\omega$ and $f_u$; Tab.~\ref{tab:abl-monitor} compares the monitor with other recovery rules.

\paragraph{Auxiliary image-VQ head.}
The pretrained vocabulary projection maps the hidden state of image code $j$ at decision time $t$ to logits for code $j$ at $t+1$. Targets within the history window occupy a later frame group, hidden from the predicting row by the block-causal mask. Targets for the latest frame lie beyond the input window. The head adds no parameters and is used during training with mean cross-entropy and weight $0.04$.

\subsection{Losses and object-level augmentations}\label{app:loss-detail}

\paragraph{Action loss and valid-step reduction.}
For a demonstrated chunk $\mathbf A_t^*$, we draw $\boldsymbol\epsilon\sim\mathcal N(\mathbf0,\mathbf I)$ and $\tau\sim\mathcal U(0,1)$ independently per sample and form $\mathbf A_t^\tau=(1-\tau)\boldsymbol\epsilon+\tau\mathbf A_t^*$. The conditional velocity and training loss are
\begin{equation}
\begin{aligned}
\hat{\mathbf v}_t^\tau
&=\mathbf v_\xi(\mathbf A_t^\tau,\tau,\mathbf c_t),\\
\mathcal L_{\mathrm{act}}
&=\mathbb E_{\tau,\boldsymbol\epsilon}
  \!\left[\operatorname{MSE}_{\mathrm{valid}}
  \!\left(\hat{\mathbf v}_t^\tau,\mathbf A_t^*-\boldsymbol\epsilon\right)\right].
\end{aligned}
\label{eq:act}
\end{equation}
Let $\mu^{\mathrm a}_{bh}$ indicate a valid control step $h$ of sample $b$, and let $d_a=7$ denote the action dimension. For predicted and target velocity tensors $\mathbf U$ and $\mathbf V$,
\begin{equation}
\begin{aligned}
\operatorname{MSE}_{\mathrm{valid}}(\mathbf U,\mathbf V)
&=\frac{1}{d_a Z_{\mathrm a}}\sum_{b,h}\mu^{\mathrm a}_{bh}
  \|\mathbf U_{bh}-\mathbf V_{bh}\|_2^2,\\
Z_{\mathrm a}&=\max\!\bigl(1,\sum\nolimits_{b,h}\mu^{\mathrm a}_{bh}\bigr).
\end{aligned}
\label{eq:action-reduction-detail}
\end{equation}
The mask excludes right-padded steps at the end of a demonstration. The reduction averages over valid sample--step positions and action coordinates, and returns zero when all steps are masked.

\paragraph{World loss.}
Indices $(b,t,k)$ identify a sample, a history decision time and an object; their targets refer to the following decision time. Let $\mathbf y^{c}=\mathbf c$ denote content, $\mathbf y^{x}\in\mathbb R^3$ position and $\mathbf y^{r}\in\mathbb R^6$ rotation. Each component has its own supervision mask and normalization:
\begin{equation}
\begin{aligned}
\mathcal L_{\mathrm{world}}&=E_c+\lambda_p(E_x+E_r),\\
E_d&=\frac{1}{Z_d}\sum_{b,t,k}\mu^d_{btk}
  \|\hat{\mathbf y}^{d}_{btk}-\mathbf y^{d,*}_{btk}\|_2^2,
  \qquad d\in\{c,x,r\},\\
Z_d&=\max\!\bigl(1,\sum\nolimits_{b,t,k}\mu^d_{btk}\bigr).
\end{aligned}
\label{eq:world}
\end{equation}
Here $\lambda_p=1$. The content mask $\mu^c$ selects tracked objects visible at both query times, excluding robot slots, padding and missing future observations. Position and rotation use $\mu^x=\mu^c v^{\rm pos}_{t+1}$ and $\mu^r=\mu^c v^{\rm rot}_{t+1}$, respectively. Squared errors are summed over coordinates and averaged over each component's valid targets; an empty component contributes zero. The detached content target is $\mathbf c^*_{btk}=\bar f_{\mathrm{cnt}}(\mathbf{raw}_{b,t+1,k})$, with unavailable geometry zero-filled as in Supp.~\ref{app:slot-features}. The main world loss updates $h_\psi$ and the trainable shared representations.

\paragraph{Auxiliary motion-monitor loss.}
The optional monitor uses the same targets, component-validity masks and reductions as Eq.~\eqref{eq:world}, applied to $g_\omega(\operatorname{sg}(\mathbf H_{k,t}),\mathbf u_t)$. Its loss is added as $\lambda_m\mathcal L_{\mathrm{mon}}$, with $\lambda_m=1$ in Stage~I and $0.5$ in Stage~II. Gradients stop at the stored trunk features, and the auxiliary gradients are clipped separately from the policy gradients.

\paragraph{Identity-code redraw.}
The augmentation retains every object and history frame while drawing a new assignment of distinct identity codes, independently of slot order. Addresses and the address components of all history slots are recomputed, and the slots are re-embedded with $f_\phi$. Pointer distributions are aligned by object identity. Both action chunks are integrated from the same Gaussian noise, with the original outputs detached as consistency targets.

Let $\mathcal B_v$ denote samples with at least one original valid object. Write $\bar{\boldsymbol\alpha}^{\mathrm{orig}}_{br}$ and $\bar{\boldsymbol\alpha}^{\mathrm{aug}}_{br}$ for pointer distributions restricted to the original objects and renormalized. The composition loss is
\begin{equation}
\begin{aligned}
\mathcal L_{\mathrm{compose}}
={}&\frac{1}{\max(1,R|\mathcal B_v|)}
  \sum_{b\in\mathcal B_v}\sum_{r=1}^{R}
  \operatorname{KL}\!\left(
  \operatorname{sg}(\bar{\boldsymbol\alpha}^{\mathrm{orig}}_{br})
  \,\Vert\,\bar{\boldsymbol\alpha}^{\mathrm{aug}}_{br}\right)\\
 &+\operatorname{MSE}\!\left(
  \operatorname{sg}(\mathbf A^{\mathrm{orig}}),\mathbf A^{\mathrm{aug}}\right).
\end{aligned}
\label{eq:compose-detail}
\end{equation}
The MSE averages over the batch and all $H\times7$ generated action coordinates. The KL term averages over roles and samples in $\mathcal B_v$; samples without valid objects contribute only to action consistency. Logarithm arguments and distribution-normalization denominators are floored at $10^{-8}$. Redraw consistency encourages the policy to preserve object selection and actions across identity-code assignments.

\paragraph{Distractor insertion.}
An object slot from another sample, including its history and address, is inserted into a free slot. Its identity code is replaced by one unused in the recipient. Equation~\eqref{eq:compose-detail} applies after dropping the inserted pointer column and renormalizing both distributions over the original objects. The original and augmented chunks again share their initial Gaussian noise. Each batch uses either redraw or insertion, with probability $0.5$ each. The composition weight ramps from $0$ to $0.1$ over the first $30\%$ of Stage~II. Slot-permutation equivariance is treated separately in Supp.~\ref{app:equivariance}.

\paragraph{Weak role supervision.}
The first two role queries are supervised with target and reference pseudo-labels from the Qwen3-VL relation graph. Using the one-based role indices of Eq.~\eqref{eq:pointer}, the loss is
\begin{equation}
\begin{aligned}
\mathcal L_{\mathrm{role}}
={}&-\frac{1}{\max(1,|\mathcal B_{\mathrm{tar}}|)}
  \sum_{b\in\mathcal B_{\mathrm{tar}}}
  \log\alpha_{b,1,j_b^{\mathrm{tar}}}\\
 &-\frac{1}{\max(1,|\mathcal B_{\mathrm{ref}}|)}
  \sum_{b\in\mathcal B_{\mathrm{ref}}}
  \log\alpha_{b,2,j_b^{\mathrm{ref}}}.
\end{aligned}
\label{eq:role-ce-detail}
\end{equation}
Each set contains samples whose corresponding label exists and identifies a valid current-frame object slot. Target and reference terms are reduced independently, so reference-only samples still contribute. Empty sets contribute zero; probabilities are floored at $10^{-8}$ before taking logarithms. The weight is $0.05$ for Stage-II steps $0$--$49{,}999$ and $0$ thereafter.

\subsection{Three-stage training}\label{app:training-detail}

Stage~0 adapts the public Chameleon-7B trunk~\citep{chameleon} to robot demonstrations. Its causal sequence contains the instruction and four groups of image, state and past-action tokens, followed by the action query and a teacher-forced suffix of up to $16\times7$ discrete future-action tokens. For each valid token $z_j$, the logit at position $j-1$ predicts $z_j$; the cross-entropy is averaged only over these targets~\citep{openvla}. The image tokenizer is frozen; the trunk, input embeddings and vocabulary projection are trainable. Our models in Tab.~\ref{tab:controls} share the Stage-0 checkpoint. The two controls without Stage~0 in Tab.~\ref{tab:abl-train} (A5, A6) start directly from the public Chameleon-7B weights.

Stage~I aligns the slot encoders and pre-execution world head on LIBERO, DROID~\citep{droid} and the same 100,000 Open X-Embodiment (OXE) training trajectories~\citep{openx} used in Stage~0, with the trunk and vocabulary projection frozen. Model (a) uses the same modules with OA disabled. Stage~II adds the pointer and action head and applies rank-32 LoRA ($\alpha=64$, dropout $0.05$) to the query, key, value, output, gate, up and down projections in all 32 layers. LIBERO-Plus perturbations and evaluation episodes are excluded throughout training.

\paragraph{Data mixture and splits.}
Stage~0 samples DROID and OXE equally. Stage~I keeps DROID and OXE at 40\% each and draws the remaining 20\% from the four standard LIBERO suites. In both stages, the RT-1, BridgeData V2, Kuka and BC-Z shares within OXE are 45\%, 30\%, 15\% and 10\%, respectively. Stage~0 contains 160,000 training and 8,000 validation trajectories; Stage~I contains 161,800 and 8,200. DROID is listed outside the OXE subtotal and contributes only trajectories with complete images, timestamps, actions and required annotations. Stage~II uses only the downstream training split: standard LIBERO for LIBERO and LIBERO-Plus, or BridgeData V2 for WidowX.

\begin{center}\small
\setlength{\tabcolsep}{4pt}
\begin{tabularx}{\linewidth}{@{}l X rrrr@{}}
\toprule
& & \multicolumn{2}{c}{Trajectories} & \multicolumn{2}{c}{Sampling probability (\%)}\\
\cmidrule(lr){3-4}\cmidrule(l){5-6}
Source & Suites or OXE configuration & Train & Validation & Stage~0 & Stage~I\\
\midrule
DROID & Official release & 60,000 & 3,000 & 50 & 40\\
LIBERO & Spatial, Object, Goal, Long & 1,800 & 200 & 0 & 20\\
\midrule
OXE: Google RT-1 & \nolinkurl{fractal20220817_data} & 45,000 & 2,250 & 22.5 & 18\\
OXE: BridgeData V2 & \nolinkurl{bridge_orig} & 30,000 & 1,500 & 15 & 12\\
OXE: Kuka & \nolinkurl{kuka} & 15,000 & 750 & 7.5 & 6\\
OXE: BC-Z & \nolinkurl{bc_z} & 10,000 & 500 & 5 & 4\\
\textit{OXE subtotal} & & 100,000 & 5,000 & 50 & 40\\
\bottomrule
\end{tabularx}
\end{center}

Simulation runs use the final Stage-II parameter EMA for evaluation. The separate content-target EMA supplies only the world target. Stages~I and II use AdamW with $(\beta_1,\beta_2)=(0.9,0.95)$, weight decay $0.05$, 2,000 warmup steps and cosine decay to $10\%$ of the peak learning rate. Gradients are clipped to norm 1.0, and training runs in BF16. Accumulation over four micro-batches gives global batches of 256 ($8\times8\times4$) in Stage~I and 128 ($8\times4\times4$) in Stage~II. Simulation fine-tuning is repeated with three random seeds. The main loss weights in Stage~II are $\lambda_w=0.5$ and $\lambda_v=0.04$; $\lambda_c$ ramps to $0.1$ over the first 30k updates of the full Stage-II schedule, and $\lambda_r=0.05$ for the first 50k updates only. The optional monitor has its own stopped-gradient loss. Algorithm~\ref{alg:oawam-train} gives one optimizer step.

\begin{table}[!ht]
\centering\small
\caption{Simulation training schedule. Parameter counts exclude frozen perception and EMA copies.}
\begin{tabular}{p{0.24\linewidth}p{0.20\linewidth}p{0.21\linewidth}p{0.23\linewidth}}
\toprule
Setting & Stage 0 & Stage I & Stage II (LoRA) \\
\midrule
Training data & DROID, OXE & LIBERO, DROID, OXE & LIBERO or BridgeData V2 \\
Main trainable parameters & $7.0$B & $24.1$M & $128.6$M \\
Optional motion monitor & --- & $5.9$M & $5.9$M \\
Training updates & $100{,}000$ & $50{,}000$ & $100{,}000$ \\
Peak learning rate & $2\times10^{-5}$ & $2\times10^{-4}$ & $2\times10^{-4}$ \\
Global batch & $512$ & $256$ & $128$ \\
Accumulation & $4$ & $4$ & $4$ \\
History & $T=4$ & $T=4$ & $T=4$ \\
Action horizon & $16$ (discrete targets) & --- & $16$ (flow target) \\
Query stride & Per source & Per source & LIBERO $16$; BridgeData V2 $4$ \\
OA operations & Off & On; off for (a) & On; off for (a) \\
Main objective & Action-token CE & $\mathcal L_{\rm world}$ & Eq.~\eqref{eq:loss} \\
Auxiliary monitor weight & --- & $1$ & $0.5$ \\
Parameter EMA & --- & --- & $0.999$ \\
Content-target EMA & --- & $0.996$ & $0.996$ \\
Route dropout & --- & --- & $0.5$ \\
\bottomrule
\end{tabular}
\end{table}

The slot modules contain 18,578,056 parameters; the world and action heads have 5,525,257 and 24,531,120, respectively. LoRA adds 79,953,920 parameters, giving a Stage-II policy total of 128,588,353. The optional motion monitor adds 5,922,597; the OA key path, address writes and reused vocabulary projection add none. Frozen perception models, including the MoGe-3 estimator, are outside these totals.

\begin{algorithm*}[!htbp]
\caption{\method{} Stage-II optimizer step.}
\label{alg:oawam-train}
\begin{algorithmic}[1]
\Require micro-batches with cached perception, per-frame proprioception, next-query slot vectors, action chunks and validity masks; main-model parameters $\theta=\{f_\mathrm{addr},f_\mathrm{cnt},f_\phi,\mathbf c,\psi,\xi,\text{LoRA}\}$; optional monitor parameters $\{f_u,\omega\}$.
\For{each of the $4$ accumulated micro-batches}
    \State draw identity codes and compute fresh, differentiable addresses for every sample
    \State close the \tok{[act\_q]}$\to$image-code route in a random half of the samples \Comment{route dropout}
    \State encode the history with the causal sequence and slot-aware trunk (Supp.~\ref{app:tokenizer-detail} and~\ref{app:oa-impl}) $\to\mathbf{H}$ \Comment{address writes pass gradients}
    \State $\mathcal L_\mathrm{world}$ from $h_\psi(\mathbf H_{k,t})$ and next-query targets, using component-validity masks \Comment{Eq.~\eqref{eq:world}}
    \State if enabled, $\mathcal L_\mathrm{mon}$ regresses the same targets from $g_\omega(\operatorname{sg}(\mathbf H),\mathbf u)$
    \State $\mathcal{L}_\mathrm{act}$: pointer-conditioned flow matching over valid action steps \Comment{Eq.~\eqref{eq:act}}
    \State $\mathcal{L}_\mathrm{vq}$: VQ code $j$ of frame $t$ predicts code $j$ of frame $t{+}1$
    \State $\mathcal{L}_\mathrm{compose}$: redraw identity codes or insert a distractor; second forward pass; aligned KL + action MSE with shared noise
    \State $\mathcal{L}_\mathrm{role}$ on the first two pointer distributions (step $<50$k)
    \State backpropagate $(\mathcal L+\lambda_m\mathcal L_\mathrm{mon})/4$; omit the auxiliary term when disabled (Eq.~\eqref{eq:loss})
\EndFor
\State average gradients across data-parallel ranks; clip main-model and auxiliary gradients separately to $1.0$; AdamW step
\State parameter EMA (decay $0.999$); content-target EMA $\bar f_\mathrm{cnt}\gets 0.996\,\bar f_\mathrm{cnt}+0.004\,f_\mathrm{cnt}$
\end{algorithmic}
\end{algorithm*}

\FloatBarrier
\section{Closed-loop execution and re-grounding}\label{app:inference-detail}

\subsection{Forecast before execution, comparison after observation}

At decision $t$, the same trunk pass supplies the next action chunk and a forecast indexed by the identity addresses. The policy stores this forecast before executing the first $K$ actions. At $t+1$, it compares the stored forecast with the newly observed states, re-grounds objects when the recovery rule fires and conditions the next action on the updated slots. Algorithm~\ref{alg:oawam-infer} shows where the observation, forecast and action enter this loop.

\begin{algorithm*}[!htbp]
\caption{\method{} policy query at time $t$ (queried every $K$ control steps).}
\label{alg:oawam-infer}
\begin{algorithmic}[1]
\Require observation $(\mathbf{I}_t,\mathbf{q}_t)$, instruction $\ell$; episode state: identities $\{(\boldsymbol{\ell}_k,\mathbf{f}_k^{(0)})\}$, addresses $\{\mathbf{addr}_k\}$, $T$-frame history, filter and tracker state, the previous query's cached forecast $\widehat{\mathcal S}_t$; hidden states $\mathbf H_{k,t-1}$ and proprioception $\mathbf q_{t-1}$ when auxiliary motion monitoring is enabled.
\State estimate geometry from RGB and align calibrated image and base-frame coordinates
\If{$t{=}0$}
    \State ground objects with Qwen3-VL prompts and SAM~3 masks; pool DINOv3 features within each mask and estimate the pose from the aligned geometry
    \State draw identity codes $\iota_k$; $\mathbf{addr}_k\gets f_\mathrm{addr}([\boldsymbol{\ell}_k;\mathbf{f}_k^{(0)}])+\mathbf{c}_{\iota_k}$ \Comment{fixed for the episode, Eq.~\eqref{eq:addr}}
\EndIf
\State track identities and assemble $\mathbf{raw}_k^t$ with component-validity flags \Comment{Supp.~\ref{app:tokenizer-detail}}
\If{$t{>}0$}
    \State read cached $\widehat{\mathcal S}_t$; optionally use $g_\omega(\mathbf H_{k,t-1},\mathbf u_{t-1})$ for monitoring; compute $e_k$ with the new observation
    \If{$e_k>\tau_{\mathrm{mon}}$ for a named object, or a named object disappeared}
        \State re-detect; re-solve identity$\leftrightarrow$detection assignment (Hungarian on label, descriptor and predicted position); admit new objects
        \State re-observe $\mathbf{raw}_k^t$ under the repaired assignment; replace the history of reassigned objects
    \EndIf
\EndIf
\State $\mathbf{cnt}_k^t\gets f_\mathrm{cnt}(\mathbf{raw}_k^t)$;\; $\tilde{\mathbf{s}}_k^t\gets f_\phi([\mathbf{addr}_k\Vert\mathbf{cnt}_k^t\Vert\boldsymbol{\pi}^t\Vert\boldsymbol{\rho}_k])$
\State build the $T$-frame block-causal sequence (image codes, proprioception and slots per frame, plus executed actions in historical groups); scatter $\tilde{\mathbf{s}}_k^t$ into the slot positions
\State $\mathbf{x}^{(0)}\gets$ embeddings with $\mathbf{x}^{(0)}_{k,\mathcal{A}}\gets\mathbf{addr}_k$
\For{$m=0,\dots,31$}
    \State $\mathbf{x}^{(m+1)}\gets\textsc{Layer}_m(\mathbf{x}^{(m)})$ with slot keys from the address coordinates \Comment{Eq.~\eqref{eq:oa}}
    \State $\mathbf{x}^{(m+1)}_{k,\mathcal{A}}\gets\mathbf{addr}_k$ \Comment{per-layer address write}
\EndFor
\State $\widehat{\mathcal S}_{t+1}\gets\{h_\psi(\mathbf H_{k,t})\}_k$; cache the forecast; retain $\mathbf H_{k,t},\mathbf q_t$ if auxiliary motion monitoring is enabled
\State $\boldsymbol{\alpha},\mathbf{r}\gets$ action pointer (Eq.~\eqref{eq:pointer});\; $\mathbf{A}_t\gets\textsc{Euler}_4(\mathbf{v}_\xi;\,[W_c\mathbf{h}_\mathrm{q};\mathbf{r}])$ \Comment{Eq.~\eqref{eq:euler-app}}
\State \Return $(\mathbf{A}_t,\widehat{\mathcal S}_{t+1})$
\end{algorithmic}
\end{algorithm*}

\subsection{Recovery trigger and reassignment}\label{app:reassignment}

The default prediction is the cached world-head forecast; when auxiliary motion monitoring is selected, $g_\omega$ instead uses stored $\mathbf H_{k,t}$ and the now-observed $\mathbf u_t$. For either source, the new observation is mapped into target space using $\bar f_{\mathrm{cnt}}$ and normalized poses. Writing $(\hat{\mathbf c}_k^{t+1},\hat{\mathbf p}_k^{t+1})$ for the selected monitoring prediction, the residual is
\begin{equation}
\begin{aligned}
e_{k,t+1}
={}&\frac{\|\hat{\mathbf c}_k^{t+1}-\mathbf c_k^{*,t+1}\|_2^2}{256}\\
&+\frac{\lambda_p}{9}\left(
v_{k,t+1}^{\rm pos}\|\hat{\mathbf x}_k^{t+1}-\mathbf x_k^{*,t+1}\|_2^2
+v_{k,t+1}^{\rm rot}\|\hat{\mathbf r}_k^{t+1}-\mathbf r_k^{*,t+1}\|_2^2\right).
\end{aligned}
\label{eq:monitor-residual}
\end{equation}
Here $\mathbf x$ and $\mathbf r$ denote the position and rotation components of $\mathbf p$. Residuals are evaluated for objects visible at both queries. Missing geometry contributes zero to its residual term; disappearance is determined by object visibility. Each predictor, checkpoint, query interval and depth source has its own threshold and position-error scale.

\paragraph{Episode-level threshold.}
With the policy and predictor frozen, we collect held-out nominal rollouts from the deployment tasks and initial-state distribution, without injected faults. One set of 100 episodes fits the position-error scale $\sigma_p$, and a separate set of 400, run with recovery disabled, calibrates the threshold. For each calibration episode with at least one evaluated residual, $s_j=\max_{t,k}e_{k,t}$ over named objects and queries; $n$ counts these scored episodes. The threshold $\tau_{\mathrm{mon}}$ is the $\lceil(n+1)(1-\alpha)\rceil$-th order statistic with $\alpha=0.05$ (rank 381 for $n=400$). Like the residual in Eq.~\eqref{eq:monitor-residual}, it is dimensionless. Under exchangeability, a new nominal episode scored the same way exceeds this threshold with probability at most $\alpha$~\citep{sentinel,faildetect,fiper}. The disappearance of a named object can also trigger re-grounding and is outside this bound. A single exceedance triggers re-grounding; a three-query cooldown follows each trigger. The clean-event column in Tab.~\ref{tab:abl-monitor} counts interventions from either trigger; one episode can contribute more than once.

The observation-change baseline uses the previous observation as its prediction. The kinematic predictor keeps objects stationary unless the gripper was closed within 8\,cm of an object at the preceding query, in which case the object follows the end effector. The auxiliary motion monitor uses the observed change in proprioception (Supp.~\ref{app:heads}). All three alternatives use the same calibration procedure.

\paragraph{Re-grounding.}
The perception stack re-detects objects with the episode's phrases. The assignment between the episode's identities and the detections minimizes $\sum_{(k,j)}\bigl[(1-\cos(\boldsymbol{\ell}_k,\boldsymbol{\ell}'_j))+(1-\cos(\mathbf{f}_k^{(0)},\mathbf{f}'_j))+d^2_{kj}/\chi^2_{3,0.99}\bigr]$ over matched pairs (Hungarian algorithm). The position term $d^2_{kj}=\|\hat{\mathbf x}_k-\mathbf x'_j\|^2/\sigma_p^2$ compares predicted and detected 3-D positions in the robot base frame; the predicted position is mapped back from the workspace-normalized coordinates of Eq.~\eqref{eq:monitor-residual}. For base-frame prediction errors $\boldsymbol\epsilon=(\epsilon_x,\epsilon_y,\epsilon_z)$ on the scale-fitting episodes,
\[
\sigma_p=\sqrt{\bigl(\widehat{\mathrm{Var}}(\epsilon_x)+\widehat{\mathrm{Var}}(\epsilon_y)+\widehat{\mathrm{Var}}(\epsilon_z)\bigr)/3}
\]
is in meters, so $d^2_{kj}$ is dimensionless. As in DeepSORT~\citep{deepsort}, pairs outside the gate ($d^2_{kj}>\chi^2_{3,0.99}=11.34$) are excluded; pairs with total cost above $1.5$ are rejected. The position term and its gate are applied only when both positions are available; other pairs use label and appearance costs. Identities without a monitoring prediction use the Kalman prediction when available. Matched identities keep their addresses and follow the new tracks; unmatched detections are admitted into free slots with fresh addresses and unused identity codes; unmatched identities keep their addresses and are marked invisible. The filter and tracker states of reassigned objects are reset, their entries in the history frames are replaced by the new observation, and their residuals are skipped at the next query. The consumed forecast is discarded after re-grounding, and the repaired observation produces the next forecast. The auxiliary monitor reuses hidden states of the regular forward pass; a re-grounding event costs one re-detection.

The pre-execution world head produces one forecast for a given history, regardless of the action chunk sampled. A track switch can yield a large residual, and so can a valid action that moves a correctly tracked object in an atypical way. Across the 1,200 clean episodes of Tab.~\ref{tab:abl-monitor}, the forecast-residual rule intervenes 61 times for disappearance or a forecast residual (0.051 per episode). Conditioning the forecast on the sampled chunk, as action-conditioned world models do~\citep{checkvla}, would separate the two sources.

\subsection{Simulation component timings}\label{app:latency}

In the default simulation settings, each query predicts $H=16$ actions. LIBERO executes $K=16$ at $20$\,Hz; SimplerEnv executes $K=4$ at $5$\,Hz. Both configurations therefore have a nominal command duration of $800$\,ms per query. The table reports measured component runtimes for the RGB policy, including geometry and the pre-execution forecast, and their sum. Instruction parsing runs only at the first query, so first-query sums exceed the nominal period and later sums fall below it. The simulator waits for each policy query.

\begin{center}\small
\begin{tabularx}{\linewidth}{@{}Xcc@{}}
\toprule
Component (ms) & LIBERO: first / later & SimplerEnv: first / later \\
\midrule
MoGe-3 ViT-L geometry & 308 / 308 & 154 / 154 \\
Geometry alignment and pose & 10 / 10 & 6 / 6 \\
SAM~3 detection/tracking and DINOv3 pooling & 175 / 105 & 95 / 55 \\
Qwen3-VL parsing (first query) & 500 / 0 & 500 / 0 \\
Temporal filtering and residual evaluation & 4 / 6 & 3 / 4 \\
Sequence construction, image VQ and transfers & 25 / 25 & 23 / 23 \\
Slot-aware 7B trunk, BF16 & 125 / 125 & 125 / 125 \\
Action head (four Euler steps) & 12 / 12 & 12 / 12 \\
Pre-execution world head and forecast cache & 2 / 2 & 2 / 2 \\
\textbf{Sum of components} & \textbf{1,161 / 593} & \textbf{920 / 381} \\
Optional motion monitor (additional) & --- / +2 & --- / +2 \\
Re-grounding, when triggered (additional) & --- / +100 & --- / +55 \\
\bottomrule
\end{tabularx}
\end{center}

\FloatBarrier
\section{Simulation benchmarks and comparison policies}\label{app:eval}

\subsection{Tasks, trials and aggregation}

The \method{} entries in Tabs.~\ref{tab:libero-simpler} and~\ref{tab:libero-plus} average three fine-tuning runs. The same LIBERO-trained policies are evaluated on standard and perturbed scenes; the policies fine-tuned on BridgeData~V2 are evaluated on WidowX. Trial counts in the table below are per training seed.

\begin{center}\small
\begin{tabularx}{\linewidth}{@{}l X X@{}}\toprule
Benchmark & Evaluation unit & Success and aggregation\\\midrule
LIBERO~\citep{libero} & Four suites. Per suite: 10 tasks $\times$ 50 initial states $\times$ 3 rollouts. & Environment task completion; mean of four suite rates.\\
LIBERO-Plus~\citep{liberoplus} & 10,030 zero-shot perturbation trials on seven axes. & Task completion; Overall pools trials, Geo Avg weights Camera, Robot init and Layout equally.\\
SimplerEnv~\citep{simplerenv} & Four WidowX visual-matching tasks; 24 episodes per task. & Final-step task completion; equal-weight mean over tasks.\\\bottomrule
\end{tabularx}\end{center}

On WidowX, the three policies completed 67 spoon-to-towel, 67 carrot-to-plate, 52 block-stacking and 70 eggplant-to-basket episodes out of $3\times24=72$ per task (256/288 overall). On standard LIBERO, model (b) completed $5{,}941$, $5{,}941$ and $5{,}944$ of $6{,}000$ trials across the three seeds ($99.0\%$, sample SD $0.03$ points). On full LIBERO-Plus, the three policies completed $27{,}240$ of $30{,}090$ trials ($90.53\%$, sample SD $0.03$ points); Geo Avg is $89.33\%$ (sample SD $0.04$ points). The table below gives per-seed successes (success rates in \%) and across-seed summaries; Geo Avg entries are rates only. Table~\ref{tab:per-seed} gives the suite and axis rates for the addressing controls. With both views, OA improves every LIBERO suite and every LIBERO-Plus axis in all three seeds. Using unrounded model (b) rates, the mean gains are $13.3$, $11.2$ and $7.5$ percentage points on Camera, Robot init and Layout and $12.4$ points under sensor noise. With only the external camera in Stage~II and at evaluation, the geometric gains are $2.8$, $2.9$ and $4.7$ points, and each geometric axis again improves in every seed. Panels B--D of Tab.~\ref{tab:per-seed} record target choice, first grasp and task completion on a grasp-task subset; their differences use unrounded rates.

\begin{center}\small
\setlength{\tabcolsep}{4.0pt}
\begin{tabularx}{\linewidth}{@{}lr*{3}{Y}rr@{}}\toprule
Metric & \makecell{Trials\\per seed} & Seed 1 & Seed 2 & Seed 3 & \makecell{Mean\\(\%)} & \makecell{SD\\(pp)} \\ \midrule
Camera & 1,599 & 1,437 (89.87) & 1,437 (89.87) & 1,439 (89.99) & 89.91 & 0.07 \\
Robot init & 1,550 & 1,401 (90.39) & 1,405 (90.65) & 1,410 (90.97) & 90.67 & 0.29 \\
Layout & 1,525 & 1,337 (87.67) & 1,333 (87.41) & 1,329 (87.15) & 87.41 & 0.26 \\
Light & 1,142 & 1,104 (96.67) & 1,100 (96.32) & 1,107 (96.94) & 96.64 & 0.31 \\
BG & 1,076 & 1,057 (98.23) & 1,059 (98.42) & 1,060 (98.51) & 98.39 & 0.14 \\
Lang & 1,537 & 1,367 (88.94) & 1,369 (89.07) & 1,368 (89.00) & 89.00 & 0.07 \\
Noise & 1,601 & 1,377 (86.01) & 1,374 (85.82) & 1,370 (85.57) & 85.80 & 0.22 \\
\midrule
Geo Avg & --- & 89.31 & 89.31 & 89.37 & 89.33 & 0.04 \\
Overall & 10,030 & 9,080 (90.53) & 9,077 (90.50) & 9,083 (90.56) & 90.53 & 0.03 \\
\bottomrule\end{tabularx}\end{center}

\begin{table*}[!ht]
\centering

\caption{A: success rate (\%) by training seed for each LIBERO suite and LIBERO-Plus perturbation. Seeds 1--3 index the three training runs of each model; runs with the same index use matched initial states. Each LIBERO suite has $1{,}500$ rollouts per seed; the seven full LIBERO-Plus axis denominators are given above. Entries are rounded percentages; model (b)'s LIBERO-Plus entries are rounded directly from the integer counts above.}
\label{tab:per-seed}
\small
\setlength{\tabcolsep}{4pt}
\begin{tabular}{llccccccccccc}
\toprule
 & & \multicolumn{4}{c}{LIBERO} & \multicolumn{7}{c}{LIBERO-Plus} \\
\cmidrule(lr){3-6} \cmidrule(lr){7-13}
Model & Seed & Spatial & Object & Goal & Long & Cam & Init & Layout & Light & BG & Lang & Noise \\
\midrule
(h) & 1 & 97.2 & 98.1 & 98.0 & 96.1 & 76.4 & 79.2 & 83.3 & 95.4 & 95.5 & 86.5 & 83.3 \\
(h) & 2 & 96.8 & 97.5 & 98.1 & 96.9 & 76.2 & 81.2 & 81.4 & 96.4 & 94.5 & 86.2 & 82.1 \\
(h) & 3 & 97.7 & 98.2 & 97.3 & 97.2 & 75.3 & 81.0 & 83.0 & 95.8 & 93.7 & 86.4 & 84.7 \\
(a) & 1 & 96.8 & 98.8 & 97.0 & 95.8 & 74.5 & 78.8 & 78.0 & 94.9 & 96.1 & 85.8 & 72.0 \\
(a) & 2 & 97.4 & 98.8 & 97.8 & 96.9 & 78.8 & 81.6 & 81.0 & 95.2 & 96.1 & 85.8 & 74.9 \\
(a) & 3 & 97.6 & 98.5 & 97.9 & 95.3 & 76.4 & 77.9 & 80.7 & 95.1 & 95.8 & 84.7 & 73.4 \\
(b) & 1 & 99.1  & 99.9  & 98.8  & 98.3  & 89.9 & 90.4 & 87.7 & 96.7 & 98.2 & 88.9 & 86.0 \\
(b) & 2 & 99.0  & 99.9  & 98.7  & 98.5  & 89.9 & 90.6 & 87.4 & 96.3 & 98.4 & 89.1 & 85.8 \\
(b) & 3 & 98.9 & 99.9  & 99.0  & 98.5  & 90.0 & 91.0 & 87.1 & 96.9 & 98.5 & 89.0 & 85.6 \\
(a$'$) & 1 & 97.7 & 98.7 & 97.3 & 97.5 & 74.0 & 81.7 & 82.2 & 95.2 & 95.1 & 90.2 & 70.8 \\
(a$'$) & 2 & 97.8 & 98.3 & 97.8 & 96.9 & 71.1 & 83.3 & 78.9 & 96.6 & 96.3 & 89.1 & 70.3 \\
(a$'$) & 3 & 97.7 & 97.9 & 97.5 & 97.1 & 75.8 & 82.0 & 80.5 & 96.4 & 96.8 & 89.1 & 71.6 \\
(b$'$) & 1 & 98.9 & 98.5 & 98.0 & 95.8 & 76.8 & 84.6 & 85.8 & 96.1 & 95.3 & 86.9 & 71.8 \\
(b$'$) & 2 & 97.9 & 98.5 & 97.8 & 96.1 & 75.8 & 83.8 & 84.5 & 95.4 & 94.5 & 87.1 & 69.4 \\
(b$'$) & 3 & 98.5 & 98.7 & 98.7 & 96.6 & 76.7 & 87.2 & 85.3 & 96.8 & 96.6 & 87.8 & 73.9 \\
\bottomrule
\end{tabular}
\end{table*}

\paragraph{Target choice and first grasp.}
Across all analyzed pairs, OA raises pointer-hit rates by 13.2, 9.3 and 4.6 points and lowers wrong-first-grasp rates by 2.2, 4.5 and 8.7 points on Camera, Robot init and Layout, respectively (Tab.~\ref{tab:controls}C). Among trials where each policy grasps the target first, mean task completion changes by $+0.7$, $+4.2$ and $+3.3$ points (Tab.~\ref{tab:per-seed}D).

\textit{Pairing and coverage.} We compare the dual-view (a)/(b) checkpoints of Tab.~\ref{tab:controls}A on tasks whose first manipulation requires grasping and lifting one specified instance. Pairs match task, initial state, perturbation instance and rollout seed. Of $1{,}177$ assigned pairs, $1{,}084$ enter the analysis (Tab.~\ref{tab:per-seed}B); 47 of the 93 excluded pairs lack a log for at least one policy. Each exclusion removes the pair for both policies, and missed detections and no-grasp episodes remain in the sample. Even if every excluded pair counted as a pointer hit for (a) and a miss for (b), (b) would still have more pointer hits (824 vs.\ 819 of $1{,}177$).

\textit{Identity and pointer.} Simulator instance labels are used only for offline scoring. At the first query, before any action, the designated target track is mapped to the physical instance with the highest mask intersection-over-union in the external view. A unique maximum of at least 0.5 against the simulator instance masks is required; ties, invisibility or insufficient overlap count as absent/ambiguous. The slot selected by the pointer is labeled with the same rule. Initial wrong bindings are counted separately from absent or ambiguous ones. A later switch is counted when an initially correct track is observed on another instance before the first stable grasp (or termination if none); a visibility gap alone is not a switch. A pointer hit requires the highest-weight slot of the target role, not the reference role, to map uniquely to the instructed physical instance; a tie for the highest weight counts as a miss. An absent target remains in the denominator.

\textit{Grasp and completion.} A stable grasp requires both fingers to hold one object at least 1\,cm above its initial height for three consecutive simulator steps. The first qualifying interval identifies the instance; simultaneous intervals count as ambiguous, and no interval as no grasp. Later corrections do not change the first-grasp category. Table~\ref{tab:per-seed}D gives environment task successes and trial counts for each category. Full-sample pointer-hit and first-grasp rates use all $N$ pairs. Completion after a target-first grasp uses each policy's own target-first trials. For each axis we average unrounded within-seed rate differences equally; a rate with zero trials is undefined.

\textit{Reading the paired results.} Pointer hit improves in all nine axis--seed comparisons, from 726 to 824 of the $1{,}084$ analyzed pairs. Summed over the nine rows of Tab.~\ref{tab:per-seed}D, OA removes 56 wrong first grasps and adds 54 correct ones: other-first trials fall from 190 to 134 and target-first trials rise from 787 to 841, while no-grasp trials change little (76 and 80). Task successes rise from 653 to 737. The two policies share a frozen perception pipeline, but stochastic SAM~3 inference can give different initial bindings on a matched pair.

\textit{Target choice after correct initial binding.} From the same analyzed pairs, we retain those in which both target tracks matched the instructed physical instance at the first query, regardless of the subsequent grasp. This leaves 935 of 1,084 pairs. The pointer hits the target in 641 of these pairs with standard attention and 725 with OA. Within-seed gains are positive in every comparison: $13.15\pm3.99$ points on Camera, $9.29\pm3.28$ on Robot init and $4.63\pm1.45$ on Layout (mean $\pm$ sample SD of three paired rate differences; panel C). These gains match the all-pair gains to within 0.1 points, so differences in initial binding do not explain the pointer gains.

\begin{table}[!ht]
\centering\small\setlength{\tabcolsep}{3pt}
\caption*{\textbf{Table~\ref{tab:per-seed} (continued). B: pair coverage and target identity.} Assigned and analyzed $N$ are pair counts; the remaining cells give (a)/(b) counts on the analyzed pairs. Initial wrong and absent/ambiguous refer to the first query; later switch counts only initially correct tracks.}
\begin{tabularx}{\linewidth}{@{}l c*{6}{Y}@{}}
\toprule Axis & Seed & Assigned & Analyzed $N$ & Initial wrong & Absent / ambiguous & Later switch & Pointer hit\\
\midrule
Camera & 1 & 122 & 112 & 10/2 & 3/3 & 3/5 & 82/93 \\
Camera & 2 & 132 & 125 & 4/4 & 2/6 & 7/6 & 73/98 \\
Camera & 3 & 128 & 122 & 9/7 & 3/4 & 8/7 & 88/100 \\
Robot init & 1 & 120 & 112 & 10/5 & 3/3 & 4/2 & 72/86 \\
Robot init & 2 & 142 & 128 & 6/3 & 6/4 & 3/5 & 91/103 \\
Robot init & 3 & 124 & 116 & 8/4 & 8/5 & 5/3 & 73/80 \\
Layout & 1 & 137 & 126 & 10/2 & 3/5 & 4/4 & 81/91 \\
Layout & 2 & 139 & 123 & 9/7 & 4/2 & 4/3 & 80/83 \\
Layout & 3 & 133 & 120 & 6/9 & 8/1 & 8/1 & 86/90 \\
\bottomrule\end{tabularx}\par\medskip
\caption*{\textbf{C: target choice and first grasp.} On all pairs, entries count events in (b) only / (a) only. On the $M$ pairs with both initial target bindings correct, hits count first-query selections of the instructed instance; $\Delta=100(\mathrm{hits}_b-\mathrm{hits}_a)/M$.}
\begin{tabularx}{\linewidth}{@{}l c Y Y r r r r@{}}
\toprule
& & \multicolumn{2}{c}{All $N$ pairs} & \multicolumn{4}{c@{}}{Both initial bindings correct}\\
\cmidrule(lr){3-4}\cmidrule(l){5-8}
Axis & Seed & Pointer hit & Wrong first grasp & $M$ & Hits (a) & Hits (b) & $\Delta$ (pp)\\
\midrule
Camera & 1 & 18/7 & 4/8 & 96 & 75 & 87 & $+12.50$ \\
Camera & 2 & 30/5 & 3/10 & 109 & 67 & 86 & $+17.43$ \\
Camera & 3 & 18/6 & 6/3 & 105 & 78 & 88 & $+9.52$ \\
Robot init & 1 & 19/5 & 1/6 & 94 & 62 & 74 & $+12.77$ \\
Robot init & 2 & 19/7 & 2/7 & 113 & 83 & 93 & $+8.85$ \\
Robot init & 3 & 12/5 & 3/9 & 96 & 61 & 67 & $+6.25$ \\
Layout & 1 & 20/10 & 1/8 & 111 & 70 & 77 & $+6.31$ \\
Layout & 2 & 14/11 & 0/11 & 106 & 69 & 73 & $+3.77$ \\
Layout & 3 & 16/12 & 2/16 & 105 & 76 & 80 & $+3.81$ \\
\bottomrule\end{tabularx}
\end{table}

\begin{table}[!ht]
\centering\small\setlength{\tabcolsep}{4pt}
\caption*{\textbf{Table~\ref{tab:per-seed} (continued). D: first stable grasp and final task outcome.} Cells give successes / trials in each first-grasp category. The four trial counts sum to $N$ for each policy; their success counts sum to its total task successes. $0/0$ denotes an empty category with an undefined conditional rate. A wrong first grasp followed by eventual success remains in the other-first category.}
\begin{tabularx}{\linewidth}{@{}l c*{8}{Y}@{}}
\toprule
& & \multicolumn{2}{c}{Target first} & \multicolumn{2}{c}{Other first} & \multicolumn{2}{c}{No grasp} & \multicolumn{2}{c}{Ambiguous}\\
Axis & Seed & (a)&(b)&(a)&(b)&(a)&(b)&(a)&(b)\\
\midrule
Camera & 1 & 72/88 & 72/84 & 4/16 & 6/12 & 0/6 & 0/10 & 0/2 & 0/6 \\
Camera & 2 & 70/87 & 82/103 & 4/23 & 7/16 & 0/13 & 0/6 & 0/2 & 0/0 \\
Camera & 3 & 69/90 & 69/91 & 4/17 & 6/20 & 0/10 & 0/8 & 1/5 & 0/3 \\
Robot init & 1 & 68/84 & 67/89 & 6/15 & 5/10 & 0/10 & 0/12 & 0/3 & 0/1 \\
Robot init & 2 & 64/91 & 85/104 & 2/24 & 6/19 & 1/9 & 0/4 & 0/4 & 0/1 \\
Robot init & 3 & 59/82 & 67/85 & 7/23 & 9/17 & 0/4 & 1/10 & 0/7 & 0/4 \\
Layout & 1 & 74/95 & 79/98 & 4/19 & 6/12 & 0/9 & 0/12 & 0/3 & 0/4 \\
Layout & 2 & 70/89 & 78/96 & 3/20 & 5/9 & 0/11 & 1/11 & 0/3 & 0/7 \\
Layout & 3 & 64/81 & 76/91 & 7/33 & 10/19 & 0/4 & 0/7 & 0/2 & 0/3 \\
\bottomrule\end{tabularx}
\end{table}

\subsection{Public baselines and sources of benchmark scores}\label{app:baseline-sources}

The ten baselines in Tabs.~\ref{tab:libero-simpler} and~\ref{tab:libero-plus} have official releases with public code and model weights: \href{https://github.com/moojink/openvla-oft/blob/main/LIBERO.md}{OpenVLA-OFT}, \href{https://github.com/Physical-Intelligence/openpi}{$\pi_0$-FAST and $\pi_{0.5}$}, \href{https://github.com/NVIDIA/Isaac-GR00T/tree/n1.5-release}{GR00T N1.5}, \href{https://github.com/OpenDriveLab/UniVLA}{UniVLA}, \href{https://github.com/yuantianyuan01/FastWAM}{Fast-WAM}, \href{https://github.com/NVlabs/cosmos-policy}{Cosmos Policy}, \href{https://github.com/Robbyant/lingbot-va}{LingBot-VA}, \href{https://github.com/ginwind/VLA-JEPA}{VLA-JEPA} and \href{https://github.com/InternRobotics/InternVLA-A-series}{InternVLA-A1.5}. Licenses and checkpoint coverage vary: for example, \href{https://huggingface.co/nvidia/GR00T-N1.5-3B}{GR00T N1.5} releases a base policy and \href{https://huggingface.co/robbyant/lingbot-va-posttrain-libero-long}{LingBot-VA} releases a LIBERO-Long checkpoint.

For standard LIBERO we use the method authors' four-suite results: OpenVLA-OFT (one policy per suite, Table I)~\citep{openvlaoft}, $\pi_0$-FAST (\href{https://github.com/Physical-Intelligence/openpi/blob/e4580662a60e528ee8e39e7073f82affec7dd407/examples/libero/README.md}{openpi, 30k steps}) and $\pi_{0.5}$ (openpi; $96.85\%$, rounded to $96.9\%$)~\citep{openpi}, UniVLA (Table I)~\citep{bu2025univla}, the base Fast-WAM (Table 2)~\citep{fastwam}, Cosmos Policy (Table 1)~\citep{cosmospolicy} and LingBot-VA (Table 2)~\citep{lingbotva}. GR00T N1.5 is the unmodified model in Table V of~\citet{baiactionerror}; the source orders Goal before Object.

On WidowX, OpenVLA-OFT is reported in Table 4 of~\citet{embodiedrone} and GR00T N1.5 in Table 2 of~\citet{internvlam1}. For $\pi_0$-FAST we use the BridgeData V2 reproduction by~\citet{jang2026mgselect}: 24 trials per task, using the unmodified policy and the task-success columns (Tables 2 and 9; $46.9\%$). For $\pi_{0.5}$ we use the independently trained supervised baseline in Table 1 of~\citet{starevla}, before RL, and report task completion rather than grasp success. For UniVLA we use the BridgeData V2 LoRA-adapted UniVLA row of Table A-III in~\citet{bu2025univla} (arXiv v3). For OpenVLA-OFT, GR00T N1.5 and UniVLA, we recompute the equal-task means from the four displayed task rates ($41.7\%$, $62.0\%$ and $48.0\%$); the sources report $41.8\%$, $61.9\%$ and $47.9\%$. The cited sources give no WidowX results for Fast-WAM, Cosmos Policy or LingBot-VA.

For zero-shot LIBERO-Plus we use OpenVLA-OFT, $\pi_0$-FAST and UniVLA from the benchmark authors' Table 2~\citep{liberoplus}; $\pi_{0.5}$, Fast-WAM and Cosmos Policy from Table 4 of~\citet{wam-vs-vla-2026}; and GR00T N1.5 from Table 5 of~\citet{jeonvpm}, which reports only an overall rate. For VLA-JEPA and InternVLA-A1.5, we use the authors' LIBERO, WidowX and LIBERO-Plus results (Tables 1/2/3~\citep{vlajepa} and 5/2/6~\citep{internvlaa15}). The literature entries in Tabs.~\ref{tab:libero-simpler} and~\ref{tab:libero-plus} use those published evaluations and the LingBot-VA weighting below; their views, controllers, training data and trial counts can differ from ours. Published Overall entries are the totals each source reports and can aggregate the axes differently (VLA-JEPA reports the unweighted seven-axis mean); \method{} leads under both aggregations, with $90.5\%$ pooled and $91.1\%$ unweighted. Our matched full LIBERO-Plus evaluations of $\pi_{0.5}$ and Cosmos Policy and the separate local instruction-rewrite tests (Tab.~\ref{tab:baseline-artifacts}) are in Supp.~\ref{app:baseline-artifacts}; Tab.~\ref{tab:real} uses our own PiPER trials.

\paragraph{Trial weighting for LingBot-VA.}
We weight the published success rate $s_i$ for each perturbation (Table~5 of~\citet{zimablue}) by its trial count $n_i$ (Table~7 of~\citet{liberoplus}). The resulting overall rate, $\sum_i n_i s_i/\sum_i n_i$, rounds to $69.5\%$, also reported in Table~5 of~\citet{mmabclp}. ZimaBlue itself is concurrent work: Table~5 of~\citet{zimablue} reports unweighted seven-axis means of $86.7\%$ in zero-shot transfer from standard LIBERO and $92.0\%$ after fine-tuning on LIBERO-Plus. On the same unweighted zero-shot mean, \method{} reaches $91.1\%$.

\subsection{Local evaluation of public checkpoints}\label{app:baseline-artifacts}

\noindent\textbf{Matched full LIBERO-Plus evaluation.} We evaluated two public policies: $\pi_{0.5}$ (\nolinkurl{pi05_libero}) and Cosmos Policy (\nolinkurl{nvidia/Cosmos-Policy-LIBERO-Predict2-2B}). For each policy, three inference runs used the same fixed weights and the same 10,030 initial-state and perturbation conditions, success criteria and per-task step limits as \method{}. The tables below report sample SD across inference runs for the public policies and across training seeds for \method{}. The run columns give success counts except for Geo Avg, which is a percentage. Gains are computed before rounding.

\begin{center}\small
\setlength{\tabcolsep}{4pt}
\begin{tabularx}{\linewidth}{@{}lr*{3}{Y}rrr@{}}\toprule
Metric & \makecell{Trials\\per run} & Run 1 & Run 2 & Run 3 & $\pi_{0.5}$ (\%) & \method{} (\%) & Gain (pp) \\
\midrule
Camera & 1,599 & 1,241 & 1,231 & 1,231 & $77.19\pm0.36$ & $89.91\pm0.07$ & +12.72 \\
Robot init & 1,550 & 1,184 & 1,213 & 1,212 & $77.61\pm1.06$ & $90.67\pm0.29$ & +13.05 \\
Layout & 1,525 & 1,287 & 1,292 & 1,280 & $84.35\pm0.40$ & $87.41\pm0.26$ & +3.06 \\
Light & 1,142 & 1,101 & 1,094 & 1,091 & $95.91\pm0.45$ & $96.64\pm0.31$ & +0.73 \\
BG & 1,076 & 1,019 & 1,024 & 1,020 & $94.89\pm0.25$ & $98.39\pm0.14$ & +3.50 \\
Lang & 1,537 & 1,281 & 1,315 & 1,308 & $84.67\pm1.17$ & $89.00\pm0.07$ & +4.34 \\
Noise & 1,601 & 1,312 & 1,346 & 1,332 & $83.07\pm1.07$ & $85.80\pm0.22$ & +2.73 \\
\midrule
Geo Avg & --- & 79.464 & 79.988 & 79.705 & $79.72\pm0.26$ & $89.33\pm0.04$ & +9.61 \\
Overall & 10,030 & 8,425 & 8,515 & 8,474 & $84.46\pm0.45$ & $90.53\pm0.03$ & +6.07 \\
\bottomrule\end{tabularx}\end{center}
Against $\pi_{0.5}$, the largest gains for \method{} are on Robot init (13.05 points) and Camera (12.72 points).

\begin{center}\small
\setlength{\tabcolsep}{4pt}
\begin{tabularx}{\linewidth}{@{}lr*{3}{Y}rrr@{}}\toprule
Metric & \makecell{Trials\\per run} & Run 1 & Run 2 & Run 3 & Cosmos Policy (\%) & \method{} (\%) & Gain (pp) \\
\midrule
Camera & 1,599 & 1,216 & 1,199 & 1,203 & $75.42\pm0.56$ & $89.91\pm0.07$ & +14.49 \\
Robot init & 1,550 & 1,002 & 992 & 989 & $64.15\pm0.44$ & $90.67\pm0.29$ & +26.52 \\
Layout & 1,525 & 1,238 & 1,266 & 1,232 & $81.66\pm1.19$ & $87.41\pm0.26$ & +5.75 \\
Light & 1,142 & 1,104 & 1,098 & 1,100 & $96.38\pm0.27$ & $96.64\pm0.31$ & +0.26 \\
BG & 1,076 & 959 & 961 & 933 & $88.38\pm1.45$ & $98.39\pm0.14$ & +10.01 \\
Lang & 1,537 & 1,245 & 1,270 & 1,248 & $81.61\pm0.89$ & $89.00\pm0.07$ & +7.40 \\
Noise & 1,601 & 1,475 & 1,477 & 1,485 & $92.38\pm0.33$ & $85.80\pm0.22$ & $-6.58$ \\
\midrule
Geo Avg & --- & 73.958 & 74.000 & 73.276 & $73.74\pm0.41$ & $89.33\pm0.04$ & +15.58 \\
Overall & 10,030 & 8,239 & 8,263 & 8,190 & $82.06\pm0.37$ & $90.53\pm0.03$ & +8.47 \\
\bottomrule\end{tabularx}\end{center}
Across the three runs, Cosmos Policy succeeded in 24,692 of 30,090 trials; \method{} succeeded in 27,240 of 30,090 across its three seeds. The geometric advantage of \method{} is largest on Robot init (26.52 points), followed by Camera (14.49) and Layout (5.75); Cosmos Policy has the higher success rate on Noise (92.38\% vs.\ 85.80\%).

\noindent\textbf{LIBERO-Object instruction rewrites.} The four public checkpoints in Tab.~\ref{tab:baseline-artifacts} are evaluated locally on LIBERO-Object, with $500$ original-instruction and $500$ separate rewritten-instruction trials per checkpoint. Re-target ranges from $32.4\%$ to $53.4\%$, and Original only from $29.0\%$ to $40.2\%$. Cosmos Policy has the highest Re-target rate ($53.4\%$), while $\pi_{0.5}$ has the lowest Original only rate ($29.0\%$). Table~\ref{tab:controls}B uses Cosmos Policy as its public reference; all our readout variants in that panel exceed its Re-target rate. The outcome categories and our matched readout controls are defined in Supp.~\ref{app:retarget}.

\begin{table}[h]
\centering
\caption{Public checkpoints and local LIBERO-Object evaluation (\%). Nominal uses $500$ original-instruction LIBERO-Object trials per policy. Re-target and Original only use the separate $500$ rewritten-instruction trials. \textbf{Bold}/\underline{underline}: best/next distinct by column, with ties sharing rank.}
\label{tab:baseline-artifacts}
\small
\setlength{\tabcolsep}{4pt}
\renewcommand{\arraystretch}{1.12}
\begin{tabularx}{\textwidth}{@{}l >{\raggedright\arraybackslash}X >{\raggedright\arraybackslash}X@{}}
\toprule
Method & Implementation & LIBERO model \\
\midrule
$\pi_{0.5}$ & \nolinkurl{Physical-Intelligence/openpi} & \nolinkurl{pi05_libero} \\
$\pi_0$-FAST & \href{https://github.com/Physical-Intelligence/openpi/blob/e4580662a60e528ee8e39e7073f82affec7dd407/scripts/serve_policy.py}{\nolinkurl{Physical-Intelligence/openpi}} & \nolinkurl{pi0_fast_libero} \\
VLA-JEPA & \nolinkurl{ginwind/VLA-JEPA} & \nolinkurl{ginwind/VLA-JEPA}, LIBERO folder \\
Cosmos Policy & \nolinkurl{NVlabs/cosmos-policy} & \nolinkurl{nvidia/Cosmos-Policy-LIBERO-Predict2-2B} \\
\bottomrule
\end{tabularx}\par\smallskip
\begin{tabularx}{\linewidth}{@{}l*{3}{Y}@{}}
\toprule
Checkpoint & Nominal $\uparrow$ & Re-target $\uparrow$ & Original only $\downarrow$\\
\midrule
$\pi_{0.5}$~\citep{pi05} & \textbf{98.0} & \underline{52.2} & \textbf{29.0} \\
$\pi_0$-FAST~\citep{pi0fast} & 96.6 & 38.4 & 38.0 \\
VLA-JEPA~\citep{vlajepa} & 96.4 & 32.4 & 40.2 \\
Cosmos Policy~\citep{cosmospolicy} & \underline{97.2} & \textbf{53.4} & \underline{30.6} \\
\bottomrule
\end{tabularx}
\end{table}

\begingroup
\setlength{\intextsep}{3pt}
\subsection{Closed-loop deployment rollouts and camera observations}
\label{app:sim-deployments}
Figures~\ref{fig:sim-deploy-spatial}--\ref{fig:sim-deploy-plus-texture-noise} show closed-loop \method{} rollouts in LIBERO and LIBERO-Plus~\citep{libero,liberoplus} and on WidowX. Each sequence contains observations acquired during policy execution, with columns following one episode from left to right. Quantitative benchmark results are reported in Tabs.~\ref{tab:libero-simpler} and~\ref{tab:libero-plus}.
\begin{figure}[H]
  \centering
  \includegraphics[page=1,width=\linewidth]{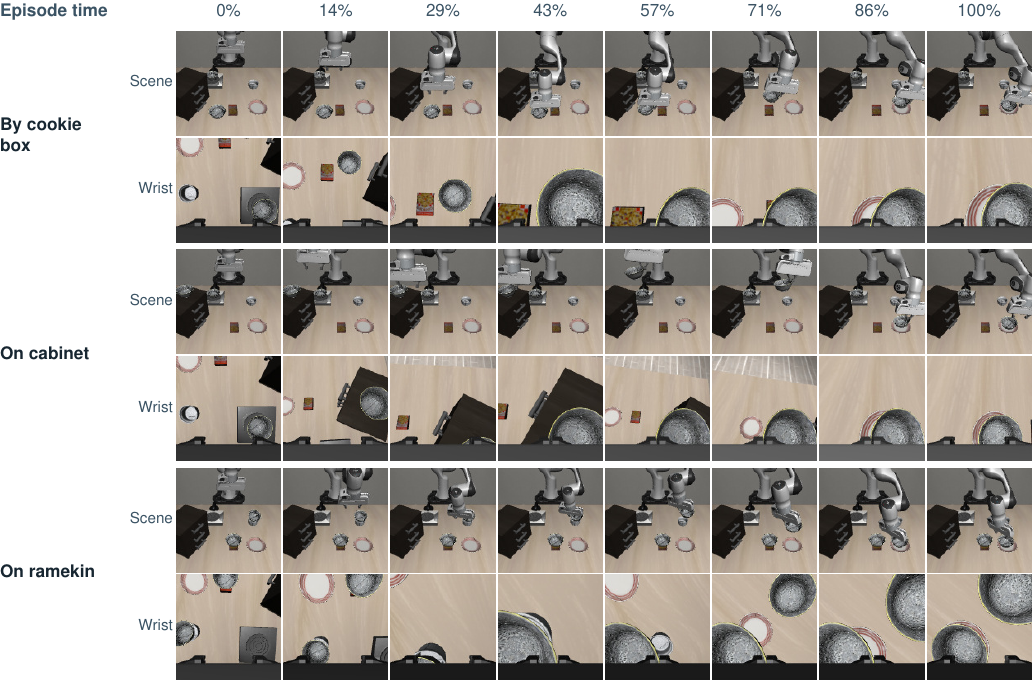}
  \caption{\textbf{LIBERO-Spatial: choosing the named bowl.} All three instructions ask for a black bowl to be placed on the plate. The bowl's starting position identifies the target: beside the cookie box, on the wooden cabinet, or on the ramekin (top to bottom). Each pair shows the scene view above and the wrist view below at the same eight instants.}
  \label{fig:sim-deploy-spatial}
\end{figure}

\vspace{9pt}

\begin{figure}[H]
  \centering
  \includegraphics[page=2,width=\linewidth]{figures/sim_sup_panels.pdf}
  \caption{\textbf{LIBERO-Object: changing the target object.} The basket is the destination in all three rollouts. From top to bottom, the requested object is the ketchup bottle, the cream cheese box and the orange juice carton. Each scene-camera strip is paired with the wrist view from the same episode.}
  \label{fig:sim-deploy-object}
\end{figure}

\vspace{9pt}

\begin{figure}[H]
  \centering
  \includegraphics[page=3,width=\linewidth]{figures/sim_sup_panels.pdf}
  \caption{\textbf{LIBERO-Goal: changing the required end state.} Top to bottom: put the wine bottle on the rack, put the bowl on top of the cabinet, and put the cream cheese in the bowl. The scene and wrist views in each pair show the same episode; the named destination differs across pairs.}
  \label{fig:sim-deploy-goal}
\end{figure}

\vspace{9pt}

\begin{figure}[H]
  \centering
  \includegraphics[page=4,width=\linewidth]{figures/sim_sup_panels.pdf}
  \caption{\textbf{LIBERO-Long: the caddy, stove and basket tasks.} Each row follows one episode from the scene camera. Top to bottom: place a book in the rear compartment of the caddy; put both moka pots on the stove; put two specified groceries in the basket. The grocery pairs in the last three rows are alphabet soup and cream cheese, alphabet soup and tomato sauce, and cream cheese and butter.}
  \label{fig:sim-deploy-long-1}
\end{figure}

\vspace{9pt}

\begin{figure}[H]
  \centering
  \includegraphics[page=5,width=\linewidth]{figures/sim_sup_panels.pdf}
  \caption{\textbf{LIBERO-Long: ordered actions and distinct destinations.} Top to bottom: put the black bowl in the bottom drawer and close it; put the white mug on the left plate and the yellow-and-white mug on the right plate; put the white mug on the plate and the pudding to its right; put the yellow-and-white mug in the microwave and close it; turn on the stove and place the moka pot on it. Each row shows a separate rollout from the scene camera.}
  \label{fig:sim-deploy-long-2}
\end{figure}

\vspace{9pt}

\begin{figure}[H]
  \centering
  \includegraphics[page=6,width=\linewidth]{figures/sim_sup_panels.pdf}
  \caption{\textbf{WidowX: closed-loop \method{} rollouts.} The first two rows show separate carrot-to-plate episodes. The next three show spoon-to-towel placement, stacking the green block on the yellow block, and placing an eggplant in a pot. Each row follows one policy rollout. The eggplant-to-pot rollout shows a task outside the benchmark; Tab.~\ref{tab:libero-simpler} evaluates eggplant-to-basket placement.}
  \label{fig:sim-deploy-widowx}
\end{figure}

\vspace{9pt}

\begin{figure}[H]
  \centering
  \includegraphics[page=7,width=\linewidth]{figures/sim_sup_panels.pdf}
  \caption{\textbf{LIBERO-Plus: camera viewpoint and lighting.} The instructions, top to bottom, are to put the bowl on the plate, put the wine bottle on the rack and turn on the stove. For each instruction, the upper strip shows a viewpoint change and the lower strip a lighting change. The two conditions are separate episodes, each sampled on its own time scale.}
  \label{fig:sim-deploy-plus-view-light}
\end{figure}

\vspace{9pt}

\begin{figure}[H]
  \centering
  \includegraphics[page=8,width=\linewidth]{figures/sim_sup_panels.pdf}
  \caption{\textbf{LIBERO-Plus: background texture and sensor noise.} The same three instructions are shown in the same order: bowl to plate, bottle to rack, and stove on. In each pair, the upper strip shows a background-texture change and the lower strip sensor noise. The strips come from separate episodes; together with the preceding figure they cover four of the seven perturbations in Tab.~\ref{tab:libero-plus}.}
  \label{fig:sim-deploy-plus-texture-noise}
\end{figure}

\endgroup

\FloatBarrier
\section{Ablations and evidence for object addressing}\label{app:ablations}

\subsection{Trunk addressing, object inputs and training controls}

The training and inference controls in Tabs.~\ref{tab:abl-train} and~\ref{tab:abl-infer} use the three training seeds and benchmark protocol of Supp.~\ref{app:eval}, with the dual-view model (b) of Tab.~\ref{tab:controls}A as their reference. Each seed has $1{,}500$ standard LIBERO trials per suite and $10{,}030$ trials on full LIBERO-Plus. Geo Avg weights Camera, Robot init and Layout equally, and Overall pools successes across all seven axes within each seed. Re-target and Multi-inst.\ are the rewritten-instruction and same-class-distractor tests of Supp.~\ref{app:retarget}. Within each seed, all models use the same initial states. All training variants use 100k Stage-II updates. Rows marked II in Tab.~\ref{tab:abl-train} start from the reference Stage-I checkpoint, except that A2 skips Stage~I and A4 starts from model~(a)'s Stage-I checkpoint. Rows marked I+II retrain both stages.

\noindent\textbf{Addressing controls (Tab.~\ref{tab:controls}A).} Addressing, not slots, drives robustness. Model (b) retains (a)'s perception, readout, route dropout, parameter count and training budget, and adds address-only trunk keys and per-layer address writes. Adding object slots, a pointer and a world head without OA gives lower mean Geo Avg than the slot-free policy (h) (78.6\% vs.\ 79.7\%) and lowers sensor-noise success from 83.4\% to 73.4\%. Addressing the same slots by identity raises Geo Avg to 89.3\% and Noise to 85.8\%, and improves every LIBERO suite and LIBERO-Plus axis in all three seeds (Tab.~\ref{tab:per-seed}). The external-camera policies (a$'$)/(b$'$) each reuse their corresponding Stage-I initialization and omit wrist images and wrist pose measurements throughout Stage~II and evaluation. Under this protocol, trunk addressing raises Geo Avg from 78.8\% to 82.3\% and improves each geometric axis in every seed.

\noindent\textbf{Trunk addressing (K1).} Both trunk operations contribute, and they are complementary. We vary them after OA Stage-I alignment, retaining the identity-key action pointer (Tab.~\ref{tab:abl-train}). V2 disables both operations. V1 retains only the per-layer address writes and uses standard keys. V3 retains the address key path and writes the address once, before the first layer. Each operation alone improves the three-seed means of all six summary metrics over V2. Geo Avg rises from 84.5\% (V2) to 86.7\% (V1) and 86.9\% (V3), and to 89.3\% with both; Overall follows the same order (84.6\%, 87.3\%, 87.6\% and 90.5\%). The full trunk's largest gain over V2 is under sensor noise (67.6\% vs.\ 85.8\%). In every seed, either operation alone improves Re-target and Multi-inst.\ over V2, and the full trunk improves both further. These results follow the keys themselves (Supp.~\ref{app:key-invariance}): only the full design keeps object-slot keys invariant at every layer.

The remaining K1 controls favor the default key input, address dimension and layer coverage: in each comparison, the default has the highest three-seed mean on all six summary metrics. Normalizing the address coordinates on their own, which makes the key exactly invariant (Supp.~\ref{app:oa-impl}), outperforms masking the normalized full residual, with Geo Avg 89.3\% vs.\ 83.7\%, Re-target 77.4\% vs.\ 72.8\% and Multi-inst.\ 80.0\% vs.\ 74.5\%. Address dimensions of 16, 32 and 64 give 82.9\%, 89.3\% and 87.0\% Geo Avg, 72.4\%, 77.4\% and 76.6\% Re-target, and 73.5\%, 80.0\% and 79.2\% Multi-inst. Applying OA in all 32 layers outperforms restricting it to the first or last 16 (Geo Avg 89.3\% vs.\ 86.8\% and 86.9\%; Noise 85.8\% vs.\ 63.0\% and 69.5\%).

OA benefits both training stages. From the Stage-0 trunk, Geo Avg is 78.6\% without OA (model (a)), 84.5\% with OA only during Stage-I alignment (V2), 85.7\% with OA only in Stage~II (A4) and 89.3\% with OA in both. Starting instead from public Chameleon-7B weights with the same 32 address coordinates, OA raises Geo Avg from 76.4\% to 85.5\%, Overall from 80.6\% to 87.6\% and Re-target from 66.6\% to 73.6\% (A5, A6), so its advantage holds without the robot-adapted Stage-0 weights.

\noindent\textbf{Address components (K3).} Each address component contributes, and the two descriptive cues are complementary. Both single-cue controls retain the identity code. With the code and label, Re-target reaches 75.2\% and Multi-inst.\ 72.2\%; with the code and initial appearance, the rates are 71.1\% and 77.2\%. Combining all three components reaches 77.4\% and 80.0\%. The label therefore matters more for following a rewritten instruction ($-6.3$ points when removed), and initial appearance for separating same-class instances ($-7.8$ points). Removing the identity code lowers Multi-inst.\ to 67.8\%, the largest drop among the address controls. All three controls also lower Geo Avg (82.6\%, 85.6\% and 86.2\% vs.\ 89.3\%).

\noindent\textbf{Object context and pose (K4).} Slots for non-target objects help the policy under geometric change. Layout success rises from 71.5\% with target and robot slots to 83.9\% after adding reference slots and 87.4\% with all object slots, and Geo Avg follows the same order (79.4\%, 85.7\% and 89.3\%). A cap of sixteen object slots gives higher Geo Avg (89.3\% vs.\ 85.2\%), Overall (90.5\% vs.\ 86.6\%) and Noise success (85.8\% vs.\ 74.5\%) than a cap of eight, which gives higher Re-target (81.8\% vs.\ 77.4\%). In every seed, including pose in the slots improves Camera, Robot init and Layout over pose-free slots, by 3.5, 4.5 and 8.9 points on average, while pose-free slots give similar target selection (Re-target 78.5\% vs.\ 77.4\%; Multi-inst.\ 80.7\% vs.\ 80.0\%). The default content excludes image-plane mask shape: adding it raises Re-target from 77.4\% to 85.2\% in every seed and leaves Multi-inst.\ similar (81.2\% vs.\ 80.0\%), but lowers Noise success from 85.8\% to 66.8\% and Geo Avg to 86.9\%.

\noindent\textbf{Training objectives and stages (K8).} Composition consistency improves the means of all six summary metrics, with the largest mean gain on the same-class distractor test: it raises Multi-inst.\ from 74.0\% to 80.0\%, Re-target from 75.6\% to 77.4\%, Geo Avg from 88.4\% to 89.3\% and standard LIBERO from 98.7\% to 99.0\% (Tab.~\ref{tab:abl-train}). Without the loss, Multi-inst.\ varies widely across seeds, from 68.5\% to 81.5\% (SD 6.73 points), and exceeds the reference in one seed. At the same 100k Stage-II budget, removing Stage-I alignment lowers Geo Avg from 89.3\% to 84.6\% and Multi-inst.\ from 80.0\% to 74.5\%. LoRA gives similar mean success to full fine-tuning (89.3\% vs.\ 89.4\% Geo Avg; 90.5\% Overall for both) while training 128.6M parameters.

\noindent\textbf{History and re-planning (K9).} More history improves success under sensor noise and on long-horizon tasks: with one, two and four input frames, Noise success is 81.5\%, 84.5\% and 85.8\%, and Long is 95.8\%, 97.9\% and 98.4\%. The interval comparison holds history length at four frames. For re-planning intervals $K=8,16,32$, the predicted horizons $H$ are $16,16,32$; history spacing is $K$, and world and image targets come from $s+K$. These settings give 90.6\%, 90.5\% and 88.4\% Overall. The default interval of 16 thus matches interval 8 to within 0.1 points at half the query frequency; interval 32, which also doubles the predicted horizon, lowers Overall by 2.1 points.

\begin{table*}[!ht]
\centering
\caption{Training ablations with 100k Stage-II updates per seed. Stage identifies the retraining required. Entries give the six summary metrics as mean success rates (\%) across three training seeds, with sample SD. Re-target and Multi-inst.\ use 500 and 200 episodes per seed. Overall pools the seven axes within each seed, and its mean and sample SD are computed across the three seed-level rates. Table~\ref{tab:readout-seeds} gives per-seed Re-target outcomes for four readout variants.}
\label{tab:abl-train}
\scriptsize\setlength{\tabcolsep}{2pt}
\begin{tabularx}{\linewidth}{@{}Xc*{6}{r}@{}}
\toprule
Variant & Stage & LIBERO & Geo Avg & Noise & Overall & Re-target & Multi-inst. \\
\midrule
\hl{Reference: model (b)} & \hl{II} & \hl{99.0\statstd{0.03}} & \hl{89.3\statstd{0.04}} & \hl{85.8\statstd{0.22}} & \hl{90.5\statstd{0.03}} & \hl{77.4\statstd{0.72}} & \hl{80.0\statstd{0.50}} \\
\midrule
\multicolumn{8}{@{}l}{\textit{K1 OA in the trunk}} \\
V2: No OA in Stage~II & II & 98.4\statstd{0.10} & 84.5\statstd{0.59} & 67.6\statstd{1.05} & 84.6\statstd{0.07} & 71.8\statstd{1.44} & 74.0\statstd{1.00} \\
V1: Key path off, address writes kept & II & 98.6\statstd{0.05} & 86.7\statstd{0.35} & 73.6\statstd{1.77} & 87.3\statstd{0.30} & 75.0\statstd{0.60} & 76.8\statstd{1.61} \\
V3: Address written before the first layer only & II & 98.8\statstd{0.12} & 86.9\statstd{0.34} & 77.0\statstd{1.13} & 87.6\statstd{0.25} & 74.3\statstd{0.76} & 77.8\statstd{1.15} \\
Masked normalized key input & II & 98.7\statstd{0.13} & 83.7\statstd{0.37} & 71.2\statstd{0.19} & 84.4\statstd{0.12} & 72.8\statstd{2.50} & 74.5\statstd{1.00} \\
Address dimension $16$ & I+II & 98.2\statstd{0.17} & 82.9\statstd{0.73} & 70.6\statstd{0.90} & 85.1\statstd{0.39} & 72.4\statstd{2.91} & 73.5\statstd{3.00} \\
Address dimension $64$ & I+II & 98.3\statstd{0.18} & 87.0\statstd{0.54} & 78.7\statstd{0.80} & 88.3\statstd{0.51} & 76.6\statstd{0.80} & 79.2\statstd{1.26} \\
OA in the last $16$ layers only & II & 98.1\statstd{0.27} & 86.9\statstd{0.62} & 69.5\statstd{0.34} & 86.5\statstd{0.15} & 75.0\statstd{1.22} & 77.5\statstd{2.65} \\
OA in the first $16$ layers only & II & 98.3\statstd{0.11} & 86.8\statstd{0.32} & 63.0\statstd{0.72} & 84.7\statstd{0.40} & 74.8\statstd{1.91} & 76.8\statstd{1.53} \\
\multicolumn{8}{@{}l}{\textit{K2 Readout}} \\
Mean-pool readout (role loss off) & II & 98.0\statstd{0.12} & 86.9\statstd{0.05} & 80.3\statstd{0.23} & 88.1\statstd{0.08} & 56.9\statstd{2.73} & 70.7\statstd{4.48} \\
No readout (\tok{[act\_q]} only) & II & 98.4\statstd{0.19} & 85.9\statstd{0.17} & 78.1\statstd{0.12} & 87.3\statstd{0.09} & 69.8\statstd{1.59} & 72.3\statstd{2.52} \\
No OA and mean-pool readout (non-addressable) & II & 97.9\statstd{0.10} & 83.9\statstd{0.11} & 78.9\statstd{0.06} & 85.9\statstd{0.11} & 54.5\statstd{3.43} & 68.3\statstd{1.26} \\
$R{=}1$ role query & II & 99.0\statstd{0.08} & 87.6\statstd{0.10} & 82.7\statstd{0.16} & 89.1\statstd{0.08} & 72.5\statstd{1.55} & 74.3\statstd{2.75} \\
$R{=}8$ role queries & II & 98.9\statstd{0.10} & 87.8\statstd{0.17} & 82.6\statstd{0.13} & 89.2\statstd{0.10} & 75.9\statstd{2.77} & 77.7\statstd{5.03} \\
$\lambda_r{=}0$ & II & 98.8\statstd{0.08} & 88.0\statstd{0.13} & 81.4\statstd{0.07} & 89.0\statstd{0.08} & 64.6\statstd{1.04} & 73.8\statstd{2.75} \\
$\lambda_r{=}0.5$ & II & 99.0\statstd{0.15} & 88.1\statstd{0.06} & 84.3\statstd{0.06} & 89.6\statstd{0.04} & 77.4\statstd{1.56} & 77.8\statstd{1.26} \\
\multicolumn{8}{@{}l}{\textit{K3 Address}} \\
No identity codes & I+II & 98.5\statstd{0.07} & 86.2\statstd{0.28} & 74.4\statstd{0.47} & 87.2\statstd{0.26} & 72.1\statstd{3.52} & 67.8\statstd{3.62} \\
Identity code + label & II & 98.1\statstd{0.09} & 82.6\statstd{0.18} & 69.9\statstd{2.34} & 84.2\statstd{0.35} & 75.2\statstd{1.73} & 72.2\statstd{5.03} \\
Identity code + initial appearance & II & 98.8\statstd{0.11} & 85.6\statstd{0.61} & 61.3\statstd{1.47} & 84.5\statstd{0.45} & 71.1\statstd{0.90} & 77.2\statstd{4.51} \\
\multicolumn{8}{@{}l}{\textit{K4 Object representation}} \\
Target and robot slots only & II & 98.2\statstd{0.07} & 79.4\statstd{0.73} & 69.9\statstd{1.20} & 82.7\statstd{0.20} & 70.3\statstd{2.72} & 65.2\statstd{2.57} \\
Target, reference and robot slots only & II & 98.8\statstd{0.09} & 85.7\statstd{0.43} & 67.7\statstd{0.95} & 85.4\statstd{0.44} & 77.1\statstd{1.33} & 78.5\statstd{0.50} \\
At most $8$ object slots & II & 98.4\statstd{0.14} & 85.2\statstd{0.44} & 74.5\statstd{1.44} & 86.6\statstd{0.17} & 81.8\statstd{2.60} & 78.8\statstd{3.75} \\
No pose in object slots & II & 98.7\statstd{0.07} & 83.7\statstd{0.35} & 75.7\statstd{1.64} & 85.6\statstd{0.31} & 78.5\statstd{1.40} & 80.7\statstd{1.61} \\
Image-plane mask shape in the content input & I+II & 98.4\statstd{0.16} & 86.9\statstd{0.42} & 66.8\statstd{0.89} & 85.5\statstd{0.27} & 85.2\statstd{1.06} & 81.2\statstd{2.52} \\
\multicolumn{8}{@{}l}{\textit{K5 Image route}} \\
No route dropout & II & 98.8\statstd{0.20} & 87.4\statstd{0.79} & 83.5\statstd{1.16} & 89.1\statstd{0.65} & 74.8\statstd{0.53} & 73.7\statstd{3.82} \\
\multicolumn{8}{@{}l}{\textit{K6 World model}} \\
No world loss or residual trigger (disappearance-only \mbox{re-grounding}) & II & 98.7\statstd{0.13} & 83.1\statstd{0.23} & 77.8\statstd{0.85} & 85.7\statstd{0.20} & 70.3\statstd{2.72} & 71.8\statstd{1.89} \\
No auxiliary motion monitor & I+II & 99.0\statstd{0.17} & 89.4\statstd{0.70} & 85.9\statstd{1.00} & 90.6\statstd{0.59} & 77.4\statstd{2.12} & 79.7\statstd{1.89} \\
Content targets without EMA & II & 98.7\statstd{0.23} & 87.7\statstd{0.50} & 83.8\statstd{1.42} & 89.3\statstd{0.71} & 73.8\statstd{1.93} & 75.0\statstd{4.27} \\
No auxiliary VQ loss & II & 98.9\statstd{0.14} & 89.0\statstd{0.48} & 85.4\statstd{0.41} & 90.3\statstd{0.27} & 77.2\statstd{0.60} & 79.3\statstd{4.04} \\
\multicolumn{8}{@{}l}{\textit{K7 Perception}} \\
O1: Simulator poses in training and evaluation (oracle) & I+II & 99.3\statstd{0.06} & 93.3\statstd{0.27} & 89.5\statstd{0.70} & 93.1\statstd{0.33} & 78.5\statstd{1.30} & 81.8\statstd{3.75} \\
\multicolumn{8}{@{}l}{\textit{K8 Training recipe}} \\
A1: No composition loss & II & 98.7\statstd{0.05} & 88.4\statstd{0.11} & 84.2\statstd{1.34} & 89.7\statstd{0.32} & 75.6\statstd{1.25} & 74.0\statstd{6.73} \\
A2: No Stage~I & II & 97.5\statstd{0.15} & 84.6\statstd{0.17} & 79.2\statstd{1.39} & 86.5\statstd{0.26} & 71.4\statstd{1.51} & 74.5\statstd{1.73} \\
A3: Full fine-tuning instead of LoRA & II & 99.1\statstd{0.12} & 89.4\statstd{0.38} & 85.7\statstd{0.50} & 90.5\statstd{0.21} & 77.4\statstd{2.09} & 80.0\statstd{4.77} \\
A4: Stage~I of (a), OA in Stage~II only & II & 98.2\statstd{0.20} & 85.7\statstd{1.19} & 81.5\statstd{0.63} & 87.7\statstd{0.49} & 73.6\statstd{0.35} & 77.0\statstd{3.00} \\
A5: No Stage~0, model (a) from Chameleon-7B & I+II & 96.3\statstd{0.20} & 76.4\statstd{0.89} & 70.8\statstd{0.19} & 80.6\statstd{0.58} & 66.6\statstd{2.09} & 69.0\statstd{1.32} \\
A6: No Stage~0, model (b) from Chameleon-7B & I+II & 97.9\statstd{0.08} & 85.5\statstd{0.22} & 82.1\statstd{1.12} & 87.6\statstd{0.22} & 73.6\statstd{2.16} & 76.0\statstd{2.18} \\
\multicolumn{8}{@{}l}{\textit{K9 Sensitivity}} \\
B1: $T{=}1$ history frame & II & 97.9\statstd{0.08} & 86.4\statstd{0.61} & 81.5\statstd{0.67} & 88.1\statstd{0.57} & 74.2\statstd{1.56} & 76.5\statstd{3.77} \\
B2: $T{=}2$ history frames & II & 98.9\statstd{0.13} & 88.9\statstd{0.31} & 84.5\statstd{0.83} & 90.1\statstd{0.30} & 76.6\statstd{1.80} & 79.0\statstd{3.97} \\
B3: Re-planning interval $8$ & II & 99.0\statstd{0.06} & 89.5\statstd{0.50} & 85.9\statstd{0.36} & 90.6\statstd{0.36} & 77.6\statstd{2.27} & 80.5\statstd{3.28} \\
B4: Re-planning interval $32$ & II & 98.2\statstd{0.21} & 86.8\statstd{0.28} & 82.2\statstd{0.04} & 88.4\statstd{0.18} & 74.6\statstd{0.69} & 77.0\statstd{3.91} \\
\bottomrule
\end{tabularx}
\end{table*}

\begin{table*}[!ht]
\centering
\caption*{\textbf{Table~\ref{tab:abl-train} (continued).} Mean success rates (\%) by standard LIBERO suite and full LIBERO-Plus axis for the same three training seeds, with sample SD.}
\scriptsize\setlength{\tabcolsep}{2.3pt}
\begin{tabularx}{\linewidth}{@{}X*{11}{r}@{}}
\toprule
 & \multicolumn{4}{c}{Standard LIBERO} & \multicolumn{7}{c}{Full LIBERO-Plus} \\
\cmidrule(lr){2-5}\cmidrule(lr){6-12}
Variant & Spatial & Object & Goal & Long & Cam & Init & Layout & Light & BG & Lang & Noise \\
\midrule
Reference: model (b) & 99.0\statstd{0.13} & 99.9\statstd{0.04} & 98.8\statstd{0.17} & 98.4\statstd{0.14} & 89.9\statstd{0.07} & 90.7\statstd{0.29} & 87.4\statstd{0.26} & 96.6\statstd{0.31} & 98.4\statstd{0.14} & 89.0\statstd{0.07} & 85.8\statstd{0.22} \\
\midrule
\multicolumn{12}{@{}l}{\textit{K1 OA in the trunk}} \\
V2: No OA in Stage~II & 98.8\statstd{0.12} & 99.0\statstd{0.20} & 98.5\statstd{0.17} & 97.2\statstd{0.18} & 86.0\statstd{0.90} & 88.4\statstd{0.26} & 79.0\statstd{0.64} & 93.5\statstd{0.27} & 98.2\statstd{0.14} & 86.3\statstd{0.85} & 67.6\statstd{1.05} \\
V1: Key path off, address writes kept & 98.9\statstd{0.29} & 99.3\statstd{0.31} & 98.6\statstd{0.28} & 97.4\statstd{0.17} & 89.8\statstd{0.71} & 90.4\statstd{1.07} & 79.9\statstd{0.72} & 96.5\statstd{0.30} & 98.3\statstd{0.53} & 88.7\statstd{0.55} & 73.6\statstd{1.77} \\
V3: Address written before the first layer only & 98.9\statstd{0.20} & 99.7\statstd{0.10} & 98.7\statstd{0.14} & 97.9\statstd{0.24} & 88.8\statstd{0.25} & 88.1\statstd{0.32} & 83.9\statstd{0.56} & 94.8\statstd{0.41} & 98.3\statstd{0.23} & 87.9\statstd{0.23} & 77.0\statstd{1.13} \\
Masked normalized key input & 98.9\statstd{0.24} & 99.4\statstd{0.17} & 98.3\statstd{0.27} & 98.2\statstd{0.27} & 83.2\statstd{0.28} & 85.4\statstd{0.32} & 82.6\statstd{0.82} & 96.1\statstd{0.22} & 97.7\statstd{0.33} & 82.2\statstd{0.79} & 71.2\statstd{0.19} \\
Address dimension $16$ & 98.5\statstd{0.53} & 99.4\statstd{0.08} & 97.8\statstd{0.28} & 96.9\statstd{0.37} & 85.9\statstd{0.16} & 84.0\statstd{1.22} & 78.7\statstd{0.86} & 96.5\statstd{0.40} & 98.3\statstd{0.51} & 88.9\statstd{0.23} & 70.6\statstd{0.90} \\
Address dimension $64$ & 98.4\statstd{0.31} & 99.5\statstd{0.18} & 98.1\statstd{0.27} & 97.3\statstd{0.17} & 89.7\statstd{0.25} & 90.6\statstd{1.10} & 80.8\statstd{0.52} & 96.5\statstd{0.61} & 98.3\statstd{0.19} & 88.8\statstd{0.90} & 78.7\statstd{0.80} \\
OA in the last $16$ layers only & 98.3\statstd{0.24} & 99.6\statstd{0.14} & 98.5\statstd{0.25} & 96.1\statstd{0.76} & 89.8\statstd{0.82} & 90.5\statstd{0.56} & 80.5\statstd{0.73} & 95.8\statstd{0.44} & 98.3\statstd{0.77} & 87.5\statstd{0.26} & 69.5\statstd{0.34} \\
OA in the first $16$ layers only & 99.0\statstd{0.10} & 99.4\statstd{0.28} & 97.9\statstd{0.18} & 96.8\statstd{0.31} & 88.7\statstd{1.27} & 89.5\statstd{0.91} & 82.1\statstd{0.60} & 93.9\statstd{0.89} & 98.2\statstd{0.30} & 84.7\statstd{0.92} & 63.0\statstd{0.72} \\
\multicolumn{12}{@{}l}{\textit{K2 Readout}} \\
Mean-pool readout (role loss off) & 98.7\statstd{0.21} & 99.3\statstd{0.18} & 97.6\statstd{0.28} & 96.4\statstd{0.15} & 88.2\statstd{0.13} & 88.4\statstd{0.23} & 84.1\statstd{0.13} & 95.6\statstd{0.15} & 98.1\statstd{0.14} & 87.2\statstd{0.23} & 80.3\statstd{0.23} \\
No readout (\tok{[act\_q]} only) & 98.7\statstd{0.29} & 99.5\statstd{0.20} & 98.1\statstd{0.25} & 97.3\statstd{0.17} & 87.7\statstd{0.11} & 87.8\statstd{0.16} & 82.2\statstd{0.27} & 95.2\statstd{0.13} & 97.8\statstd{0.23} & 87.5\statstd{0.26} & 78.1\statstd{0.12} \\
No OA and mean-pool readout (non-addressable) & 98.4\statstd{0.04} & 99.1\statstd{0.25} & 97.7\statstd{0.20} & 96.4\statstd{0.23} & 85.6\statstd{0.14} & 85.4\statstd{0.29} & 80.7\statstd{0.19} & 94.0\statstd{0.13} & 97.6\statstd{0.25} & 84.9\statstd{0.26} & 78.9\statstd{0.06} \\
$R{=}1$ role query & 98.9\statstd{0.08} & 99.8\statstd{0.18} & 98.8\statstd{0.23} & 98.3\statstd{0.24} & 89.0\statstd{0.29} & 89.2\statstd{0.21} & 84.6\statstd{0.11} & 96.2\statstd{0.27} & 98.2\statstd{0.27} & 88.8\statstd{0.30} & 82.7\statstd{0.16} \\
$R{=}8$ role queries & 98.8\statstd{0.27} & 99.8\statstd{0.24} & 98.7\statstd{0.13} & 98.3\statstd{0.24} & 89.2\statstd{0.22} & 89.7\statstd{0.16} & 84.5\statstd{0.20} & 96.0\statstd{0.13} & 98.2\statstd{0.25} & 88.7\statstd{0.28} & 82.6\statstd{0.13} \\
$\lambda_r{=}0$ & 98.8\statstd{0.07} & 99.6\statstd{0.17} & 98.6\statstd{0.20} & 98.2\statstd{0.27} & 89.1\statstd{0.19} & 89.3\statstd{0.28} & 85.6\statstd{0.17} & 96.1\statstd{0.20} & 98.2\statstd{0.19} & 88.4\statstd{0.30} & 81.4\statstd{0.07} \\
$\lambda_r{=}0.5$ & 98.9\statstd{0.28} & 99.8\statstd{0.18} & 98.7\statstd{0.18} & 98.4\statstd{0.27} & 89.4\statstd{0.25} & 89.6\statstd{0.19} & 85.3\statstd{0.17} & 96.3\statstd{0.18} & 98.3\statstd{0.11} & 88.7\statstd{0.25} & 84.3\statstd{0.06} \\
\multicolumn{12}{@{}l}{\textit{K3 Address}} \\
No identity codes & 98.7\statstd{0.10} & 99.4\statstd{0.07} & 97.6\statstd{0.18} & 98.3\statstd{0.08} & 91.0\statstd{0.28} & 87.7\statstd{0.56} & 80.0\statstd{0.69} & 95.1\statstd{0.23} & 98.4\statstd{0.23} & 89.5\statstd{1.13} & 74.4\statstd{0.47} \\
Identity code + label & 98.5\statstd{0.10} & 99.1\statstd{0.19} & 98.2\statstd{0.14} & 96.5\statstd{0.10} & 83.6\statstd{1.47} & 86.0\statstd{0.06} & 78.3\statstd{1.10} & 92.4\statstd{0.41} & 97.9\statstd{0.16} & 88.1\statstd{0.90} & 69.9\statstd{2.34} \\
Identity code + initial appearance & 99.1\statstd{0.15} & 99.2\statstd{0.08} & 99.0\statstd{0.14} & 97.9\statstd{0.12} & 85.4\statstd{0.94} & 87.0\statstd{0.30} & 84.3\statstd{1.02} & 93.9\statstd{0.48} & 98.4\statstd{0.25} & 88.6\statstd{0.58} & 61.3\statstd{1.47} \\
\multicolumn{12}{@{}l}{\textit{K4 Object representation}} \\
Target and robot slots only & 98.4\statstd{0.10} & 99.0\statstd{0.10} & 98.6\statstd{0.07} & 96.6\statstd{0.12} & 87.5\statstd{0.56} & 79.3\statstd{1.41} & 71.5\statstd{0.36} & 93.7\statstd{0.84} & 98.1\statstd{0.11} & 86.9\statstd{0.71} & 69.9\statstd{1.20} \\
Target, reference and robot slots only & 99.1\statstd{0.13} & 99.8\statstd{0.12} & 98.6\statstd{0.23} & 97.5\statstd{0.24} & 87.0\statstd{1.03} & 86.1\statstd{0.49} & 83.9\statstd{0.27} & 96.5\statstd{0.32} & 98.8\statstd{0.56} & 85.2\statstd{1.03} & 67.7\statstd{0.95} \\
At most $8$ object slots & 98.6\statstd{0.12} & 99.2\statstd{0.23} & 97.9\statstd{0.50} & 97.8\statstd{0.18} & 88.3\statstd{0.27} & 87.4\statstd{0.27} & 80.0\statstd{1.04} & 96.2\statstd{0.27} & 98.6\statstd{0.56} & 87.9\statstd{0.63} & 74.5\statstd{1.44} \\
No pose in object slots & 98.6\statstd{0.18} & 99.8\statstd{0.17} & 99.3\statstd{0.10} & 97.1\statstd{0.10} & 86.4\statstd{0.72} & 86.2\statstd{0.84} & 78.5\statstd{0.55} & 94.9\statstd{1.17} & 98.3\statstd{0.51} & 86.0\statstd{0.30} & 75.7\statstd{1.64} \\
Image-plane mask shape in the content input & 98.9\statstd{0.15} & 99.6\statstd{0.17} & 98.9\statstd{0.27} & 96.2\statstd{0.19} & 89.0\statstd{0.58} & 88.6\statstd{1.00} & 83.2\statstd{0.19} & 93.4\statstd{0.94} & 98.4\statstd{0.19} & 85.9\statstd{0.51} & 66.8\statstd{0.89} \\
\multicolumn{12}{@{}l}{\textit{K5 Image route}} \\
No route dropout & 98.8\statstd{0.37} & 99.6\statstd{0.23} & 98.6\statstd{0.12} & 98.2\statstd{0.42} & 88.0\statstd{1.62} & 90.3\statstd{0.38} & 83.9\statstd{0.90} & 96.1\statstd{0.35} & 98.3\statstd{0.33} & 88.7\statstd{0.55} & 83.5\statstd{1.16} \\
\multicolumn{12}{@{}l}{\textit{K6 World model}} \\
No world loss or residual trigger (disappearance-only \mbox{re-grounding}) & 98.7\statstd{0.20} & 99.7\statstd{0.15} & 98.4\statstd{0.20} & 98.0\statstd{0.23} & 82.6\statstd{0.44} & 85.5\statstd{0.76} & 81.2\statstd{1.05} & 95.0\statstd{1.35} & 97.9\statstd{0.54} & 86.2\statstd{0.23} & 77.8\statstd{0.85} \\
No auxiliary motion monitor & 99.0\statstd{0.24} & 99.9\statstd{0.08} & 98.8\statstd{0.24} & 98.3\statstd{0.14} & 89.7\statstd{0.64} & 90.9\statstd{1.00} & 87.6\statstd{0.74} & 96.7\statstd{0.75} & 98.5\statstd{0.28} & 89.1\statstd{0.95} & 85.9\statstd{1.00} \\
Content targets without EMA & 98.7\statstd{0.50} & 99.7\statstd{0.37} & 98.4\statstd{0.61} & 98.0\statstd{0.29} & 88.5\statstd{0.72} & 89.6\statstd{0.87} & 85.0\statstd{0.10} & 96.3\statstd{0.35} & 98.0\statstd{0.60} & 88.5\statstd{1.64} & 83.8\statstd{1.42} \\
No auxiliary VQ loss & 99.0\statstd{0.29} & 99.8\statstd{0.07} & 98.7\statstd{0.17} & 98.1\statstd{0.23} & 89.6\statstd{0.69} & 90.3\statstd{0.54} & 87.1\statstd{0.57} & 96.7\statstd{0.71} & 98.2\statstd{0.39} & 88.8\statstd{0.62} & 85.4\statstd{0.41} \\
\multicolumn{12}{@{}l}{\textit{K7 Perception}} \\
O1: Simulator poses in training and evaluation (oracle) & 99.4\statstd{0.34} & 99.8\statstd{0.08} & 99.1\statstd{0.37} & 98.8\statstd{0.04} & 93.9\statstd{0.41} & 94.2\statstd{0.50} & 91.8\statstd{0.57} & 97.1\statstd{0.79} & 98.3\statstd{0.46} & 89.3\statstd{0.57} & 89.5\statstd{0.70} \\
\multicolumn{12}{@{}l}{\textit{K8 Training recipe}} \\
A1: No composition loss & 98.9\statstd{0.12} & 99.5\statstd{0.15} & 98.5\statstd{0.00} & 97.8\statstd{0.08} & 89.0\statstd{0.44} & 90.3\statstd{0.39} & 85.8\statstd{0.27} & 96.1\statstd{0.26} & 98.2\statstd{0.56} & 88.6\statstd{0.78} & 84.2\statstd{1.34} \\
A2: No Stage~I & 97.7\statstd{0.13} & 98.8\statstd{0.47} & 98.0\statstd{0.14} & 95.6\statstd{0.69} & 84.9\statstd{0.88} & 86.1\statstd{0.92} & 82.7\statstd{1.53} & 94.9\statstd{0.83} & 97.5\statstd{0.80} & 86.3\statstd{0.04} & 79.2\statstd{1.39} \\
A3: Full fine-tuning instead of LoRA & 99.1\statstd{0.10} & 99.8\statstd{0.12} & 98.9\statstd{0.34} & 98.6\statstd{0.37} & 90.0\statstd{0.54} & 90.5\statstd{0.21} & 87.6\statstd{0.66} & 96.3\statstd{0.63} & 98.4\statstd{0.16} & 89.1\statstd{0.17} & 85.7\statstd{0.50} \\
A4: Stage~I of (a), OA in Stage~II only & 98.5\statstd{0.48} & 99.2\statstd{0.30} & 98.3\statstd{0.44} & 97.0\statstd{0.44} & 86.1\statstd{0.76} & 87.5\statstd{1.88} & 83.5\statstd{0.95} & 95.6\statstd{0.31} & 98.0\statstd{0.89} & 87.4\statstd{0.15} & 81.5\statstd{0.63} \\
A5: No Stage~0, model (a) from Chameleon-7B & 96.4\statstd{0.20} & 98.1\statstd{0.17} & 96.5\statstd{0.50} & 94.2\statstd{0.47} & 74.6\statstd{1.07} & 77.3\statstd{0.74} & 77.3\statstd{1.50} & 93.9\statstd{0.40} & 95.6\statstd{0.78} & 83.7\statstd{0.97} & 70.8\statstd{0.19} \\
A6: No Stage~0, model (b) from Chameleon-7B & 98.0\statstd{0.54} & 98.9\statstd{0.31} & 98.0\statstd{0.43} & 96.5\statstd{0.13} & 85.6\statstd{0.34} & 87.5\statstd{0.47} & 83.3\statstd{0.07} & 95.3\statstd{0.95} & 97.7\statstd{0.37} & 87.1\statstd{0.98} & 82.1\statstd{1.12} \\
\multicolumn{12}{@{}l}{\textit{K9 Sensitivity}} \\
B1: $T{=}1$ history frame & 98.3\statstd{0.30} & 99.2\statstd{0.29} & 98.3\statstd{0.23} & 95.8\statstd{0.27} & 87.1\statstd{0.85} & 88.3\statstd{0.90} & 83.9\statstd{0.26} & 95.7\statstd{1.45} & 97.9\statstd{0.25} & 87.6\statstd{1.43} & 81.5\statstd{0.67} \\
B2: $T{=}2$ history frames & 99.0\statstd{0.10} & 99.7\statstd{0.07} & 98.8\statstd{0.32} & 97.9\statstd{0.20} & 89.6\statstd{0.64} & 90.1\statstd{0.47} & 86.9\statstd{0.85} & 96.6\statstd{0.40} & 98.2\statstd{0.43} & 88.7\statstd{0.17} & 84.5\statstd{0.83} \\
B3: Re-planning interval $8$ & 99.0\statstd{0.20} & 99.9\statstd{0.12} & 98.8\statstd{0.28} & 98.5\statstd{0.55} & 89.8\statstd{0.90} & 90.8\statstd{0.88} & 87.9\statstd{0.57} & 96.5\statstd{0.43} & 98.4\statstd{0.79} & 89.0\statstd{0.20} & 85.9\statstd{0.36} \\
B4: Re-planning interval $32$ & 98.6\statstd{0.15} & 99.7\statstd{0.27} & 98.3\statstd{0.31} & 96.3\statstd{0.23} & 87.6\statstd{0.44} & 88.1\statstd{1.10} & 84.6\statstd{1.36} & 95.5\statstd{0.97} & 98.0\statstd{0.52} & 88.2\statstd{0.86} & 82.2\statstd{0.04} \\
\bottomrule
\end{tabularx}
\end{table*}

\subsection{Following a rewritten instruction}\label{app:retarget}

Each Re-target trial on LIBERO-Object changes the requested object while holding the scene, basket and initial state fixed. We assign 50 initial states to each of ten ordered scene--target pairs, for 500 trials per checkpoint. Re-target counts trials in which the new object is placed in the basket, including those with both objects in the basket. Original only counts trials in which the original object is placed in the basket and the new object is not. Neither means neither object is in the basket. For the two supervised pointers, we split Re-target into mutually exclusive final outcomes: only the new target is in the basket, or both targets are in it. All rates use the 500 assigned trials as the denominator. A separate 200-episode multi-instance test adds a same-class distractor only when the instruction still identifies a unique object; success requires placing that object.

\noindent\textbf{Readout and role supervision (K2).} From mean pooling, each step toward a supervised, identity-keyed readout improves target selection, and the Re-target and Original-only comparisons in Tab.~\ref{tab:readout-seeds} hold in all three seeds. All readouts share the OA Stage-I initialization, the full Stage-II schedule and the test-time recovery rule (Supp.~\ref{app:reassignment}). With role supervision off, the identity-key pointer raises Re-target from 56.9\% with mean pooling to 64.6\% and lowers Original-only placements from 29.1\% to 20.3\%. Supervising the same pointer raises Re-target to 77.4\% and lowers Original-only placements to 11.3\%. An equally supervised state-key pointer reaches 69.8\% Re-target, the same rate as the action query without an object readout (Tab.~\ref{tab:abl-train}), and places only the original object in 15.2\% of trials. Across 1,500 trials per readout, the identity key gives 114 more new-target placements than the state key (1,161 vs.\ 1,047; 7.60 percentage points). Of that difference, 83 trials come from placing only the new target (988 vs.\ 905; 5.53 points) and 31 from placing both targets (173 vs.\ 142; 2.07 points). The new-only gain is 20, 37 and 26 trials in the three seeds, respectively. The best public checkpoint on this test, Cosmos Policy, reaches 53.4\% Re-target (Tab.~\ref{tab:baseline-artifacts}).

Making the policy addressable step by step improves the three-seed mean of every summary metric in Tab.~\ref{tab:abl-train}. Starting from a policy without trunk OA and with mean-pooled slots, we enable trunk OA, replace mean pooling with the identity-key pointer and add role supervision. These steps raise Geo Avg from 83.9\% to 86.9\%, 88.0\% and 89.3\%, Overall from 85.9\% to 88.1\%, 89.0\% and 90.5\%, Re-target from 54.5\% to 56.9\%, 64.6\% and 77.4\%, and Multi-inst.\ from 68.3\% to 70.7\%, 73.8\% and 80.0\%; standard LIBERO and Noise success also rise at every step. With mean pooling fixed, trunk OA improves mean success on all seven LIBERO-Plus axes.

Four role queries match or exceed the reported means for one and eight queries on every summary metric; Re-target is 72.5\%, 77.4\% and 75.9\%, and Multi-inst.\ is 74.3\%, 80.0\% and 77.7\%, for $R=1$, 4 and 8. The default role-loss schedule ($\lambda_r=0.05$ for the first 50k updates) and the stronger $\lambda_r=0.5$ setting give the same Re-target (77.4\%); the default gives higher Geo Avg (89.3\% vs.\ 88.1\%) and Multi-inst.\ (80.0\% vs.\ 77.8\%).

\begin{table}[!ht]
\centering
\caption{Readout variants after 100k Stage-II updates, evaluated on 500 rewritten-instruction trials per seed. Re-target (New) is the sum of New only and Both; the remaining trials end with only the original target or neither target in the basket. Dashes: split given for supervised pointers only. Below: mean $\pm$ sample SD (\%) over three training seeds; \textbf{bold}/\underline{underline}: best/next distinct per ranked column.}
\label{tab:readout-seeds}
\small\setlength{\tabcolsep}{4pt}
\begin{tabularx}{\linewidth}{@{}X*{6}{r}@{}}\toprule
Readout / role loss & Seed & New & New only & Both & Original only & Neither \\
\midrule
Mean pooling / off & 1 & 282 & --- & --- & 147 & 71 \\
Mean pooling / off & 2 & 272 & --- & --- & 155 & 73 \\
Mean pooling / off & 3 & 299 & --- & --- & 134 & 67 \\
Pointer (identity key) / off & 1 & 320 & --- & --- & 93 & 87 \\
Pointer (identity key) / off & 2 & 320 & --- & --- & 105 & 75 \\
Pointer (identity key) / off & 3 & 329 & --- & --- & 107 & 64 \\
Pointer (state key) / on & 1 & 353 & 310 & 43 & 76 & 71 \\
Pointer (state key) / on & 2 & 354 & 299 & 55 & 72 & 74 \\
Pointer (state key) / on & 3 & 340 & 296 & 44 & 80 & 80 \\
Pointer (identity key) / on & 1 & 388 & 330 & 58 & 59 & 53 \\
Pointer (identity key) / on & 2 & 390 & 336 & 54 & 55 & 55 \\
Pointer (identity key) / on & 3 & 383 & 322 & 61 & 55 & 62 \\
\bottomrule\end{tabularx}\par\smallskip
\begin{tabularx}{\linewidth}{@{}X*{5}{r}@{}}\toprule
Readout / role loss & Re-target $\uparrow$ & New only $\uparrow$ & Both & Original only $\downarrow$ & Neither \\
\midrule
Mean pooling / off & 56.87\statstd{2.73} & --- & --- & 29.07\statstd{2.12} & 14.07\statstd{0.61} \\
Pointer (identity key) / off & 64.60\statstd{1.04} & --- & --- & 20.33\statstd{1.51} & 15.07\statstd{2.30} \\
Pointer (state key) / on & \underline{69.80}\statstd{1.56} & \underline{60.33}\statstd{1.47} & 9.47\statstd{1.33} & \underline{15.20}\statstd{0.80} & 15.00\statstd{0.92} \\
Pointer (identity key) / on & \textbf{77.40}\statstd{0.72} & \textbf{65.87}\statstd{1.40} & 11.53\statstd{0.70} & \textbf{11.27}\statstd{0.46} & 11.33\statstd{0.95} \\
\bottomrule\end{tabularx}
\end{table}

\subsection{Future-state supervision and recovery}\label{app:recovery-controls}

\paragraph{Learning from future object states (K6).}
The three recovery-off rows of Tab.~\ref{tab:world-ablation}A compare Stage-II supervision for future image codes and for each object's next content and pose. Within a seed, all three policies share the Stage-I initialization, prediction heads, other objectives and 100k training updates. The world and image loss weights $(\lambda_w,\lambda_v)$ are $(0,0.04)$ for image codes only, $(0.5,0)$ for object states only and $(0.5,0.04)$ for both. The image head reads each current image-code state, which precedes the frame's proprioception and slots in token order; the world head reads slot states, which attend to the whole current frame. All three use the same 4,674 LIBERO-Plus geometry trials per seed, with initial grounding and tracking active and both recovery triggers disabled.

Object states alone outperform image codes alone by 3.15 points on Camera, 2.06 on Robot init and 4.13 on Layout. The paired Geo Avg gain is $3.11\pm1.19$ points (mean $\pm$ sample SD of the three within-seed differences). Adding image-code supervision to object-state prediction yields a further $0.47\pm0.08$ points. All three axes follow this ordering in every seed. Adding object-state supervision to image-code prediction gives $3.59\pm1.17$ points on Geo Avg and $0.78\pm0.40$ on standard LIBERO.

\begin{center}
\begin{minipage}{\linewidth}\footnotesize
\noindent\textbf{Table~\ref{tab:world-ablation}A (continued): standard LIBERO with recovery off.} Image-code supervision is enabled in both rows ($\lambda_v=0.04$).
Entries in the upper panel give successes in seed order 1/2/3. Each suite has 1,500 trials per seed. The lower panel gives mean success (\%) and sample SD across seeds.\par\smallskip
\setlength{\tabcolsep}{3pt}
\begin{tabularx}{\linewidth}{@{}l*{5}{Y}@{}}
\toprule
$\lambda_w$ & Spatial & Object & Goal & Long & Total / 6,000\\
\midrule
0 & 1466/1479/1476 & 1491/1487/1496 & 1470/1463/1476 & 1452/1440/1447 & 5879/5869/5895\\
0.5 & 1479/1488/1485 & 1492/1498/1496 & 1473/1488/1483 & 1466/1469/1466 & 5910/5943/5930\\
\midrule
\multicolumn{6}{@{}l}{\emph{Mean $\pm$ sample SD}}\\
0 & $98.24\pm0.45$ & $99.42\pm0.30$ & $97.98\pm0.43$ & $96.42\pm0.40$ & $98.02\pm0.22$\\
0.5 & $98.93\pm0.31$ & $99.69\pm0.20$ & $98.76\pm0.51$ & $97.80\pm0.12$ & $98.79\pm0.28$\\
\bottomrule
\end{tabularx}
\end{minipage}
\end{center}

In the full system, object-state forecasts matter far more than image-code prediction: removing only the auxiliary VQ loss changes Geo Avg from 89.3\% to 89.0\% (Tab.~\ref{tab:abl-train}), whereas removing the world loss together with the residual trigger, which leaves only disappearance-triggered re-grounding at test time, lowers Geo Avg from 89.3\% to 83.1\%, Re-target from 77.4\% to 70.3\% and Multi-inst.\ from 80.0\% to 71.8\%. Without the auxiliary motion monitor, whose loss updates only that branch, the three means remain close to the reference (89.4\%, 77.4\% and 79.7\%). The EMA content target gives higher mean success and smaller sample SDs. It raises Geo Avg from 87.7\% to 89.3\% and reduces its SD from 0.50 to 0.04 points; the Re-target and Multi-inst.\ SDs fall from 1.93 to 0.72 and from 4.27 to 0.50 points.

\paragraph{Recovery on natural geometric perturbations.}
Each seed uses one 100k model~(b) checkpoint ($\lambda_w=0.5$) on the same 4,674 LIBERO-Plus geometry trials, without injected swaps. All rules retain first-frame grounding and ordinary tracking. With recovery enabled, disappearance of a named object triggers re-grounding. For visible objects, the observation-change rule compares the current state with the previous observation; the forecast-residual rule compares it with the cached forecast. Both are calibrated as in Supp.~\ref{app:reassignment}, with the same $\alpha=0.05$. All three active rules share re-detection, the three-query cooldown and reassignment with the predicted position.

Disappearance alone raises Geo Avg by 3.24 points over recovery off. Adding observation change gains another 1.14 points; adding the forecast residual instead gains 1.79 points. The residual exceeds observation change by 0.69 points on Camera, 0.43 on Robot init and 0.83 on Layout (0.65 on Geo Avg), with the same ordering on all three axes in every seed. On standard LIBERO, recovery off and the forecast-residual rule score 98.8\% and 99.0\%, respectively (Tab.~\ref{tab:world-ablation}A).

\begin{center}
\begin{minipage}{\linewidth}
\footnotesize
\noindent\textbf{Table~\ref{tab:world-ablation}A (continued): geometry by forecast target and recovery trigger.} Axis entries are task successes in seed order 1/2/3. Change and Residual include disappearance and add the named trigger for visible objects. Geo Avg weights the three within-seed axis rates equally. The lower panel reports mean success (\%) and sample SD (percentage points) across seeds.\par\smallskip
\setlength{\tabcolsep}{3pt}
\begin{tabularx}{\linewidth}{@{}l l*{4}{Y}@{}}
\toprule
Forecast targets & Trigger & Camera / 1,599 & Robot init / 1,550 & Layout / 1,525 & Geo Avg (\%)\\
\midrule
Image codes & Off & 1302/1289/1311 & 1293/1303/1314 & 1173/1150/1187 & 80.59/80.03/81.53\\
Object states & Off & 1349/1361/1343 & 1329/1332/1345 & 1235/1250/1214 & 83.70/84.34/83.46\\
Both & Off & 1355/1371/1351 & 1333/1338/1352 & 1243/1257/1224 & 84.08/84.83/83.99\\
Both & Disappearance & 1411/1397/1426 & 1392/1380/1384 & 1301/1285/1302 & 87.79/86.89/87.95\\
Both & Change & 1430/1416/1434 & 1398/1392/1406 & 1327/1312/1322 & 88.88/88.13/89.03\\
Both & Residual & 1437/1437/1439 & 1401/1405/1410 & 1337/1333/1329 & 89.31/89.31/89.37\\
\midrule
\multicolumn{6}{@{}l}{\emph{Mean $\pm$ sample SD}}\\
Image codes & Off & $81.34\pm0.69$ & $84.09\pm0.68$ & $76.72\pm1.23$ & $80.72\pm0.76$\\
Object states & Off & $84.49\pm0.57$ & $86.15\pm0.55$ & $80.85\pm1.19$ & $83.83\pm0.46$\\
Both & Off & $84.99\pm0.66$ & $86.52\pm0.64$ & $81.40\pm1.09$ & $84.30\pm0.46$\\
Both & Disappearance & $88.26\pm0.91$ & $89.38\pm0.39$ & $84.98\pm0.63$ & $87.54\pm0.57$\\
Both & Change & $89.22\pm0.59$ & $90.24\pm0.45$ & $86.58\pm0.50$ & $88.68\pm0.48$\\
Both & Residual & $89.91\pm0.07$ & $90.67\pm0.29$ & $87.41\pm0.26$ & $89.33\pm0.04$\\
\bottomrule
\end{tabularx}
\end{minipage}
\end{center}

\begin{center}
\begin{minipage}{\linewidth}
\footnotesize
\noindent\textbf{Recovery cost and paired outcomes on the same geometry trials.} The left block gives episodes with at least one intervention / total interventions; both are zero with recovery off. The right block counts matched trials that changed from failure to success / from success to failure. Each seed contributes 4,674 trials.\par\smallskip
\setlength{\tabcolsep}{3pt}
\begin{tabularx}{\linewidth}{@{}l*{6}{Y}@{}}
\toprule
& \multicolumn{3}{c}{Episodes / interventions} & \multicolumn{3}{c}{Paired outcome changes}\\
\cmidrule(lr){2-4}\cmidrule(l){5-7}
Seed & Disapp. & Change & Residual & Off $\to$ Disapp. & Disapp. $\to$ Change & Disapp. $\to$ Residual\\
\midrule
1 & 296/398 & 461/622 & 553/743 & 195/22 & 68/17 & 93/22\\
2 & 216/290 & 377/521 & 497/675 & 117/21 & 79/21 & 130/17\\
3 & 299/393 & 458/603 & 529/690 & 205/20 & 65/15 & 80/14\\
\midrule
All & 811/1,081 & 1,296/1,746 & 1,579/2,108 & 517/63 & 212/53 & 303/53\\
\bottomrule
\end{tabularx}
\end{minipage}
\end{center}

Across 14,022 trials per rule, disappearance alone intervenes in 5.78\% of episodes, observation change in 9.24\% and the forecast residual in 11.26\%. At these thresholds, the residual yields 91 more successes than observation change (20, 55 and 16 by seed) at the cost of 362 additional interventions.

\paragraph{Injected track swaps at a fixed checkpoint.}
At the third query, we exchange the tracks assigned to the target and its most similar distractor by initial DINOv3 appearance. Their addresses stay fixed. For each rule and seed, 40 tasks and ten fixed initial states give 400 assigned swap trials and 400 clean controls. Task success uses all assigned trials. Detection means re-grounding within two queries of an applied swap, and its denominator is the number of applied swaps (360, 356 and 364 by seed). Delay is the median among detections. Clean events count all interventions, so one episode can contribute several.

The forecast residual is the more informative trigger. At the same policy checkpoint within each seed, a trigger starts re-detection and reassignment to stored identities (Supp.~\ref{app:reassignment}). The residual detects 76.2\% of applied swaps, against 61.7\% for observation change, at a median delay of zero queries, and it intervenes less often on clean episodes (0.051 vs.\ 0.082 per episode). After injected swaps, task success is 68.0\% with the forecast residual, 64.7\% with observation change and 56.3\% without re-grounding. In every seed, the forecast residual detects more swaps and triggers fewer clean interventions than observation change, and its swap-task success is equal or higher. On clean trials, success is 97.5\% with the forecast residual and 98.1\% without re-grounding.

On seed 1 (Tab.~\ref{tab:abl-monitor}B), the forecast residual at the default threshold gives higher swap-task success (65.0\%) than the kinematic predictor (62.5\%), the auxiliary motion monitor (63.8\%) and periodic re-grounding every five queries (63.3\%), which intervenes 1.165 times per clean episode. Raising $\alpha$ from 0.01 to 0.1 increases detected swaps from 199 to 284 of 360 and clean interventions from 11 to 32. The default $\alpha=0.05$ gives up one point of swap-task success relative to $\alpha=0.1$ (65.0\% vs.\ 66.0\%) for fewer than half the clean interventions (14 vs.\ 32). The forecast also guides reassignment: removing its position term leaves detection unchanged (257 of 360) but lowers swap-task success from 65.0\% to 60.5\%. Oracle re-grounding at the known swap time reaches 76.3\%.

\begin{table*}[!ht]
\centering
\caption{Recovery after injected track swaps. A: mean $\pm$ sample SD over three training seeds; each rule has $400$ assigned swap trials and $400$ clean trials per seed. Detection uses applied swaps ($360$, $356$, $364$). B: additional rules on seed 1, with the panel-A rules repeated. Delay: median queries until detection ($0$: same query). Scheduled ($\ast$) interventions use fixed intervals and do not count as detections; oracle ($\dagger$) interventions use the known swap time. \textbf{Bold}/\underline{underline}: best/next distinct within each panel; detection and delay rank adaptive rules, task success excludes the oracle, and clean event counts are unranked.}
\label{tab:abl-monitor}
\footnotesize\setlength{\tabcolsep}{2.5pt}
\begin{tabularx}{\linewidth}{@{}X*{4}{r}@{}}
\toprule
\multicolumn{5}{@{}l}{\textbf{A. Three-seed summaries}}\\
Rule & Detected (\%) $\uparrow$ & Swap (\%) $\uparrow$ & Clean (\%) $\uparrow$ & Clean events/ep.\\
\midrule
No re-grounding & --- & 56.3\statstd{3.40} & \textbf{98.1}\statstd{0.63} & 0 (fixed) \\
Observation change & \underline{61.7}\statstd{2.65} & \underline{64.7}\statstd{3.59} & 97.4\statstd{1.04} & 0.082\statstd{0.010} \\
Forecast residual & \textbf{76.2}\statstd{4.22} & \textbf{68.0}\statstd{2.70} & \underline{97.5}\statstd{1.25} & 0.051\statstd{0.018} \\
\bottomrule
\end{tabularx}\par\medskip
\begin{tabularx}{\linewidth}{@{}X*{6}{r}@{}}
\toprule
\multicolumn{7}{@{}l}{\textbf{B. Additional rules on seed 1}}\\
Rule & Detected/applied & Delay & Clean events & Events/ep. & Swap (\%) & Clean (\%) \\
\midrule
No re-grounding & --- & --- & 0 (fixed) & 0.000 & 59.0 & \textbf{98.8} \\
Periodic re-grounding every $5$ queries$\ast$ & 0/360 & --- & 466 & 1.165 & 63.3 & \underline{97.8} \\
Observation change & 214/360 & \underline{1} & 34 & 0.085 & 63.3 & 96.3 \\
Kinematic predictor & 221/360 & \textbf{0} & 34 & 0.085 & 62.5 & 97.5 \\
Auxiliary motion monitor & 242/360 & \textbf{0} & 18 & 0.045 & 63.8 & 97.3 \\
\hl{Forecast residual, $\alpha{=}0.05$ (ours)} & \hl{\underline{257/360}} & \hl{\textbf{0}} & \hl{14} & \hl{0.035} & \hl{\underline{65.0}} & \hl{97.5} \\
Forecast residual, $\alpha{=}0.01$ & 199/360 & \textbf{0} & 11 & 0.028 & 62.8 & 97.5 \\
Forecast residual, $\alpha{=}0.1$ & \textbf{284/360} & \textbf{0} & 32 & 0.080 & \textbf{66.0} & 97.3 \\
Forecast residual, two consecutive exceedances & 137/360 & \underline{1} & 7 & 0.018 & 61.5 & 97.5 \\
Reassignment without the position term & \underline{257/360} & \textbf{0} & 14 & 0.035 & 60.5 & 97.0 \\
Oracle re-grounding at the swap$\dagger$ & 360/360 & 0 & 0 & 0.000 & 76.3 & 98.8 \\
\bottomrule
\end{tabularx}
\end{table*}

\begin{table*}[!ht]
\centering
\caption*{\textbf{Table~\ref{tab:abl-monitor} (continued).} Per-seed counts for the three rules of panel A. Each row has $400$ assigned swap and $400$ clean trials. Clean events count interventions, not affected episodes. Delay is the within-seed median.}
\footnotesize\setlength{\tabcolsep}{4pt}
\begin{tabularx}{\linewidth}{@{}X*{7}{r}@{}}
\toprule
Rule & Seed & Applied & Detected & Delay & Swap/400 & Clean/400 & Clean events\\
\midrule
No re-grounding & 1 & 360 & --- & --- & 236 & 395 & 0 \\
No re-grounding & 2 & 356 & --- & --- & 230 & 392 & 0 \\
No re-grounding & 3 & 364 & --- & --- & 210 & 390 & 0 \\
Observation change & 1 & 360 & 214 & 1 & 253 & 385 & 34 \\
Observation change & 2 & 356 & 230 & 1 & 248 & 391 & 28 \\
Observation change & 3 & 364 & 222 & 1 & 275 & 393 & 36 \\
Forecast residual & 1 & 360 & 257 & 0 & 260 & 390 & 14 \\
Forecast residual & 2 & 356 & 282 & 0 & 281 & 395 & 19 \\
Forecast residual & 3 & 364 & 284 & 0 & 275 & 385 & 28 \\
\bottomrule
\end{tabularx}
\end{table*}

\subsection{Dependence on observations and sensitivity to perception}\label{app:actq}

\noindent\textbf{Direct image access (K5).} With the route open at test time, route dropout raises Geo Avg from 87.4\% to 89.3\% and Layout from 83.9\% to 87.4\%, and Re-target and Multi-inst.\ gain 2.6 and 6.3 points (Tab.~\ref{tab:abl-train}). Dropout trains the action query to use the remaining inputs when its direct access to image codes is closed. The trained policy keeps most of its geometric robustness without the action query's direct access to image codes: closing this route at inference retains 99.0\% standard LIBERO and 88.5\% Geo Avg (vs.\ 89.3\%), while mean Multi-inst.\ falls from 80.0\% to 76.7\%. Because proprioception, slot and historical action tokens can still attend to image codes (Supp.~\ref{app:sequence}), this test measures reliance on the direct route alone. The open route adds image evidence under sensor noise (85.8\% open, 83.0\% closed; Tab.~\ref{tab:abl-infer}). The action query's direct attention follows the same pattern: slots receive 0.38 of its attention mass on standard LIBERO, 0.43 under camera changes and 0.45 under layout changes, but 0.30 under sensor noise, where image codes receive 0.43 (Tab.~\ref{tab:actq}).

\begin{table}[h]
\centering
\caption{Mean attention mass of \textnormal{\textsc{[act\_q]}} by token type, averaged over heads, episodes and the last eight trunk layers ($m=24,\dots,31$). Each row sums to one. Slot mass sums over all instances; Tab.~\ref{tab:controls}C reports target-role pointer-hit gains.}
\label{tab:actq}
\small
\begin{tabular}{lccc}
\toprule
Episodes & Slots & Image codes & Other \\
\midrule
LIBERO (nominal) & 0.38 & 0.39 & 0.23 \\
LIBERO-Plus camera & 0.43 & 0.32 & 0.25 \\
LIBERO-Plus layout & 0.45 & 0.31 & 0.24 \\
LIBERO-Plus noise & 0.30 & 0.43 & 0.27 \\
\bottomrule
\end{tabular}
\end{table}

\begin{table*}[!ht]
\centering
\caption{Inference controls for policies trained with 100k Stage-II updates. Entries give mean success rates (\%) across three training seeds, with sample SD. Each seed has $6{,}000$ standard LIBERO trials and $10{,}030$ full LIBERO-Plus trials. The target-selection diagnostics use 500 Re-target and 200 Multi-inst.\ episodes per seed. X8 reuses the reference Multi-inst.\ trials. The two calibration controls use only the $1{,}599$ Camera trials. Dashes denote metrics outside the evaluation.}
\label{tab:abl-infer}
\scriptsize\setlength{\tabcolsep}{2.8pt}
\begin{tabularx}{\linewidth}{@{}X*{8}{r}@{}}
\toprule
Variant & LIBERO & Geo Avg & Overall & Camera & Noise & Re-target & Multi-inst. & Long \\
\midrule
\hl{Reference: model (b)} & \hl{99.0\statstd{0.03}} & \hl{89.3\statstd{0.04}} & \hl{90.5\statstd{0.03}} & \hl{89.9\statstd{0.07}} & \hl{85.8\statstd{0.22}} & \hl{77.4\statstd{0.72}} & \hl{80.0\statstd{0.50}} & \hl{98.4\statstd{0.14}} \\
\midrule
\multicolumn{9}{@{}l}{\textit{K5 Image route}} \\
\tok{[act\_q]}$\to$image codes closed at evaluation & 99.0\statstd{0.10} & 88.5\statstd{0.24} & 89.5\statstd{0.14} & 89.0\statstd{0.29} & 83.0\statstd{0.30} & 75.5\statstd{1.33} & 76.7\statstd{4.31} & 98.3\statstd{0.17} \\
\multicolumn{9}{@{}l}{\textit{K7 Perception}} \\
P1: No temporal filter & 98.6\statstd{0.14} & 88.3\statstd{0.18} & 89.0\statstd{0.51} & 88.6\statstd{1.09} & 81.6\statstd{1.85} & 76.1\statstd{0.70} & 77.3\statstd{4.37} & 97.3\statstd{0.40} \\
P2: Single-frame pose (no tracking) & 97.8\statstd{0.20} & 85.8\statstd{0.59} & 87.8\statstd{0.37} & 85.8\statstd{1.07} & 80.0\statstd{0.41} & 76.5\statstd{0.42} & 78.0\statstd{6.54} & 95.5\statstd{0.92} \\
P3: Pose fusion without the per-view error term & 98.8\statstd{0.09} & 87.2\statstd{0.35} & 89.3\statstd{0.18} & 85.8\statstd{0.20} & 84.6\statstd{1.03} & 77.1\statstd{1.55} & 79.8\statstd{2.47} & 97.9\statstd{0.37} \\
P4: Exact simulated depth at evaluation (oracle) & 99.3\statstd{0.13} & 91.3\statstd{0.22} & 91.9\statstd{0.18} & 92.2\statstd{1.24} & 88.2\statstd{0.36} & 77.6\statstd{2.62} & 81.0\statstd{1.32} & 99.1\statstd{0.37} \\
P5: Simulator masks at evaluation (oracle) & 99.1\statstd{0.26} & 90.7\statstd{0.31} & 91.8\statstd{0.17} & 91.2\statstd{0.42} & 89.1\statstd{0.88} & 79.4\statstd{2.16} & 84.0\statstd{1.32} & 99.0\statstd{0.54} \\
P6: Camera axis with true instead of nominal extrinsics & --- & --- & --- & 94.4\statstd{0.07} & --- & --- & --- & --- \\
P7: Camera axis with the wrist-anchored offset estimate & --- & --- & --- & 92.6\statstd{0.22} & --- & --- & --- & --- \\
P8: Wrist view masked at evaluation & 97.1\statstd{0.22} & 81.6\statstd{0.59} & 85.2\statstd{0.74} & 77.9\statstd{1.67} & 77.3\statstd{0.66} & 73.9\statstd{1.62} & 75.7\statstd{3.33} & 94.5\statstd{0.48} \\
\multicolumn{9}{@{}l}{\textit{K7 Stress tests}} \\
X1: Pose noise $1$\,cm\,/\,$5^\circ$ & 98.8\statstd{0.03} & 88.5\statstd{0.37} & 89.9\statstd{0.27} & 89.1\statstd{0.72} & 84.6\statstd{0.50} & 76.9\statstd{2.05} & 79.2\statstd{3.79} & 97.8\statstd{0.42} \\
X2: Pose noise $2$\,cm\,/\,$10^\circ$ & 98.0\statstd{0.18} & 86.3\statstd{0.66} & 88.3\statstd{0.55} & 86.9\statstd{1.13} & 82.2\statstd{1.00} & 75.9\statstd{2.02} & 78.0\statstd{1.32} & 96.1\statstd{0.39} \\
X3: Pose noise $5$\,cm\,/\,$20^\circ$ & 93.7\statstd{0.13} & 79.5\statstd{0.77} & 81.8\statstd{0.08} & 79.7\statstd{0.58} & 73.4\statstd{1.09} & 71.5\statstd{2.95} & 72.2\statstd{2.08} & 88.7\statstd{0.47} \\
X4: Target mask lost for $1$ query & 98.8\statstd{0.20} & 88.8\statstd{0.12} & 89.9\statstd{0.21} & 89.4\statstd{0.29} & 84.4\statstd{0.27} & 76.7\statstd{2.14} & 79.2\statstd{0.76} & 97.8\statstd{0.28} \\
X5: Target mask lost for $3$ queries & 97.6\statstd{0.15} & 86.2\statstd{0.23} & 87.6\statstd{0.28} & 86.5\statstd{0.92} & 80.7\statstd{0.83} & 74.8\statstd{2.08} & 75.5\statstd{4.36} & 95.7\statstd{0.33} \\
X6: Target mask lost for $5$ queries & 94.7\statstd{0.38} & 81.9\statstd{0.61} & 83.5\statstd{0.49} & 81.5\statstd{0.97} & 75.2\statstd{1.20} & 70.7\statstd{0.31} & 70.0\statstd{0.50} & 90.9\statstd{0.60} \\
X7: Target and distractor bindings swapped at $t{=}0$ & 84.9\statstd{0.31} & 75.0\statstd{0.37} & 76.2\statstd{0.11} & 76.0\statstd{0.39} & 70.2\statstd{1.38} & 53.8\statstd{2.11} & 54.3\statstd{3.88} & 80.9\statstd{0.50} \\
X8: Same-class distractor added & 95.4\statstd{0.20} & 84.5\statstd{0.70} & 85.8\statstd{0.46} & 85.6\statstd{0.90} & 79.9\statstd{1.07} & 72.2\statstd{2.95} & 80.0\statstd{0.50} & 93.5\statstd{0.45} \\
\multicolumn{9}{@{}l}{\textit{K9 Sensitivity}} \\
B5: $1$ Euler step & 98.2\statstd{0.26} & 87.5\statstd{0.37} & 89.0\statstd{0.31} & 88.2\statstd{0.72} & 83.4\statstd{1.27} & 75.6\statstd{0.92} & 78.0\statstd{7.37} & 96.5\statstd{0.88} \\
B6: $10$ Euler steps & 99.1\statstd{0.23} & 89.5\statstd{0.20} & 90.7\statstd{0.32} & 90.1\statstd{0.47} & 86.0\statstd{1.22} & 77.6\statstd{2.46} & 80.5\statstd{2.29} & 98.5\statstd{0.00} \\
\bottomrule
\end{tabularx}
\end{table*}

\begin{table*}[!ht]
\centering
\caption*{\textbf{Table~\ref{tab:abl-infer} (continued).} Standard LIBERO suites (1,500 trials per suite and seed) and full LIBERO-Plus axes. Each seed has 1,599 Camera, 1,550 Robot init, 1,525 Layout, 1,142 Light, 1,076 Background, 1,537 Language and 1,601 Noise trials. Dashes denote axes outside the evaluation.}
\scriptsize\setlength{\tabcolsep}{2.3pt}
\begin{tabularx}{\linewidth}{@{}X*{11}{r}@{}}
\toprule
 & \multicolumn{4}{c}{Standard LIBERO} & \multicolumn{7}{c}{Full LIBERO-Plus} \\
\cmidrule(lr){2-5}\cmidrule(lr){6-12}
Variant & Spatial & Object & Goal & Long & Cam & Init & Layout & Light & BG & Lang & Noise \\
\midrule
\hl{Reference: model (b)} & \hl{99.0\statstd{0.13}} & \hl{99.9\statstd{0.04}} & \hl{98.8\statstd{0.17}} & \hl{98.4\statstd{0.14}} & \hl{89.9\statstd{0.07}} & \hl{90.7\statstd{0.29}} & \hl{87.4\statstd{0.26}} & \hl{96.6\statstd{0.31}} & \hl{98.4\statstd{0.14}} & \hl{89.0\statstd{0.07}} & \hl{85.8\statstd{0.22}} \\
\midrule
\multicolumn{12}{@{}l}{\textit{K5 Image route}} \\
\tok{[act\_q]}$\to$image codes closed at evaluation & 98.9\statstd{0.14} & 99.9\statstd{0.08} & 98.8\statstd{0.18} & 98.3\statstd{0.17} & 89.0\statstd{0.29} & 90.0\statstd{0.28} & 86.5\statstd{0.29} & 96.2\statstd{0.25} & 98.2\statstd{0.14} & 88.5\statstd{0.23} & 83.0\statstd{0.30} \\
\multicolumn{12}{@{}l}{\textit{K7 Perception}} \\
P1: No temporal filter & 99.0\statstd{0.20} & 99.7\statstd{0.07} & 98.6\statstd{0.08} & 97.3\statstd{0.40} & 88.6\statstd{1.09} & 90.2\statstd{0.75} & 86.2\statstd{0.76} & 95.4\statstd{1.15} & 97.8\statstd{0.85} & 88.2\statstd{0.85} & 81.6\statstd{1.85} \\
P2: Single-frame pose (no tracking) & 98.3\statstd{0.64} & 99.7\statstd{0.15} & 97.8\statstd{0.54} & 95.5\statstd{0.92} & 85.8\statstd{1.07} & 87.5\statstd{1.10} & 84.0\statstd{0.27} & 95.9\statstd{0.13} & 98.3\statstd{0.25} & 88.2\statstd{0.16} & 80.0\statstd{0.41} \\
P3: Pose fusion without the per-view error term & 98.7\statstd{0.18} & 99.8\statstd{0.04} & 98.6\statstd{0.17} & 97.9\statstd{0.37} & 85.8\statstd{0.20} & 89.6\statstd{0.29} & 86.2\statstd{0.58} & 96.0\statstd{0.31} & 98.4\statstd{0.43} & 88.8\statstd{0.46} & 84.6\statstd{1.03} \\
P4: Exact simulated depth at evaluation (oracle) & 99.5\statstd{0.04} & 99.9\statstd{0.07} & 98.6\statstd{0.17} & 99.1\statstd{0.37} & 92.2\statstd{1.24} & 92.2\statstd{1.11} & 89.6\statstd{0.67} & 96.6\statstd{0.54} & 98.7\statstd{0.48} & 89.2\statstd{1.01} & 88.2\statstd{0.36} \\
P5: Simulator masks at evaluation (oracle) & 98.9\statstd{0.33} & 99.8\statstd{0.14} & 98.8\statstd{0.21} & 99.0\statstd{0.54} & 91.2\statstd{0.42} & 90.9\statstd{0.75} & 90.0\statstd{0.35} & 97.3\statstd{0.66} & 98.6\statstd{0.33} & 89.5\statstd{1.05} & 89.1\statstd{0.88} \\
P6: Camera axis with true instead of nominal extrinsics & --- & --- & --- & --- & 94.4\statstd{0.07} & --- & --- & --- & --- & --- & --- \\
P7: Camera axis with the wrist-anchored offset estimate & --- & --- & --- & --- & 92.6\statstd{0.22} & --- & --- & --- & --- & --- & --- \\
P8: Wrist view masked at evaluation & 97.8\statstd{0.42} & 98.8\statstd{0.20} & 97.3\statstd{0.25} & 94.5\statstd{0.48} & 77.9\statstd{1.67} & 83.4\statstd{1.29} & 83.7\statstd{0.47} & 95.9\statstd{0.51} & 98.2\statstd{0.52} & 87.4\statstd{1.86} & 77.3\statstd{0.66} \\
\multicolumn{12}{@{}l}{\textit{K7 Stress tests}} \\
X1: Pose noise $1$\,cm\,/\,$5^\circ$ & 98.8\statstd{0.04} & 99.8\statstd{0.08} & 99.0\statstd{0.23} & 97.8\statstd{0.42} & 89.1\statstd{0.72} & 89.7\statstd{0.50} & 86.9\statstd{1.18} & 96.5\statstd{0.22} & 98.4\statstd{0.21} & 89.1\statstd{0.89} & 84.6\statstd{0.50} \\
X2: Pose noise $2$\,cm\,/\,$10^\circ$ & 98.1\statstd{0.33} & 99.6\statstd{0.17} & 98.1\statstd{0.12} & 96.1\statstd{0.39} & 86.9\statstd{1.13} & 87.4\statstd{1.31} & 84.6\statstd{0.63} & 95.7\statstd{0.35} & 98.0\statstd{0.14} & 88.0\statstd{0.52} & 82.2\statstd{1.00} \\
X3: Pose noise $5$\,cm\,/\,$20^\circ$ & 94.1\statstd{0.50} & 96.8\statstd{0.28} & 95.3\statstd{0.54} & 88.7\statstd{0.47} & 79.7\statstd{0.58} & 80.8\statstd{0.81} & 78.0\statstd{1.29} & 91.0\statstd{1.01} & 94.0\statstd{1.46} & 82.3\statstd{0.14} & 73.4\statstd{1.09} \\
X4: Target mask lost for $1$ query & 98.6\statstd{0.45} & 99.5\statstd{0.07} & 99.1\statstd{0.07} & 97.8\statstd{0.28} & 89.4\statstd{0.29} & 90.3\statstd{0.38} & 86.7\statstd{0.32} & 96.1\statstd{0.57} & 97.9\statstd{0.05} & 88.4\statstd{1.10} & 84.4\statstd{0.27} \\
X5: Target mask lost for $3$ queries & 97.9\statstd{0.35} & 99.1\statstd{0.13} & 97.8\statstd{0.17} & 95.7\statstd{0.33} & 86.5\statstd{0.92} & 88.7\statstd{0.80} & 83.3\statstd{0.82} & 95.1\statstd{0.05} & 97.3\statstd{0.19} & 86.8\statstd{1.64} & 80.7\statstd{0.83} \\
X6: Target mask lost for $5$ queries & 95.3\statstd{0.23} & 97.2\statstd{0.23} & 95.4\statstd{0.81} & 90.9\statstd{0.60} & 81.5\statstd{0.97} & 84.2\statstd{0.62} & 79.9\statstd{0.76} & 91.9\statstd{0.76} & 94.9\statstd{0.14} & 83.1\statstd{0.86} & 75.2\statstd{1.20} \\
X7: Target and distractor bindings swapped at $t{=}0$ & 85.3\statstd{0.54} & 87.9\statstd{0.13} & 85.4\statstd{1.30} & 80.9\statstd{0.50} & 76.0\statstd{0.39} & 77.3\statstd{0.46} & 71.7\statstd{0.69} & 84.9\statstd{0.61} & 86.6\statstd{1.31} & 72.3\statstd{1.64} & 70.2\statstd{1.38} \\
X8: Same-class distractor added & 95.7\statstd{0.32} & 97.8\statstd{0.46} & 94.6\statstd{0.85} & 93.5\statstd{0.45} & 85.6\statstd{0.90} & 87.0\statstd{0.06} & 81.0\statstd{1.17} & 92.8\statstd{0.83} & 95.4\statstd{0.43} & 84.2\statstd{0.29} & 79.9\statstd{1.07} \\
\multicolumn{12}{@{}l}{\textit{K9 Sensitivity}} \\
B5: $1$ Euler step & 98.6\statstd{0.43} & 99.4\statstd{0.08} & 98.4\statstd{0.43} & 96.5\statstd{0.88} & 88.2\statstd{0.72} & 89.0\statstd{0.23} & 85.3\statstd{1.00} & 96.0\statstd{0.46} & 97.9\statstd{0.19} & 88.4\statstd{0.11} & 83.4\statstd{1.27} \\
B6: $10$ Euler steps & 99.1\statstd{0.17} & 99.8\statstd{0.31} & 98.9\statstd{0.57} & 98.5\statstd{0.00} & 90.1\statstd{0.47} & 90.8\statstd{0.13} & 87.5\statstd{0.97} & 96.5\statstd{0.51} & 98.5\statstd{0.43} & 89.2\statstd{0.73} & 86.0\statstd{1.22} \\
\bottomrule
\end{tabularx}
\end{table*}

\noindent\textbf{Pose, tracking and initial binding (K7).} Each stage of the perception pipeline contributes at fixed checkpoints (Tab.~\ref{tab:abl-infer}). The temporal filter on appearance and mask features raises Noise success from 81.6\% to 85.8\%. Replacing tracked poses with single-frame estimates lowers Geo Avg to 85.8\% and Long to 95.5\%, compared with 88.3\% and 97.3\% without the temporal filter (P1). Accounting for per-view systematic error in pose fusion raises Camera success from 85.8\% to 89.9\%. Perception accuracy limits success: training and evaluating with simulator poses raises Geo Avg from 89.3\% to 93.3\% and Overall from 90.5\% to 93.1\% (O1, Tab.~\ref{tab:abl-train}). Exact depth and simulator masks at evaluation give similar Overall success (91.9\% and 91.8\%). In every seed, exact depth gives the higher Geo Avg (91.3\% vs.\ 90.7\%) and simulator masks the higher Multi-inst.\ (84.0\% vs.\ 81.0\%). Their Re-target rates (79.4\% with masks, 77.6\% with depth) differ by less than the across-seed SD.

The wrist view anchors object poses under camera changes. On the Camera axis, masking it at inference lowers success from 89.9\% to 77.9\%, and using it to estimate the external-camera offset raises success to 92.6\%, against 94.4\% with true extrinsics. Table~\ref{tab:controls}A separately evaluates the external-camera policies (a$'$)/(b$'$).

Success falls steadily as injected pose noise grows: from $1$\,cm/$5^\circ$ to $2$\,cm/$10^\circ$ and $5$\,cm/$20^\circ$, Geo Avg falls from $88.5\%$ to $86.3\%$ and $79.5\%$, and Long from $97.8\%$ to $96.1\%$ and $88.7\%$. Withholding the target mask for one, three and five consecutive queries progressively lowers success; at five queries, Long falls to $90.9\%$, Re-target to $70.7\%$ and Multi-inst.\ to $70.0\%$.

Correct initial binding is particularly important for acting on the requested instance. Swapping the physical instances bound to the target and distractor addresses at the first query (X7) lowers Re-target from 77.4\% to 53.8\% and Multi-inst.\ from 80.0\% to 54.3\%. Swaps injected at the third query are detected by the forecast residual in 76.2\% of applied swaps (Tab.~\ref{tab:abl-monitor}). Adding a same-class distractor to the rewritten-instruction trials (X8) gives 72.2\% Re-target on the same denominator. For Multi-inst., X8 reuses the reference trials: 159, 160 and 161 successes out of 200 per seed (mean 80.0\%, SD 0.50 points).

\noindent\textbf{Euler integration (K9).} At fixed checkpoints, increasing the action solver from one to four Euler steps raises Overall from 89.0\% to 90.5\% and Long from 96.5\% to 98.4\%. Ten steps reach 90.7\% and 98.5\% (Tab.~\ref{tab:abl-infer}), so four steps capture most of the gain at less than half the integration cost of ten.

\FloatBarrier
\section{Real-robot tasks, deployment and perturbations}\label{app:real}

\subsection{Tasks and success criteria}

On AgileX PiPER, T1 and T2 use one arm and T3--T5 use two. Evaluation uses held-out scene--instruction pairs and 20 trials per task. In T2, the instruction names one of three identical cups by its starting position; T5 requires both a handoff and a correct final placement. Wrong objects or destinations, partially completed instructions and timeouts count as failures. A synchronized bimanual episode counts as one trial. For multi-view \method{} rollouts of cup placement, block stacking and related tabletop tasks, see Supp.~\ref{app:real-sequences}.

\noindent\begin{tabularx}{\linewidth}{@{}l X c@{}}
\toprule
Task & Instruction and success criterion & Limit\\
\midrule
T1 & Place the named pink, green or white cup upright on the named blue or white tray and release it. Both colors must match. & 60\,s \\
T2 & Place in the basket the cup that started at the named left, middle or right position. The three cups are identical; positions are defined in the robot base frame. No other cup may enter the basket. & 60\,s \\
T3 & The left arm moves the named base block to the marked zone; the right arm stacks the named top block on it. The stack must remain stable for 3\,s after release. & 60\,s \\
T4 & One arm puts the named fruit in the bowl and the other puts the named vegetable in the basket. Both placements must be correct, in either order. Each of five layouts has two instructions, each tested twice after reset. & 180\,s \\
T5 & From 8--10 objects, the left arm selects the named item and releases it in the marked transfer zone. The right arm picks it up there and places it alone on the tray. & 60\,s \\
\bottomrule
\end{tabularx}

\subsection{Shared demonstrations and policy adaptation}\label{app:real-training}

Each method was fine-tuned once per arm configuration (seed 42) on the same 20 nominal teleoperated demonstrations per task: 40 single-arm and 60 synchronized bimanual trajectories. Within each configuration, we ran each policy with its own deployment settings on the same test scenes and instructions, scoring success by the task criteria above. The perturbed trials reuse the nominal single-arm checkpoints of $\pi_{0.5}$, Fast-WAM and \method{}. Training used eight 80\,GB A100 GPUs in BF16; inference used one 80\,GB A100.

The single arm has a parallel gripper and front (external) and wrist RealSense D435i cameras. The bimanual policy controls both arms from three RGB streams: the front camera and the two wrist cameras. All methods receive language and proprioception; none receives measured depth. \method{} derives geometry from RGB (Supp.~\ref{app:monocular-depth}). The VLA baselines and Cosmos Policy take $224\times224$ images. Fast-WAM packs its three bimanual views into a $384\times320$ image and uses only the current observation. The PiPER controller accepts commands at 25\,Hz: six absolute joint targets (radians) and one gripper opening in $[0,1]$ per arm. The bimanual action ($d_a=14$) lists the six joint targets and the gripper opening of the left arm, followed by those of the right arm. We compute action normalization statistics from the training split for that arm configuration.

\method{} queries synchronously. While the controller holds the last command, the policy acquires the observations and predicts the next 16 commands; in both arm configurations, the controller then executes all 16 at 25\,Hz, which takes $16/25=0.64$\,s. Each query cycle therefore lasts $0.64\,\mathrm{s}+L_{\rm query}$, where $L_{\rm query}$ covers observation acquisition, perception and policy inference.

\noindent\textit{Policy adaptation.}\par\nobreak

\begingroup
\small
\setlength{\tabcolsep}{3pt}
\renewcommand{\arraystretch}{1.08}
\noindent\begin{tabularx}{\linewidth}{@{}l X X@{}}
\toprule
Method & Base model & Trained components / implementation \\
\midrule
$\pi_{0.5}$ & \texttt{pi05\_base} & Full policy; openpi JAX, sharded training \\
OpenVLA-OFT & \texttt{openvla/openvla-7b} & Rank-32 LoRA, action head and state projection; FiLM on bimanual tasks (OFT+) \\
GR00T N1.5 & \texttt{nvidia/GR00T-N1.5-3B} & Action DiT and adapters; vision/language backbone frozen; four denoising steps \\
Fast-WAM & Wan2.2-TI2V-5B; action expert initialized by Wan-weight interpolation & Video/action MoT; VAE and text encoder frozen; ZeRO-2 \\
Cosmos Policy & Cosmos-Predict2-2B-Video2World, 480p/16fps weights & Full policy; FSDP \\
\method{} & Chameleon-7B OA-WAM & Rank-32 trunk LoRA and policy modules; perception frozen; checkpoint at 50k updates \\
\bottomrule
\end{tabularx}
\endgroup
\par\smallskip
\noindent\textit{Training and execution settings.} Each policy predicts $H$ actions per query and executes the first $K$.\par\nobreak

\begingroup
\small
\setlength{\tabcolsep}{5pt}
\noindent\begin{tabularx}{\linewidth}{@{}l*{5}{Y}@{}}
\toprule
Policy & Updates & Global batch & LR & $H$ & $K$\\
\midrule
$\pi_{0.5}$ & 20k & 64 & $5\times10^{-5}$ & 50 & 10\\
OpenVLA-OFT & 100k & 32 & $5\times10^{-4}$ & 25 & 25\\
GR00T N1.5 & 20k & 128 & $10^{-4}$ & 16 & 8\\
Fast-WAM & 30k & 1,024 & $10^{-4}$ & 32 & 16\\
Cosmos Policy & 50k & 200 & $10^{-4}$ & 50 & 50\\
\midrule
\method{} (single arm) & 50k & 64 & $10^{-4}$ & 16 & 16\\
\method{} (bimanual) & 50k & 32 & $10^{-4}$ & 16 & 16\\
\bottomrule
\end{tabularx}
\par\nobreak\smallskip
\noindent{\footnotesize\textit{Training notes.} LR is the peak learning rate of each schedule. \method{} accumulates gradients over four micro-batches on eight GPUs, with two examples per GPU micro-batch for single-arm training and one for bimanual training ($8\times2\times4=64$ and $8\times1\times4=32$). GR00T uses cosine decay with 5\% warmup; Fast-WAM uses cosine decay, AdamW (weight decay $0.01$) and gradient accumulation. OpenVLA-OFT reduces its learning rate to $5\times10^{-5}$ after 50k updates; Cosmos Policy starts its decay after 20k updates.\par}
\endgroup

\subsection{Nominal and perturbed deployments}

With 20 trials per task, the five-task mean in Tab.~\ref{tab:real} equals the number of successes out of 100. Treating trials within each task as independent Bernoulli draws, the Wilson 95\% intervals for the fixed \method{} checkpoints on T1--T5 are $[76.4,99.1]$, $[83.9,100.0]$, $[64.0,94.8]$, $[69.9,97.2]$ and $[53.1,88.8]\%$, respectively.

\noindent\begin{tabularx}{\linewidth}{@{}l*{3}{Y}@{}}
\toprule
Method & Single arm (T1--T2) & Bimanual (T3--T5) & Five tasks \\
\midrule
$\pi_{0.5}$ & 23/40 & 32/60 & 55/100 \\
OpenVLA-OFT & 23/40 & 21/60 & 44/100 \\
GR00T N1.5 & \underline{34/40} & 29/60 & 63/100 \\
Fast-WAM & \underline{34/40} & \underline{39/60} & \underline{73/100} \\
Cosmos Policy & 33/40 & 38/60 & 71/100 \\
\textbf{\method{}} & \textbf{39/40} & \textbf{50/60} & \textbf{89/100} \\
\bottomrule
\end{tabularx}

\paragraph{Matched single-arm perturbations.}
The 480 additional deployments cover five T1 and three T2 conditions, three policies and 20 trials per condition and policy. Each condition changes one factor in the nominal scene--instruction pairs. For the camera shift, ten trials move the external camera horizontally and yaw it $15^\circ$; ten raise it and pitch it $10^\circ$. The wrist camera stays on its mount, and the policy keeps the nominal extrinsics of the moved external camera. Two unseen joint configurations each provide ten start-pose trials. A held-out lookalike cup leaves the instruction unambiguous; in the occlusion condition, a screen hides one-third to one-half of the target from the external camera without blocking the grasp. The background condition changes only the tablecloth.

\begin{table}[H]
\centering
\caption{\textbf{Single-arm perturbations on PiPER.} Entries give successes/20 (change in percentage points from the same policy and task under nominal conditions). Per column, \textbf{bold} and \underline{underlining} mark the best and second-best counts and changes separately, including ties.}
\label{tab:real-perturbations}
\small
\setlength{\tabcolsep}{2pt}
\begin{tabularx}{\linewidth}{@{}l*{8}{Y}@{}}
\toprule
 & \multicolumn{5}{c}{T1: cup to tray} & \multicolumn{3}{c}{T2: cup to basket}\\
\cmidrule(lr){2-6}\cmidrule(lr){7-9}
Method & Camera & Start pose & Lookalike cup & Occlusion & Tablecloth & Camera & Start pose & Tablecloth\\
\midrule
$\pi_{0.5}$ & 4/20\,($-50$) & 6/20\,($-40$) & \underline{10/20}\,($\underline{-20}$) & 9/20\,($\underline{-25}$) & 13/20\,($\underline{-5}$) & 2/20\,($\underline{-35}$) & 4/20\,($\underline{-25}$) & 6/20\,($\underline{-15}$) \\
Fast-WAM & \underline{10/20}\,($\underline{-35}$) & \underline{12/20}\,($\underline{-25}$) & 9/20\,($-40$) & \underline{13/20}\,($\mathbf{-20}$) & \underline{16/20}\,($\underline{-5}$) & \underline{9/20}\,($-40$) & \underline{11/20}\,($-30$) & \underline{14/20}\,($\underline{-15}$) \\
\textbf{\method{}} & \textbf{15/20}\,($\mathbf{-20}$) & \textbf{18/20}\,($\mathbf{-5}$) & \textbf{19/20}\,($\mathbf{0}$) & \textbf{15/20}\,($\mathbf{-20}$) & \textbf{19/20}\,($\mathbf{0}$) & \textbf{14/20}\,($\mathbf{-30}$) & \textbf{18/20}\,($\mathbf{-10}$) & \textbf{19/20}\,($\mathbf{-5}$) \\
\bottomrule
\end{tabularx}
\end{table}

\method{} achieved the highest success in all eight conditions and the smallest drop from its nominal success in each, tied with Fast-WAM under occlusion (Tab.~\ref{tab:real-perturbations}). On T1, it completed 19/20 trials in nominal scenes, with the lookalike cup and with the changed tablecloth; in the last two, Fast-WAM fell to 9/20 and 16/20. Moving the external camera reduced success for all three policies on both tasks; \method{} retained 15/20 on T1 and 14/20 on T2, losing 20 and 30 points, while the baselines lost 35 to 50 points.

\begingroup
\setlength{\intextsep}{3pt}
\setlength{\abovecaptionskip}{2pt}
\subsection{Multi-view deployment rollouts on PiPER}\label{app:real-sequences}
Figures~\ref{fig:real-seq-blue}--\ref{fig:real-seq-fruits} show seven closed-loop \method{} rollouts. Figures~\ref{fig:real-seq-blue}, \ref{fig:real-seq-white} and~\ref{fig:real-seq-stack} cover T1 and T3; the other four show additional tabletop tasks excluded from the T1--T5 success rates. Each column shows a third-person view recorded only for visualization (not a policy input), the front view and the active wrist view (top to bottom). Front and wrist frames share a frame index; the third-person view is aligned by arm motion. The eight columns show selected instants of each rollout, from left to right. In the bimanual rollouts (Figs.~\ref{fig:real-seq-stack}--\ref{fig:real-seq-fruits}), labels below the wrist row identify the active arm: the wrist row switches from left to right for stacking and sorting and from right to left for fruit collection.

\begin{figure}[H]
  \centering
  \includegraphics[page=1,width=\linewidth]{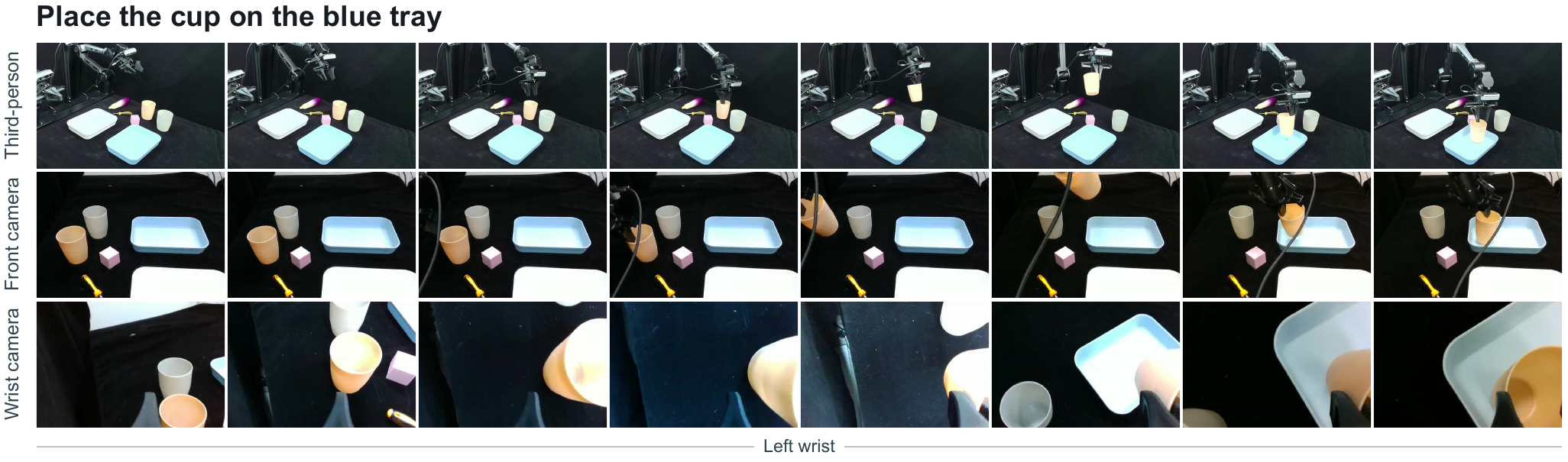}
  \caption{\textbf{Cup placement on the blue tray (T1).} The left gripper lifts the cup and releases it upright on the blue tray.}
  \label{fig:real-seq-blue}
\end{figure}

\begin{figure}[H]
  \centering
  \includegraphics[page=2,width=\linewidth]{figures/realworld-appendix-panels.pdf}
  \caption{\textbf{Cup placement on the white tray (T1).} The left arm places the cup on the white tray beside the blue one.}
  \label{fig:real-seq-white}
\end{figure}

\begin{figure}[H]
  \centering
  \includegraphics[page=3,width=\linewidth]{figures/realworld-appendix-panels.pdf}
  \caption{\textbf{Block placement in the cream-colored bin.} The left gripper carries one block over the rim and releases it inside.}
  \label{fig:real-seq-cream}
\end{figure}

\begin{figure}[H]
  \centering
  \includegraphics[page=4,width=\linewidth]{figures/realworld-appendix-panels.pdf}
  \caption{\textbf{Block placement in the orange bin.} The wrist view follows the block from the tabletop into the orange bin.}
  \label{fig:real-seq-orange}
\end{figure}

\begin{figure}[H]
  \centering
  \includegraphics[page=5,width=\linewidth]{figures/realworld-appendix-panels.pdf}
  \caption{\textbf{Bimanual block stacking (T3).} The left arm positions the red base block (columns 1--4); the right arm places the blue block on top (columns 5--8). For display only, the same color correction is applied to all eight front-camera frames.}
  \label{fig:real-seq-stack}
\end{figure}

\begin{figure}[H]
  \centering
  \includegraphics[page=6,width=\linewidth]{figures/realworld-appendix-panels.pdf}
  \caption{\textbf{Sorting produce between a tray and a bowl.} The left arm places a red item on the tray and an eggplant in the bowl (columns 1--4). The right arm adds a banana to the tray and a yellow item to the bowl (columns 5--8).}
  \label{fig:real-seq-sort}
\end{figure}

\begin{figure}[H]
  \centering
  \includegraphics[page=7,width=\linewidth]{figures/realworld-appendix-panels.pdf}
  \caption{\textbf{Collecting three pieces of fruit on one tray.} The right arm moves the yellow and orange fruit to the white tray (columns 1--5); the left arm adds the green fruit (columns 6--8).}
  \label{fig:real-seq-fruits}
\end{figure}
\endgroup

\end{document}